%% file: ms.tex
\documentclass[acmtog,nonacm]{acmart}

\AtBeginDocument{%
  \providecommand\BibTeX{{%
    \normalfont B\kern-0.5em{\scshape i\kern-0.25em b}\kern-0.8em\TeX}}}

\input{common-packages}

\begin{document}

\title{3-D Material Style Transfer for Reconstructing Unknown Appearance in Complex Natural Materials}


\author{Shashank Ranjan}
\author{Corey Toler-Franklin}
\orcid{1234-5678-9012-3456}
\email{ctoler@cise.ufl.edu}
\affiliation{%
\institution{University of Florida}
\country{USA}}

\renewcommand{\shortauthors}{Toler-Franklin, C.  et al.}

%------------------------------------------------------------------------

\input{format-arxiv-body.tex}

%------------------------------------------------------------------------

\bibliographystyle{ACM-Reference-Format}
\bibliography{references.bib}

%------------------------------------------------------------------------
\input{appendix}

%------------------------------------------------------------------------

\end{document}

%% file: common-packages.tex
\usepackage{framed,multirow}

\usepackage{amssymb}
\usepackage{latexsym}

\usepackage{url}
\usepackage{xcolor}
\definecolor{newcolor}{rgb}{.8,.349,.1}

\usepackage{hyperref}

\usepackage[switch,pagewise]{lineno} %Required by command \linenumbers below

\usepackage{epsfig}
\usepackage{graphicx}
\usepackage{amsmath}
\usepackage{amssymb}
\usepackage{bm}
\usepackage[english]{babel}
\usepackage{commath} 
\usepackage{hhline} 
\usepackage{colortbl}
\usepackage[ruled]{algorithm2e} % For algorithms
\renewcommand{\algorithmcfname}{ALGORITHM}
\SetAlFnt{\small}
\SetAlCapFnt{\small}
\SetAlCapNameFnt{\small}
\SetAlCapHSkip{0pt}
\IncMargin{-\parindent}
\usepackage{adjustbox}
\usepackage{multirow}
\UseRawInputEncoding
\usepackage{caption}
\usepackage[maxfloats=40]{morefloats}
\usepackage{subfigure}
\usepackage{upgreek}

\definecolor{LightGray}{gray}{0.9}

%% file: format-arxiv-body.tex
%------------------------------------------------------------------------

\input{format-CAG-abstract}

%------------------------------------------------------------------------
%------------------------------------------------------------------------
\input{subjectcategories}

\input{keywords}

%------------------------------------------------------------------------

%------------------------------------------------------------------------
\input{fig-teaser}

%------------------------------------------------------------------------

\maketitle

%------------------------------------------------------------------------
\input{intro}

%------------------------------------------------------------------------
%------------------------------------------------------------------------
\input{previouswork}

%------------------------------------------------------------------------
	
%------------------------------------------------------------------------
\input{overview}
%------------------------------------------------------------------------

%------------------------------------------------------------------------
\input{uvtexturecapture}

%------------------------------------------------------------------------

%------------------------------------------------------------------------
\input{model}

%------------------------------------------------------------------------

%------------------------------------------------------------------------
\input{results}

%------------------------------------------------------------------------

%------------------------------------------------------------------------
\input{applications}

%------------------------------------------------------------------------

%------------------------------------------------------------------------
\input{conclusion}

%------------------------------------------------------------------------

%------------------------------------------------------------------------
\input{acknowledgements}

%------------------------------------------------------------------------

%% file: format-CAG-abstract.tex
\begin{abstract}
\input{abstract-body}

\end{abstract}

%% file: abstract-body.tex
We propose a \hbox{3-D} material style transfer framework for reconstructing invisible (or faded) appearance properties in complex natural materials. Our algorithm addresses the technical challenge of transferring appearance properties from one object to another of the same material when both objects have intricate, noncorresponding color patterns. Eggshells, exoskeletons, and minerals, for example, have patterns composed of highly randomized layers of organic and inorganic compounds. These materials pose a challenge as the distribution of compounds that determine surface color changes from object to object and within local pattern regions. Our solution adapts appearance observations from a material property distribution in an exemplar to the material property distribution of a target object to reconstruct its unknown appearance. We use measured reflectance in \hbox{3-D} bispectral textures to record changing material property distributions. Our novel implementation of spherical harmonics uses principles from chemistry and biology to learn relationships between color (hue and saturation) and material composition and concentration in an exemplar. The encoded relationships are transformed to the property distribution of a target for color recovery and material assignment. Quantitative and qualitative evaluation methods show that we replicate color patterns more accurately than methods that only rely on shape correspondences and coarse-level perceptual differences. We demonstrate applications of our work for reconstructing color in extinct fossils, restoring faded artifacts and generating synthetic textures.

%% file: subjectcategories.tex

\ccsdesc[500]{Computing methodologies~Texturing}
\ccsdesc[500]{Computing methodologies~Mesh models}
\ccsdesc[500]{Computing methodologies~Mesh geometry models}


%% file: keywords.tex
%\begin{keyword}
%% Keywords
%\KWD \hbox{3-D} Style Transfer \sep Texture Mapping \sep Image %Registration \sep Image Enhancement \sep Multi-Spectral Imaging  \sep %\hbox{3-D} Scanning, Conformal Geometry \end{keyword}

\keywords{\hbox{3-D} Style Transfer, Texture Mapping, Image Registration, Image Enhancement, Multi-Spectral Imaging, \hbox{3-D} Scanning,  Conformal Geometry }

%% file: fig-teaser.tex
\begin{teaserfigure}
\input{teaser-body}

\end{teaserfigure}

%% file: teaser-body.tex
\centering
\includegraphics[width=1.0\hsize]{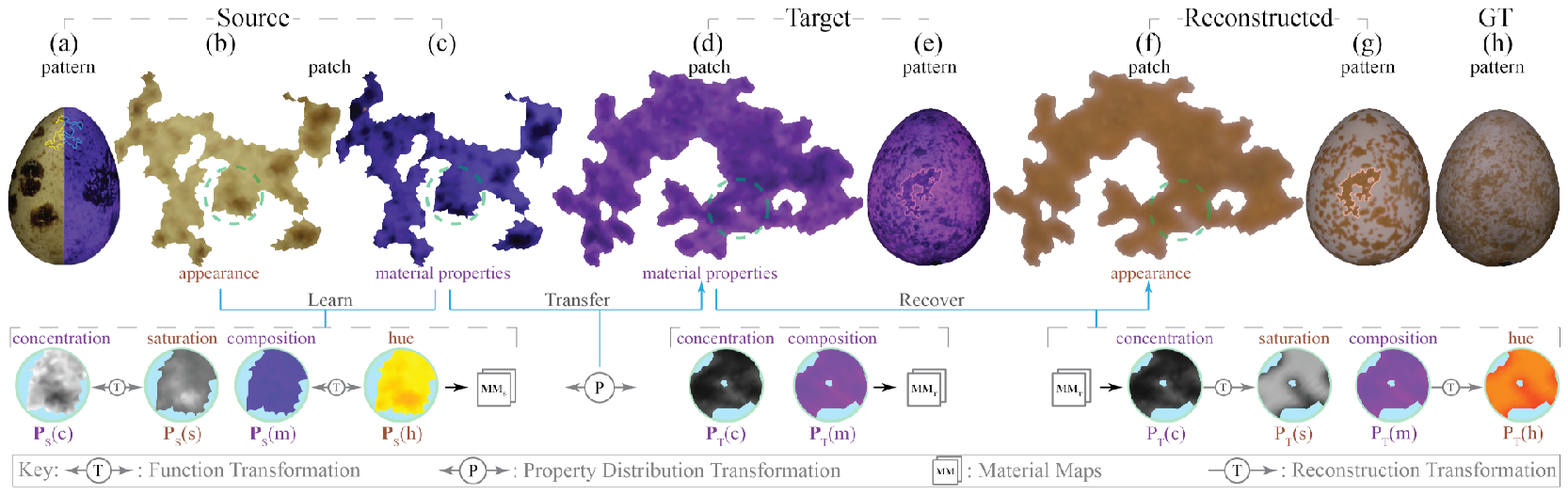}
\caption{\label{fig:teaser}%
A conceptual overview of our approach. We learn the relationships between (c) the material structure and (b) the appearance on a source patch and adapt it to (d) the material structure of a target patch to reconstruct (f) its appearance.}

%\centering
%\includegraphics[width=0.9\hsize]{../images/teaser/teaser_v007.eps}
%\caption{\label{fig:teaser}%
%The diagram shows an application of our work for reconstructing color %patterns in faded (colorless) shell fossils. Our algorithm uses known %relationships between diffuse reflection and physically-based measured %material properties (composition and concentration) to transfer \hbox{3-%D} color patterns from a source object (c) to a target object (d-e) even %when there are non-corresponding color variations.  Although patches on %the modern  \emph{cone shell} (c) have similar shapes to patches on a %shell fossil of an ancestral species (e), the color variations within %pattern boundaries will differ. We compute a mapping between changes in %color saturation (f) and material concentration recorded in a bispectral %reflectance map (e and g) and formulate transfer functions to %reconstruct hue, saturation and value (a) for the new  \hbox{3-D} fossil %texture (b).}

%% file: intro.tex
\section{Introduction}
\label{sec:intro}
Creating textures that replicate highly randomized color patterns found in natural objects in a manner that appears authentic is challenging~\cite{Walter2012,chen2012,akeninemoeller2006}. The problem is compounded when objects have faded colors or missing details due to aging or exposure to environmental elements~\cite{Bellini2016,lu2007}. Low-resolution color textures in \hbox{3-D} models may also require detail enhancement~\cite{Gheche2017TextureRG}. Consider how you would reproduce the pattern in the avion eggshell in Figure~\ref{fig:teaser}h if the original had little or no color. Even with automated optimizations~\cite{Ray2010}, generating uv-maps (\hbox{2-D} projections of \hbox{3-D} geometry) with standard texture mapping requires extensive manual effort, post processing and prior knowledge of the original color~\cite{Yuksel2017}. Less cumbersome alternatives~\cite{Mallett2019,Mallett2020,Yuksel2019} use mesh colors and hardware acceleration to avoid uv-mapping, but still require appearance data from a source. Procedural methods produce reasonably randomized textures~\cite{LAGAE2010312,Galerne2012,Schuster2020} but are not adaptable to the original pattern if its color is unknown and the pattern is unique. 

Style transfer methods that transfer shape and appearance from an exemplar texture on a source model to a target model, while retaining the target's structure~\cite{Bruckner2007,nguyen2012,ma2014}, are better suited to solve this problem. Current methods use image-space features from the source, and color and shape priors from the target for finding correspondences. These methods work well for synthesizing textures for synthetic objects even when the source and target have different typologies~\cite{chen2012}. However, little work has been done for pattern-to-pattern style transfer where changes in hue and saturation within local regions (patches) on both objects are complex (Figure~\ref{fig:teaser}b and f), and only traces (if any) of color information exists on the target. Context aware methods~\cite{lu2007} use observed changes in the source geometry (due to chemical processes) to alter the target's shape and appearance. We aim to restore the target's original color pattern. 
 
We propose a physically based solution that transfers appearance properties from an exemplar to a target object in a manner that conforms to the target's unique material property distribution without prior knowledge of the target's appearance. Preserving material properties is important when color restoration (of bio-materials and artifacts) must permit material analysis.  We build upon two concepts from biology and chemistry. First, photometric behaviors like fluorescence~\cite{FecheyrLippens2015,Hullin2010a} can be used to measure material properties even when color is eroded~\cite{Hendricks2015}. Second, the mixture of constituent materials and their molecular density directly influences surface appearance~\cite{thomas2015,hedegaard06,lewis1939color}.

A custom measurement system records material properties for two objects of the same material. We use spherical harmonics to learn relationships between observed appearance and changes in measured material composition and concentration in an exemplar (Figure~\ref{fig:teaser}a-c). We reparameterize the encoded relationships to correspond to the target's material distribution (Figure~\ref{fig:teaser}d) before reconstructing it's hue and saturation (Figure~\ref{fig:teaser} \emph{bottom right}), and assigning materials and shape details (Figure~\ref{fig:teaser} f). Areas of high material concentration appear more saturated, and different mixtures of compounds have noticeably different hues (Figure~\ref{fig:teaser}g reconstruction, h ground truth). 

Our style transfer approach will be a valuable contribution to computer graphics applications for texture reconstruction~\cite{Gheche2017TextureRG}and detail enhancement~\cite{berkiten_learning_2017}. Our contributions include: 

\begin{itemize}

\item An ultraviolet illumination system that captures \hbox{3-D}  patterns from continuous strips of bispectral reflectance. 

\item The use of ultraviolet radiation for measuring material properties, and the separation of material composition and concentration from bispectral reflectance maps.

\item A method for using spherical harmonics to learn relationships between property distribution functions of observed object appearance (hue and saturation) and measured material properties (composition and concentration). 

\item A material transformation that adapts learned relationships between the appearance and material property distribution functions in a source (exemplar) to the material property distribution function in a target object for reconstructing its unknown appearance (hue, saturation and shape detail). 

\end {itemize}

%% file: previouswork.tex
\section{Related Work}
\label{sec:prevouswork}

Our work combines style (shape and appearance) transfer with methods that capture shapes and textures from real objects. Here we review relevant related work in these areas.

\paragraph{\textbf{Style Transfer}} Our work relates to \hbox{3-D} material style transfer methods that transfer image-based features from a source (image or video) to a target \hbox{3-D} scene using a two-phase process: material extraction and material assignment using combinatorial optimizations~\cite{nguyen2012}. Shape analogies~\cite{bohan2000ludi} from cognitive science studies facilitate style transfer in \hbox{3-D} shapes~\cite{ma2014} by examining structural differences between the source and target. Style transfer functions have been combined with image-based relighting~\cite{Bruckner2007} or other shading methods to relate shapes, color and illumination for stylized rendering~\cite{Sloan01,Fiser16}.

Our framework shares similarities with data-driven methods that create styles by imaging time-varying changes in materials. Context aware methods observe changes in geometry due to chemical processes (like rusting) and apply these styles to synthetic objects~\cite{lu2007}. Reflectance exemplars have also been used to model weathered materials~\cite{Wang2006}. Our work is distinguishable from these examples because we image physical changes in materials exposed to ultraviolet radiation~\cite{wiley2011} and adapt learned relationships between material structure and surface appearance~\cite{lewis1939color} on a source to an existing target.

Surface texture transfer methods map coarse and fine-scale patterns from a texture on one model to textures on another~\cite{chen2012}. Most examples transfer patterns comprised of two solid colors to a blank slate (a model without materials). These methods do not extend to natural patterns which have modulating colors and patterns with different frequencies. 

Detail transfer techniques add complexity to simple \hbox{3-D} models like those found in digital repositories~\cite{wang16}. These methods learn to predict similarity combinations in high-quality~\hbox{3-D} models, and then transfer learned geometric features to the target model~\cite{berkiten_learning_2017}. Our detail enhancement term is derived directly from material measurements in our data. 

Color reconstruction is an important application of our work. Example-based color transfer~\cite{Chang2004} applies perceptual metrics to cluster colors, and optimizes color differences between images~\cite{Hou2007}. Enhancement methods restore texture detail~\cite{Gheche2017TextureRG}. We use measured materials to compute missing color data.

\paragraph{\textbf{Texture Synthesis}} Our work mapping \hbox{3-D} material property distributions to surface color relates to spatially varying texture synthesis methods that correlate variations in geometric shape with diffuse color. Examples use dimensionality reduction of overcomplete feature sets~\cite{akeninemoeller2006}, local neighborhood examples~\cite{Lefebvre2005}, \hbox{2-D} texture samples~\cite{Schuster2020} or texture extraction~\cite{lai05} to create geometry consistent textures with minimal artifacts. 

Biologically inspired simulations use procedural methods to simulate reaction-diffusion systems, cellular automata, and pigmentation patterns from living beings ~\cite{Malheiros2017}. The resulting \hbox{2-D} texture patterns are plausible and can be used to evaluate other texture simulation methods. However, uv-mapping is required for \hbox{3-D} models, and the process does not permit material analysis or replication of a specific object. Clonal mosaics~\cite{Walter2012} link changes in shape with changes in appearance, but are optimized for two-color patterns in fur. Other rule-based procedural textures incorporate noise, randomized patterns~\cite{Galerne2012}, or seeds~\cite{efros1999texture}, and in some cases exemplars~\cite{LAGAE2010312}. 

Neural networks have been used to synthesize high-resolution multi-scale textures from exemplars~\cite{snelgrove2017high} as well as transfer learned style between images~\cite{Ulyanov2016} or from one model to another. Neural networks for data-driven analysis of materials~\cite{LING201719} is also becoming common in material science.

\paragraph{\textbf{Texture Mapping}} We build upon a broad area of image-based methods that map coordinates of textured patterns to arbitrary \hbox{3-D} shape. Well-known challenges with artifacts, seams~\cite{Ray2010} and distortions complicate an already time-consuming manual process. Alternatives to traditional methods~\cite{Yuksel2019} like mesh colors~\cite{Mallett2019,Mallett2020} provide significant speed-ups like hardware acceleration~\cite{Yuksel2016,Yuksel2017}, but still require source appearance. 

\paragraph{\textbf{Conformal mapping}} We use conformal mapping~\cite{gu2004b,Yau2016ComputationalCG} from differential geometry to compare complex shapes. Based on the theory of Riemann surfaces, these methods map one genus zero surface to another~\cite{haker,xianfeng,levy} using angle and scale preserving methods~\cite{wang2005uniform}. Other methods find surface correspondences by solving PDEs over \hbox{4-D} hypersurfaces~\cite{dinh2005}.

\paragraph{\textbf{Material Analysis}} We incorporate principles in physical science that relate material properties to colors in nature~\cite{lewis1939color}. In Section~\ref{sec:materialanalysis} we discuss  how fluorescence~\cite{wiley2011} can record material properties and explain how these properties are associated with appearance. Throughout the paper we show examples from biological and geological specimens~\cite{Tullett1975,FecheyrLippens2015,thomas2015,Brulez2016,Hendricks2015,williams2017,KROGER1948} common in natural history collections~\cite{page2015}. 

\paragraph{\textbf{Multispectral Imaging Systems}} Our acquisition system can be categorized with computer graphics systems that use controlled illumination to capture geometry and reflectance. Related work combines multispectral imaging and \hbox{3-D} scanning to measure hyperspectral patterns on solid objects for biological studies~\cite{minim2012}. Radiometric measurement systems combine multispectral imaging with color transforms for pigment identification~\cite{minkim2011}. Spherical harmonic illumination~\cite{tunwattanapong2013} and other multi-illumination photography systems recover diffuse and specular shape attributes for photorealistic relighting~\cite{Malzbender01,tolerfranklin2021,rgbn07}. 

We use bispectral reflectance~\cite{wiley2011, Lakowicz2006} to capture data for our material analysis. Bispectral measurement systems have been used to measure spatially varying bispectral bidirectional reflectance and reradiation distribution functions (BBRDFs)~\cite{Hullin2010b,Wilkie2006} for spectral rendering. Monochromators measure bispectral radiance factors to reconstruct fluorescent object appearance under arbitrary lighting~\cite{tominaga2017appearance}. Typically, these methods require complex, time intensive measurements at multiple wavelengths. Bispectral reflectance improves shape recovery~\cite{Hullin2008b} during \hbox{3-D} scanning. Other variants have been used for chemical analysis in physical science disciplines~\cite{wiley2011,Lakowicz2006,thomas2015,FecheyrLippens2015}.

%% file: overview.tex
\section{Overview}
\label{sec:overview}

Here we introduce our pattern characterization including a definition of terms and associated data types. We also provide an overview of our style transfer algorithm.

Materials in our dataset exhibit complex patterns composed of layers of naturally occurring inorganic and organic compounds (like proteinaceous pigments) that fluoresce when exposed to ultraviolet radiation~\cite{Brulez2016,thomas2015,KROGER1948}. Fluorescence~\cite{ wiley2011} occurs when shortwave radiation hitting a surface is almost instantaneously emitted as longer waves of visible light (bispectral reflectance~\cite{Hullin2010b}). A description of the biological materials and minerals in our dataset is provided in Section~\ref{sec:materialanalysis}. 
\input{fig-nomenclature}
\input{fig-overview}

Figure~\ref{fig:nomenclature} illustrates our pattern characterization. A patch is a distinctly identifiable region of similar color. A pattern consists of a background patch of a solid color and randomly disbursed foreground patches that have modulating color. The background is sometimes visible through thin material layers in the foreground. A patch is represented as a \hbox{3-D} mesh, a \hbox{2-D} boundary contour, a \hbox{3-D} diffuse reflectance map, and a \hbox{3-D} bispectral reflectance map. The mesh and boundary contour are used to compute area and shape properties respectively, with a curvature property computed as derivatives of the surface normal. Diffuse reflectance encodes appearance properties like color (hue and saturation). Bispectral reflectance encodes material properties like the type of molecules in the material (composition) and molecular density (concentration).

Figure~\ref{fig:overview} (and our supplemental video) presents an overview of our algorithm. We start with a source and a target object of the same material. Our material measurement system (Section~\ref{sec:materialanalysis}) captures and computes per-vertex appearance properties of the source and per-vertex physical material properties for both the source and the target. Next, conformal mapping projects these per-vertex properties into spherical space. An optional alignment step aligns corresponding points to orient the objects. Patches are extracted over appearance and material properties. A pattern matching technique uses shape and material similarity functions to assign each target patch to the closest matched source patch. We use spherical harmonics to compute property distribution functions over the per-vertex appearance and material property values and ultimately learn the relationship between the two distribution functions. A novel material transform transfers this learned relationship to the target, using the known material property distribution of the target to infer the unknown appearance. Materials are transferred from the source to the target, and an inverse mapping from spherical space to object space is applied to generate the final result. 

%% file: fig-nomenclature.tex
\begin{figure}[t]
\centering
\includegraphics[width=1.0\linewidth]{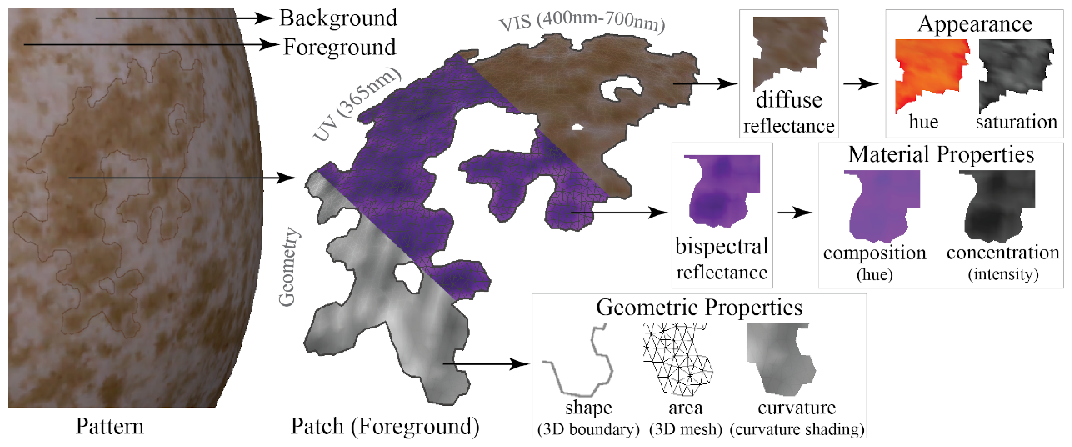}
\caption{\label{fig:nomenclature}%
Pattern Characterization}
\end{figure}

%% file: fig-overview.tex
\begin{figure*}[t]
\centering
\includegraphics[width=1.0\linewidth]{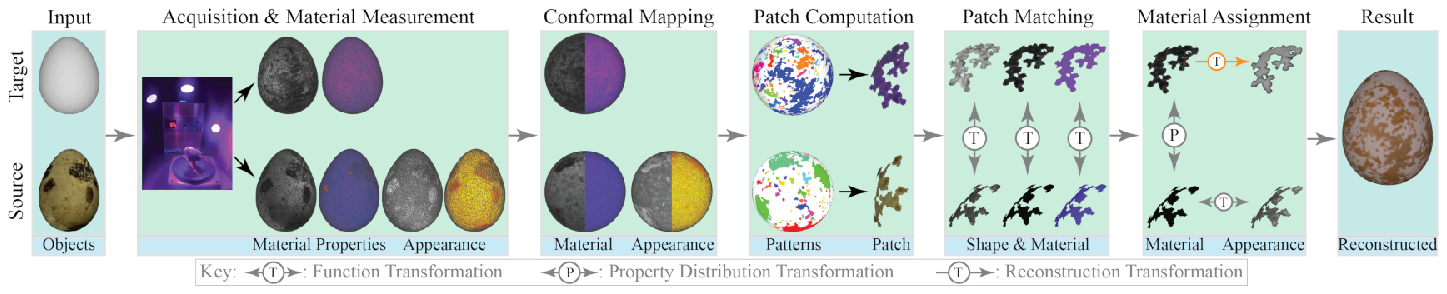}
\caption{\label{fig:overview}%
An overview of the style transfer process.}
\end{figure*}

%% file: uvtexturecapture.tex
\section{Material Properties and Appearance}
\label{sec:materialanalysis}

We propose a \hbox{3-D} material style transfer framework for complex natural materials that adapts appearance observations from a material property distribution in one object to the material property distribution of another to reconstruct its unknown appearance. To accurately simulate intricate color patterns, it is necessary to understand how the material structure of an object influences its surface appearance. In this section, we introduce the materials in our dataset, and for completeness, identify physical relationships from the biology and chemistry literature upon which we build our mathematical model.

\input{fig-uv-3dtexexamples}

Color patterns in our data are formed by compounds that have accumulated in layers due to biological~\cite{hedegaard06} or geological processes~\cite{KROGER1948}. Such natural materials are common in biological specimens and minerals like plant leaves, avion eggshells, cone shells and resinous coal (Figure~\ref{fig:uv-3dtexexamples}). We work with fluorophores which are found in many natural materials such as carbonates.

Photometric properties of the substances in our data provide quantifiable information about the underlying material structure. Fluorescence occurs when microscopic structures absorb shortwave radiation like ultraviolet waves ($100nm-400nm$) and almost instantaneously emit longer wavelengths in the visible spectrum ($400-700nm$)~\cite{Jun2018}. This occurs when an orbital electron of a molecule transitions from a state of excitation to a relaxed ground state by emitting a photon~\cite{wiley2011}. This bispectral reflectance~\cite{Hullin2010a} is shown in Figure~\ref{fig:uv-3dtexexamples} \emph{bottom row}.

\subsection{Composition:} 
Fluorophores are uniquely identifiable by their absorption and emission spectra~\cite{wiley2011}. Thus, different materials will emit different colors in the bispectral map. The difference in hue between foreground and background patches on shell fossils (Figure~\ref{fig:patternextract}c) is a classic example~\cite{Hendricks2015}. We measure material composition as the hue (in hsv color space) in the bispectral map. 

\subsection{Concentration}
The relationship between emitted fluorescent intensity $I_{f}$ and material (fluorophore) concentration $c$ is defined as:

\begin{equation}
I_{f} = kI_{o}\phi\varepsilon bc
\label{equ:concentration}
\end{equation}

\noindent where $I_{o}$ is the intensity of the incident radiation, $\phi$ is the quantum yield which is the ratio of emitted photons to absorbed photons ~\cite{wiley2011},  $\varepsilon$ is the material's molar absorptivity which measures how well a chemical species absorbs a given wavelength of light, $b$ is the length of the incident ray's path through the material, and $k$ is a system dependent constant of proportionality. We know that higher concentration increases the scattering coefficient of a substance resulting in a higher emission intensity~\cite{Hullin2008a}. In chemistry, this relationship between fluorescent emission intensity and material concentration is considered directly proportional~\cite{wiley2011,jung2018} as concentration levels are usually within a low threshold. Rare exceptions occur for very large material concentrations (we do not account for this). We use the intensity of the fluorescent emission to record concentration.

\subsection{Shape Detail}
Coarse and fine level shape detail are important for realism in reconstructed textures~\cite{wang16,chen2012,lai05}. As material concentration and composition change on the surface, the orientation of surface normals are also changing. Changes in shape are clearly visible in bispectral reflectance maps (see ridges at the shell base and side in Figure~\ref{fig:patternextract}c). Our light positions maximize contrast between foreground and background patches (Section~\ref{sec:measurementsystem}).  We use the light-dependent value component of the bispectral reflectance (in hsv color space) to record shape detail. 

\input{fig-chemistry}

\subsection{Pattern Coloration}

Continuing with our avion eggshell example, Figure~\ref{fig:chemistry}  illustrates how material composition and concentration influences pattern coloration~\cite{Brulez2016,lewis1939color}. Eggshells consists of $4\%$ organic and $96\%$ inorganic material~\cite{FecheyrLippens2015}. The latter is $98\%$ calcium carbonate with calcium phosphates and metal ions forming the remainder.  A top layer is composed of a mixture of two tetrapyrrole pigments, protoporphyrin IX and biliverdin~\cite{thomas2015,Brulez2016}. The color (hue) of this layer changes as the relative proportions of the two pigments change~\cite{FecheyrLippens2015}. In Figure~\ref{fig:chemistry}, changing the ratio of the pigments molecules from 1:3 to 1:2 influcences coloration. Differences in these ratios make the source bispectral reflectance bluish compared to the target's in Figure~\ref{fig:teaser}c and d.

The relationship between molecular concentration and color saturation is well-known in Raman Spectroscopy~\cite{hedegaard06,thomas2015}, a method for identifying different compounds in pigments by studying the elastic scattering effects in photons~\cite{wiley2011, Lakowicz2006}. Color saturation varies with the overall concentration of the pigments on the surface~\cite{hedegaard06}. Color appears more vibrant as pigment concentration increases~\cite{hedegaard06}.  Figure~\ref{fig:chemistry} \emph{bottom row} increases concentration by doubling the molecular count. Other proteins or nanostructures may contribute to coloration by selectively absorbing certain wavelengths, or enhancing light reflectance~\cite{FecheyrLippens2015}. Table~\ref{fig:material_measurements} summarizes the associations between material properties and appearance.

\input{fig-material_measurements}

\section{Acquisition and Measurement System}
\label{sec:measurementsystem}

When capturing appearance and materials for our data, it is difficult to find registration points to associate material measurements extracted from bispectral reflectance maps with \hbox{3-D} geometry on colorless objects. Our custom ultraviolet illumination system resolves this issue by capturing low resolution \hbox{3-D} fluorescent patterns to which we align patterns in the bispectral reflectance maps before extracting materials. 

\subsection{\hbox{3-D} Pattern Acquisition}
\label{sec:patternaquisition}

Our illumination configuration exposes the top (camera-facing) surface of an object to point sources of incident ultraviolet radiation. The ultraviolet radiation is produced by a $1-1.5ft$ diameter hemispherical band of five fixed ultraviolet LED arrays (Figure~\ref{fig:3dscanneruv}a and Figure~\ref{fig:lights}), mounted in tripods and focused toward the object from above. A calibration step (discussed below) positions and orients each LED to produce continuous bands of bispectral reflectance on the surface.

LEDs of the same wavelength are selected according to the material's absorption and emission spectra. We work in a light-sealed darkroom to restrict surfaces to this excitation wavelength. We can measure a broad range of materials by including UVA ($315nm-400nm$) and UVC ($100nm-280nm$) sources at $395nm$ (LEDwholesalers 51-LED ultraviolet), $365nm$ (Ultrafire A100 LED) and $254nm$ (Raytech ultraviolet $365nm$,$254nm$). The latter is linear and used for precision measurements only. 

The object to be scanned sits on a multidrive, a two axis programable rotating platform, attached to a Next Engine HD \hbox{3-D} laser triangulation scanner centered beneath the LED arrays. One multidrive axis rotates $360^{\circ}$ and the other has an $80^{\circ}$ tilt range. The object is rotated to $28$ positions for tilt angles between $-35^{\circ}$  to $35^{\circ}$ at $2.5^{\circ}$ intervals for four orientations: $0^{\circ}$, $90^{\circ}$, $180^{\circ}$, $270^{\circ}$. The scanner's camera (one of two visible $3.0$ megapixel CMOS sensors) is $9.5$ inches from the object center. For each scan orientation we image a continuous pattern strip (Figure~\ref{fig:3dscanneruv}b) for all exposed object surfaces.

The overlapping strips ($112$ total) are aligned with \emph{Geomagic Design X} (3D Systems). Mesh smoothing in post-processing removes seams between scans along lines of maximum curvature where luminance shifts abruptly. The final texture is a pattern map with fluorescent emission from the pattern and reflected emission from the scanner's flash. A final scanning pass captures the visible diffuse reflection to complete the model.

\input{fig-uvcapturesystem}

\input{fig-lights}

We use low-cost off-the-shelf imaging components and avoid specialized bispectral measurement equipment. This limits our ability to adjust internal camera and light settings. System calibration is required to: (1) account for differences in irradiance in ultraviolet sources, (2) maximize irradiance, (3) maximize pattern contrast and (4) avoid over saturation or underexposure. 

Ultraviolet irradiance at a point on the surface depends on the angle of incidence (Cosine Law) of the source and its distance to the point. Before positioning LED arrays, we compensate for differences in total irradiance from inconsistent battery power and manufacturing disparities. We use a digital ultraviolet meter (General Tools UV513AB $280nm-400nm$) to measure the average irradiance of each source at a fixed distance ($8$ inches) at full charge, adjusting the battery or power until irradiance is consistent ($10\frac{{\mu}W}{cm^{2}}$, $10\frac{{\mu}W}{cm^{2}}$, and $55\frac{{\mu}W}{cm^{2}}$ average irradiance  for $254nm$, $365nm$ and $395nm$ sources respectively).

The intensity and spectra of light emitted from the surface depends on the incoming incident wavelength and its irradiance at the point. The incident angle with respect to surface curvature determines contrast. The sensor was unable to record patterns without a flash. We maximize brightness and contrast between foreground and background patches by positioning LED arrays along three degrees of freedom: $\theta$ and $\phi$ in spherical coordinates, and height ($1-1.5$ft diameter) for the initial view only until a pattern is detectable by the sensor.

\subsection{Measurements from Bispectral Reflectance}

We generate high-resolution bispectral reflection maps $\mathbf{R}_{bis}$ by imaging each surface under one LED array with a DSLR camera. Visible diffuse reflection is also recorded. Using methods from Section~\ref{sec:patternaquisition}, we align each $\mathbf{R}_{bis}$ to the \hbox{3-D} geometry using the pre-aligned \hbox{3-D} pattern textures as correspondences. Figure~\ref{fig:patternextract}a-d shows the colorless diffuse map, acquired \hbox{3-D} pattern, high-resolution \hbox{3-D} bispectral reflectance map and $\mathbf{R}_{bis}$.

Composition is the hue of per-vertex $\mathbf{R}_{bis}$ values in $HSV$ color space. To recover material concentration, we convert per-vertex $\mathbf{R}_{bis}$ values to $YUV$ color space and store luminance $Y$. This represents linear space brightness and image space detail. These measurements work under two conditions: (1) the source is in the UVC range and (2) objects are not shiny. We found that wavelengths of $365nm$ or longer leak visible light producing specularities that cause reconstruction errors (Figure~\ref{fig:limitations_specularity}).  Removing reflection components from $\mathbf{R}_{bis}$ resolves this (Section~\ref{sec:discussion}). It takes about three hours to capture one dataset.

\input{fig-fluorescentTexture}

%% file: fig-uv-3dtexexamples.tex
\begin{figure}[h]
\centering
\includegraphics[width=1.0\hsize]{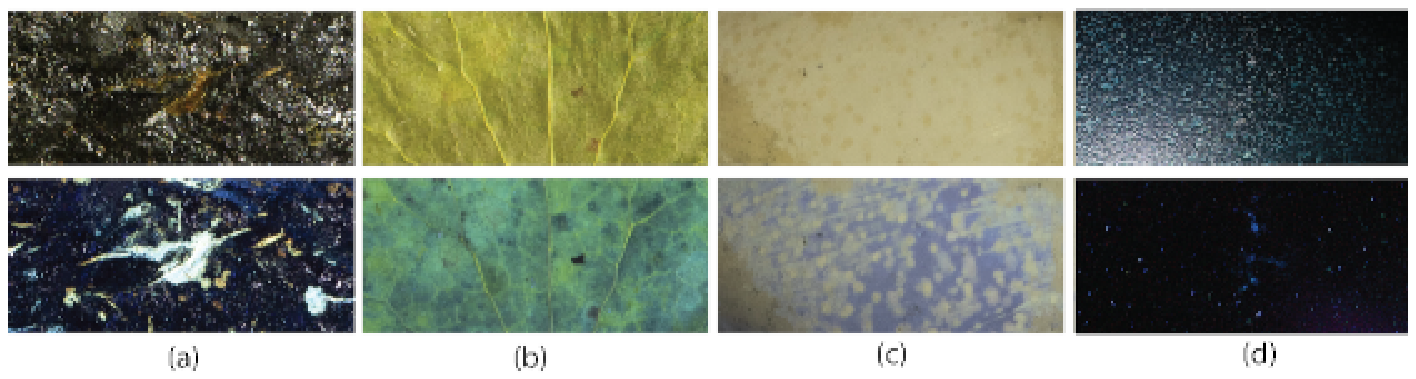}
\caption{\label{fig:uv-3dtexexamples}%
A selection of natural materials from our dataset illuminated with visible LED lights (top row) and then illuminated with a $254nm$ ultraviolet lamp. Organic and inorganic compounds in these composite materials exhibit fluorescence. Photographs were taken with a DLSR camera. (a) resinous coal (b) plant leaf (c) cone shell (d) avian eggshell (Zoom into d to see fluorescence). \emph{Galax urceolata} leaf, FLAS 263100, FLMNH. \emph{Cypraea maculifera} shell, Catalog ID 106972, AMNH. }
\end{figure}

%% file: fig-chemistry.tex
\begin{figure}[h]
  \centering
    \includegraphics[width=1.0\linewidth]{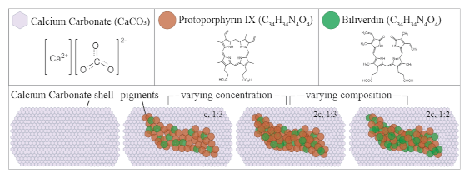}
    \caption[Avian egg structure]{The changes in the molecular composition and density of constituent materials influence hue and saturation observed on the surface. Here, calcite, and mixtures of pigments give the eggshell its color pattern.}
	\label{fig:chemistry}
\end{figure}

%% file: fig-material_measurements.tex
\begin{table}[h]
\setlength\tabcolsep{1.5pt}
\begin{tabular}{p{0.26\linewidth}|p{0.5\linewidth}|p{0.175\linewidth}}
\hline
\small Material Property & \small Measurement & \small Appearance \\
\hline
\small concentration, $c$ & \cellcolor{cyan!10} \small bispectral intensity, $I_{f} = kI_{o}\phi\varepsilon bc$ & \cellcolor{red!10} \small saturation, $s$ \\
\small composition, $m$ & \cellcolor{red!10} \small bispectral hue, $h_{f}$ & \cellcolor{red!10} \small hue, $h$ \\ 
\hline
\end{tabular}
\caption{\label{fig:material_measurements}%
Summary of material property measures and appearance associations. Blue: YUV color space, Red: HSV color space.}
\end{table}

%% file: fig-uvcapturesystem.tex
\begin{figure}[t]
  \centering
    \includegraphics[width=0.8\linewidth]{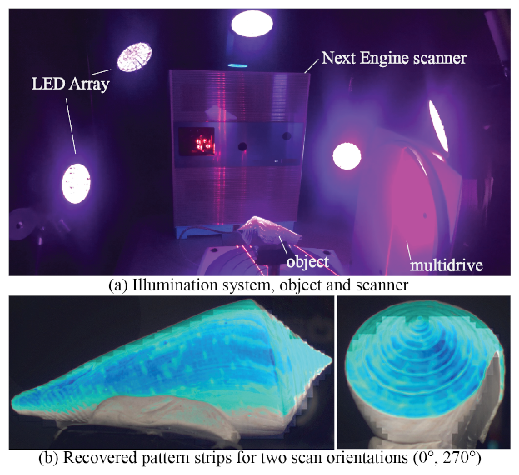}
    \caption[3D UV scanning system]{Scanning system for capturing pattern textures with 3D geometry. \emph{Conus delesertti} fossil, UF117269, FLMNH.}
	\label{fig:3dscanneruv}
\end{figure}

%% file: fig-lights.tex
\begin{figure}[h]
  \centering
    \includegraphics[width=0.8\linewidth]{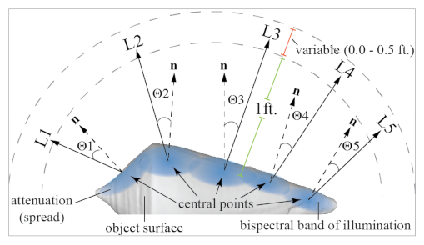}
    \caption[Light Apparatus]{LED configuration.}
	\label{fig:lights}
\end{figure}

%% file: fig-fluorescentTexture.tex
\begin{figure}[h]
\centering
\begin{tabular}{cccc}

\includegraphics[width=0.2\linewidth]{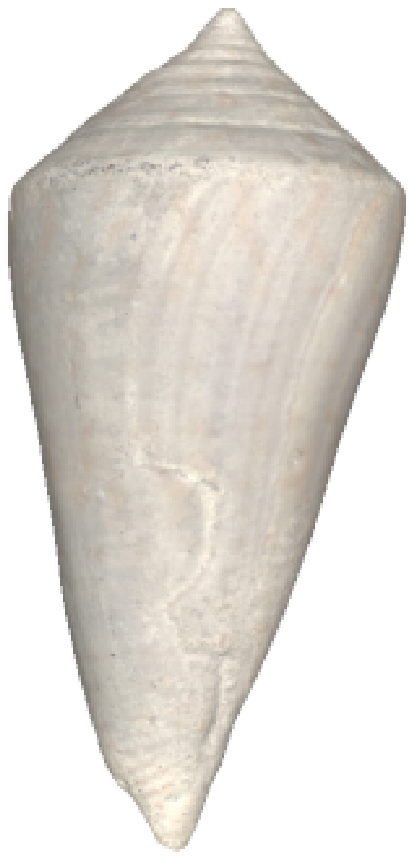}
&\includegraphics[width=0.2\linewidth]{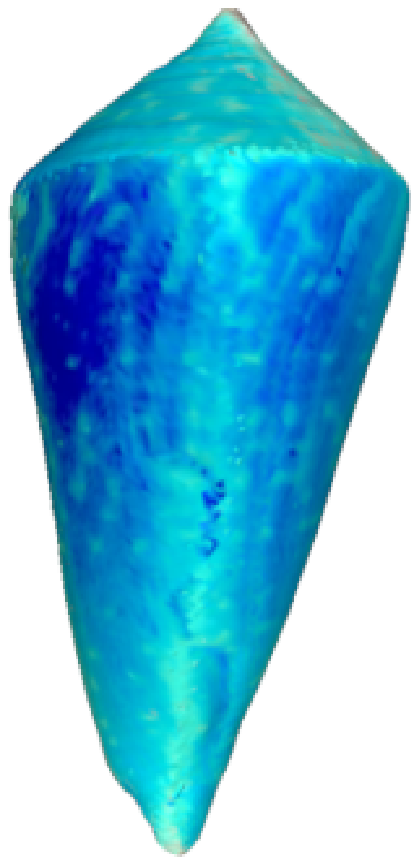}
&\includegraphics[width=0.2\linewidth]{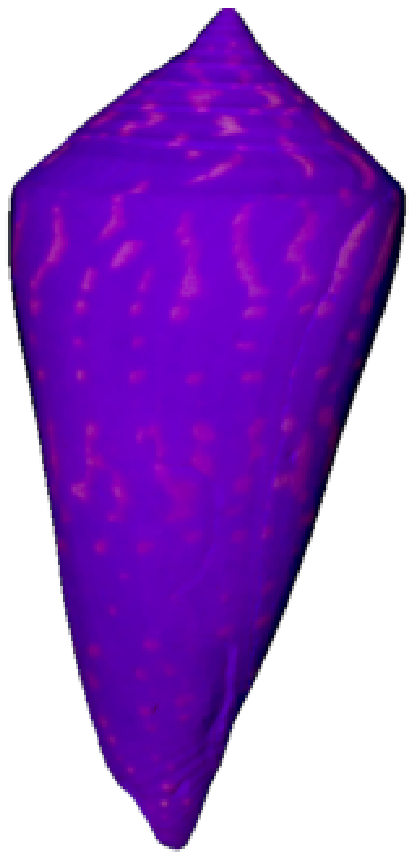}
&\includegraphics[width=0.2\linewidth]{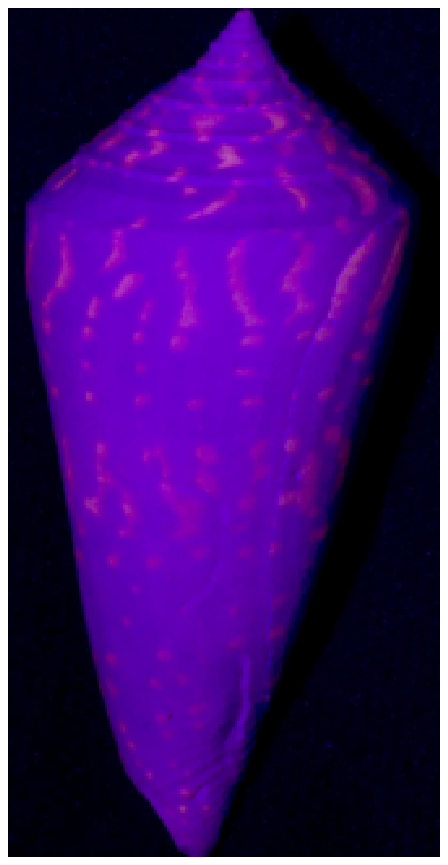}
\\

(a) & (b) & (c) & (d) \\

\end{tabular}
\caption[UV fluorescence texture mapping]{\hbox{3-D} model of a shell fossil acquired with our system. (a) \hbox{3-D} visible texture, (b) low-resolution \hbox{3-D} pattern texture, (c) high-resolution \hbox{3-D} bispectral texture. (d) high-resolution bispectral map. \emph{Conus delesertti} fossil, UF117269, FLMNH.}
\label{fig:patternextract}
\end{figure}

%% file: model.tex
\section{\hbox{3-D} Style Transfer Algorithm}
\label{sec:model}

We now present our \hbox{3-D} material style transfer algorithm. The inputs are $\mathbf{O}_{src}$ and $\mathbf{O}_{tar}$, the acquired source and target \hbox{3-D} meshes (Section~\ref{sec:measurementsystem}). We store per-vertex material properties for both meshes and appearance properties for $\mathbf{O}_{src}$. The goal is to generate the missing target appearance in a manner that adapts to its material structure without losing physical associations between materials and surface appearance. For convenience, a summary of symbols used in this section is presented in Table~\ref{tab:symbols}.

To begin, we conformally map~\cite{wang2005uniform,gu2004b} $\mathbf{O}_{src}$ and $\mathbf{O}_{tar}$ to a parameterized sphere to make the topologies easy to compare (See~\ref{sec:appendharmonicenergy}). Optional alignment~\cite{haker,gu2004b} using pre-defined user-specified landmarks improves later matching steps for cases where narrow elongated patches occur in tapered regions (see tip of eggshell). 

\input{fig-symbols}

\subsection{Patch Computation}

Our patch computation step uses existing processing methods to compute geometric features (Figure~\ref{fig:nomenclature}). Graph-cut segmentation~\cite{Felzenszwalb04} is applied to pre-filtered $I_f$ values which are first projected to $uv$ texture space from sphere space. This avoids special processing for patches that span two or more textures. We use bilateral filtering and anisotropic diffusion with white pixel thresholding.  Extracted components are given unique patch ids and the background patch is labeled. The segmentation map is mapped back to sphere space (Figure~\ref{fig:segmentation} a) and patch boundaries are computed using a half-edge data structure.

\label{sec:patchcomputation}

\input{fig-segmentation}

The material boundary sometimes extends beyond the patch boundary, intersecting a boundary face (Figure~\ref{fig:segmentation}b). This is expected when working with discrete triangles. We add neighboring vertices from the original mesh to extend the patch boundary. Blending functions create a smooth transition between foreground and background (during material assignment). Area $A$ is the total area of mesh faces in patch $\mathbf{p}$. Vertices store normal curvature (see supplemental) averaged over adjacent edges.

We generate property distribution functions (PDFs), linear combinations of orthonormal basis functions that describe how geometric, material and appearance properties change with respect to position over the surface of $\mathbf{p}$. For every source and target patch $\mathbf{p}_{src}$ and $\mathbf{p}_{tar}$, we apply a spherical harmonic transformation $f_{n}^m$ to generate the PDF $P(\rho)=\{\Pi_{\rho}^{i}|0 \leq i \leq n\}$ which is a set of coefficient vectors $\Pi_{\rho}^{i}$, one for each band $i$ of $f_{n}^m$ from a linear least squares fitting process. The set $\rho=\{\kappa, m, c, h, s\}$  for $\mathbf{p}_{src}$  and $\rho=\{\kappa, m, c \}$ for $\mathbf{p}_{tar}$ where $\kappa$ is curvature, $m$ is composition,  $c$ is concentration, $h$ is hue and $s$ is saturation. 

Spherical harmonic basis functions are well-suited for our purposes because they are defined on a unit sphere and are orthonormal for a given order. We only consider coefficients within the patch boundary and set positions outside $\mathbf{p}$ to $0$.  To achieve the fitting accuracy required for later processing steps, we consider the entire sphere evaluating all spherical directions. Setting $n = 150$ works well as discussed in Section~\ref{sec:discussion}.

\subsection{Property Distribution Mapping}
We aim to match each target patch to the most similar source patch considering all geometric and material properties: shape, area, curvature, composition and concentration. At the heart of our algorithm are property distribution maps (PDMs), unique bidirectional mappings that transform a PDF $P(\rho)$ over one surface to a PDF $P^\prime(\rho)$ on another. A PDM is a set matrices $Q(\rho)=\{\mathbf{T}_{\rho}^{i}|0 \leq i \leq n\}$ where $i$ is a spherical harmonic band and $\rho$ is a geometric or material property. These constructs are the mathematical links for adapting learned relationships in $\mathbf{p}_{src}$ to the structure of $\mathbf{p}_{tar}$. Similar concepts have been explored for spherical harmonic lighting in graphics.

Recall that the basis functions of a spherical harmonic band $i$ are orthonormal. Therefore, we can compute each $\mathbf{T}_{\rho}^{i}$ as a set of scaling and rotation matrices that transform the coefficient vector $\Pi^{i}_{\rho, src}$ to vector $\Pi^{i}_{\rho, tar}$ in a $m$-d space where $m$ is the number of spherical harmonic functions in band $i$: 

\begin{equation}
	\Pi^{i}_{\rho, tar} = \mathbf{T}^{i} \Pi^{i}_{\rho, src}
	\label{equ:transformeqn}
\end{equation}

To compute $\mathbf{T}_{\rho}^{i}$, we normalize $\Pi^{i}_{\rho, src}$ and $\Pi^{i}_{\rho, tar}$ to unit vectors $\hat{\Pi}^{i}_{\rho, src}$  $\hat{\Pi}^{i}_{\rho, tar}$ respectively, generating the corresponding scaling matrices $\mathbf{S}^{i}_{\rho, src}$ and  $\mathbf{S}^{i}_{\rho, tar}$ (Equation~\ref{equ:scalingmatrixsource}, Equation~\ref{equ:scalingmatrixtarget}) where $l^{i}_{\rho, src}$ and $l^{i}_{\rho, tar}$ are the lengths of $\Pi^{i}_{\rho, src}$ and $\Pi^{i}_{\rho, tar}$ respectively and $\mathbf{I}$ is an identity matrix. We use the Householder Reflection method~\cite{press2007} to compute a rotation matrix $\mathbf{R}^{i}_{\rho}$ that transforms $\hat{\Pi}^{i}_{\rho, src}$ to $\hat{\Pi}^{i}_{\rho, tar}$. Finally, $\mathbf{T}^{i}_{\rho}$ is computed as product of $\mathbf{R}^{i}_{\rho}$ with $\mathbf{S}^{i}_{\rho, src}$ and $\mathbf{S}^{i}_{\rho, tar}$ as shown in Equation~\ref{equ:transformmatrix}.

\begin{equation}
	\mathbf{S}^{i}_{\rho, src} = l^{i}_{\rho, src} \mathbf{I}
	\label{equ:scalingmatrixsource}
\end{equation}

\begin{equation}
	\mathbf{S}^{i}_{\rho, tar} = l^{i}_{\rho, tar} \mathbf{I}
	\label{equ:scalingmatrixtarget}
\end{equation}

\begin{equation}
	\mathbf{T}^{i}_{\rho} = \mathbf{S}^{i}_{\rho, tar} \mathbf{R}^{i}_{\rho} {\mathbf{S}^{i}_{\rho, src}}^{-1}
	\label{equ:transformmatrix}
\end{equation}

\subsection{Patch Matching}

We formulate patch matching as an energy minimization problem where the cost of matching $\mathbf{p}_{src}$ to $\mathbf{p}_{tar}$ is a weighted sum of the shape, area, curvature, composition, and concentration matching costs ($\mathbf{C}_{sh}$, $\mathbf{C}_{A}$, $\mathbf{C}_{\kappa}$, $\mathbf{C}_{m}$, $\mathbf{C}_{c}$ respectively): 

\begin{equation}
 \underset{\mathbf{p}_{src}}{\min} \, \mathbf{C}(\mathbf{p}_{tar},\mathbf{p}_{src}) = (\alpha \mathbf{C}_{sh} + \beta \mathbf{C}_{A} + \gamma \mathbf{C}_{\kappa} + \delta \mathbf{C}_{m} + \lambda \mathbf{C}_{c})(\mathbf{p}_{tar},\mathbf{p}_{src}) 
 \label{equ:costfunction_new}
 \end{equation}

We measure, $\mathbf{C}_{sh}$, the difference in the boundary contour of $\mathbf{p}_{src}$ and $\mathbf{p}_{tar}$,  as the differences in their average harmonic energies (See~\ref{sec:appendharmonicenergy}). Averages are computed over all edges.

Cost $\mathbf{C}_{A}$ measures the difference in patch areas. For $\mathbf{C}_{\kappa}$, $\mathbf{C}_{m}$ and $\mathbf{C}_{c}$, we use the determinant $\lvert \mathbf{T}^{i} \rvert$ of the linear transformation matrix $\mathbf{T}^{i}$ to measure how the distribution of the property changes at each level (spherical harmonic order) $i$ when mapping the PDF from $\mathbf{p}_{src}$ to $\mathbf{p}_{tar}$. Recall that the absolute determinant of a linear transformation matrix in a $m$-$d$ space measures the change in the scale of a $m$-$vector$, induced by applying the linear transformation to that vector. We minimize $\mathbf{C}$ for all combinations of $\mathbf{p}_{src}$ and $\mathbf{p}_{tar}$. 

The matching cost function for a property $\rho$, where $\rho \in \{\kappa, c, m\}$ is:

\begin{equation}
 \mathbf{C}_{\rho}(\mathbf{p}_{tar},\mathbf{p}_{src}) = \sum_{i = 0}^{n}\lvert 1 - \lvert \mathbf{T}_{\rho}^{i} \rvert \rvert
\label{equ:rhomatchingcost}
\end{equation}

\noindent The cost of mapping a $P(\rho)$ on $\mathbf{p}_{src}$ to PDF $P^{\prime}(\rho)$ over $\mathbf{p}_{tar}$ is the sum of the transformation costs at each spherical harmonic band $i$. The determinant of a transformation matrix of two functions with identical coefficients vectors is an identify matrix. We first compute the PDM $Q(\rho)$ and then compute the transformation cost at each band $i$ as the absolute difference between the determinant of the transformation matrix $\mathbf{T}_{\rho}^{i}$ and $1$ (the determinant of an identity matrix), where $\mathbf{T}_{\rho}^{i} \in Q(\rho), 0 \leq i \leq n$ and $n$ is the highest order of spherical harmonics used to compute the PDFs. The final mapping cost $\mathbf{C}_{\rho} $ is the sum of the transformation costs across all the bands. 

The user parameters $\alpha$, $\beta$, $\gamma$, $\delta$ and $\lambda$ control the influence of $\mathbf{C}_{sh}$, $\mathbf{C}_{A}$, $\mathbf{C}_{\kappa}$, $\mathbf{C}_{m}$, $\mathbf{C}_{c}$ respectively cover $\mathbf{C}$. These parameters can be adjusted for matching different pattern types. We use equal weights ($0.2$) when matching objects with similar pattern shape and color (Figure~\ref{fig:similar_shape_and_pattern_test} Test 1). If the source and target objects have similar pattern shape but different pattern colors (Figure~\ref{fig:similar_shape_and_pattern_test} Test 4), we decrease the contribution of composition to the overall cost ($\upalpha = 0.2$, $\upbeta = 0.2$, $\upgamma = 0.2$, $\updelta = 0.35$ and $\uplambda = 0.05$). For Figure~\ref{fig:teaser} we match patches using parameter values $\upalpha = 0.2$, $\upbeta = 0.2$, $\upgamma = 0.2$, $\updelta = 0.25$ and $\uplambda = 0.15$.

%\input{fig-spherespace}

%Figure~ref{fig:spherespace} shows a patch matching result.

\subsection{Material Assignment}
\input{fig-material_assn_algo}

Given matching patch pairs, we need to define a mapping that adapts  appearance in $\mathbf{p}_{src}$ to the material structure of $\mathbf{p}_{tar}$ so that we can compute functions for it's saturation and hue. We do this with a specialized PDM called a material map (MM). MMs are different because they map relationships between a material PDF and an appearance PDF on the same surface. We first compute material maps $\tau(c, s)$ and $\tau(m, h)$ on the $\mathbf{p}_{src}$ using the householder technique described previously. The results are sets of per-band transformation matrices $\tau(c, s)=\{\mathbf{T}^{i}_{c, s}, 0 \leq i \leq n\}$ and $\tau(m, h)=\{\mathbf{T}^{i}_{m, h}, 0 \leq i \leq n\}$ respectively. 

We now have a PDM that maps concentration on $\mathbf{p}_{src}$ to concentration on $\mathbf{p}_{tar}$, and a MM that maps the saturation to concentration on the source. To recover saturation on $\mathbf{p}_{tar}$, we compute a new PDM, PDM $Q(s)$ that maps the saturation on the source to the saturation on the target  while taking into account the change in the concentration distribution from the $\mathbf{p}_{src}$ to $\mathbf{p}_{tar}$. The PDM $Q(s)=\{\mathbf{T}^{i}_{s} | 0 \leq i \leq n\}$ can be computed using standard matrix operations as a per-band chain of matrix multiplications:

\begin{equation}
\mathbf{T}^{i}_{s} = \mathbf{T}^{i}_{c, s} \mu_{s}\sigma^{i}_{s}\mathbf{T}^{i}_{c} {\mathbf{T}^{i}_{c, s}}^{-1}
\label{equ:saturationtransform}
\end{equation}

\noindent The user parameter $\mu_{s}$ increases or decreases the concentration distribution of $\mathbf{p}_{tar}$. Per-band frequency scaling weights $\sigma^{i}_{s}$ control the variation in concentration over $\mathbf{p}_{tar}$:

\begin{equation}
\sigma^{i}_{s} = (n + 1) + (f_{s} \dot i)
\label{equ:saturationfreqweights}
\end{equation}

\noindent where $f_{s}$ is a user parameter that acts on each spherical harmonic band separately and can be used to simulate different levels of detail. Figure~\ref{fig:parameter_change} shows the effect of varying $\mu_{s}$ and $f_{s}$. 

Finally, for each band $i$, we apply $\mathbf{T}^{i}_{s}$ to $\Pi^{i}_{s, src}$ to compute the saturation coefficients $\Pi^{i}_{s, tar}$ for $\mathbf{p}_{tar}$:

\begin{equation}
\Pi^{i}_{s, tar} = \mathbf{T}^{i}_{s} \Pi^{i}_{s, src}
\label{equ:saturationcoeffs}
\end{equation}

\noindent We repeat the process, computing PDM $Q(h)=\{\mathbf{T}^{i}_{h} | 0 \leq i \leq n\}$  for hue to compute the set of transformation matrices $\mathbf{T}^{i}_{h}$ that transform hue on the $\mathbf{p}_{src}$ to  hue on $\mathbf{p}_{tar}$:

\begin{equation}
\mathbf{T}^{i}_{h} = \mathbf{T}^{i}_{h,m} \mu_{h}\mathbf{T}^{i}_{m} {\mathbf{T}^{i}_{h, m}}^{-1}
\label{equ:huetransform}
\end{equation}

\noindent where $\mu_{h}$ is a user parameter that controls the distribution of color pigments of materials over $\mathbf{p}_{tar}$. The parameter $\mu_{h}$ manipulates the proportion of different color generating compounds in the material. We apply $\mathbf{T}^{i}_{h}$ to $\Pi^{i}_{h, src}$ to compute the hue coefficients $\Pi^{i}_{h, tar}$ on the target patch:

\begin{equation}
\Pi^{i}_{h, tar} = \mathbf{T}^{i}_{h} \Pi^{i}_{h, src}
\label{equ:huecoeffs}
\end{equation}

\noindent We repeat the process for the background patch and combine the foreground and background patches into a single mesh. 

A \hbox{3-D} gaussian weighting function blends the saturation and hue in a neighborhood of boundary vertices to make a smooth color transition between the foreground and background materials. The final vertex color combines the reconstructed hue and saturation with the value component from $\mathbf{R}_{bis}$ (in HSV color space) for $\mathbf{p}_{tar}$ which we convert to $RGB$ color space. Inverse spherical harmonic mapping is applied to the reconstructed spherical mesh to generate the final result.

\input{fig-parameter_change}

%% file: fig-symbols.tex
\begin{table}[h]
\setlength\tabcolsep{1.5pt}
\begin{tabular}{p{0.15\linewidth}|p{0.85\linewidth}}
\hline
\small Symbol & \small Meaning \\
\hline
\small $f_{n}^m$ & \small spherical harmonic transformation \\
\small $P(\rho)$ &  \small property distribution function \\ 
\small $\rho$ & \small a set of properties \\
\small $I_f$ & \small fluorescent emission intensity \\
\small $\mathbf{R}_{bis}$ & \small bispectral reflectance map \\
\small $c$ & \small material concentration \\
\small $m$ & \small material composition \\
\small $h$ & \small hue in hsv color space \\
\small $s$ & \small saturation in hsv color space \\
\hline
\end{tabular}
\caption{\label{tab:symbols}%
Summary of symbols.}
\end{table}

%% file: fig-segmentation.tex
\begin{figure}[h!]
\centering
\includegraphics[width=1.0\linewidth]{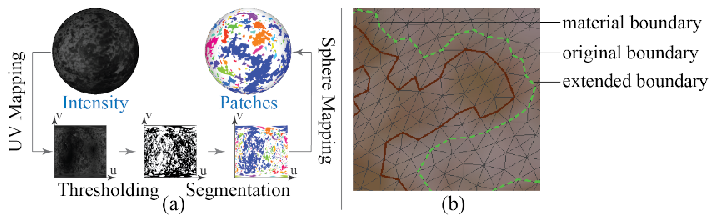}
\caption{\label{fig:segmentation}
(a) Pattern segmentation process. (b) Resolving differences between patch and material boundaries.}
\end{figure}

%% file: fig-material_assn_algo.tex
\begin{figure}[h]
\centering
\includegraphics[width=1.0\linewidth]{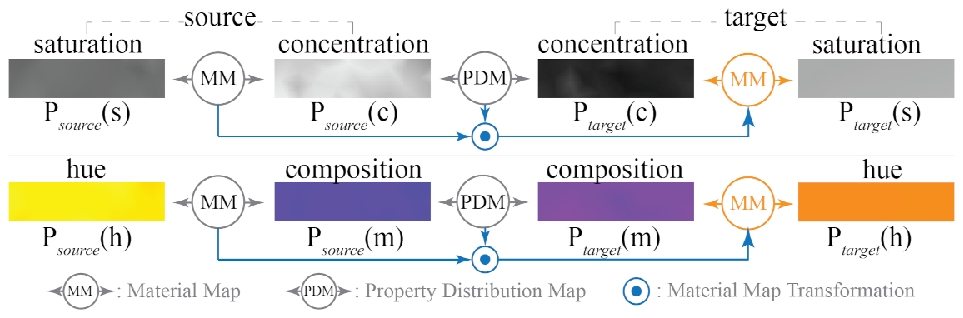}
\caption{\label{fig:material_assn_algo}%
Material assignment process.}
\end{figure}

%% file: fig-parameter_change.tex
\begin{figure}[h]
\centering
\includegraphics[width=0.85\linewidth]{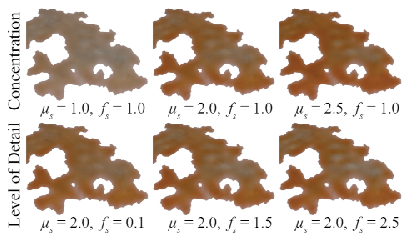}
\setlength{\belowcaptionskip}{-10pt} 
\caption{\label{fig:parameter_change}%
Simulating different levels of material concentration and detail by varying (top) scale parameter $\mu_{s}$ and (bottom) frequency parameter $f_{s}$.}
\end{figure}

%% file: results.tex
\input{fig-similar_shape_and_pattern_test_v002}

\input{fig-alleggs}

\section{Experimental Results}
\label{sec:results}

To validate the efficacy of our approach, we compare reconstructed hue and saturation with per-vertex ground truth values from our measurement system. One goal is to determine whether we can reconstruct appearance for objects with different shapes, scales, and color pattern variations from a single source exemplar. Qualitative evaluation from a user study assesses whether the reconstruction is a plausible replica. We also investigate algorithm performance on a complex composite material that has fluorescent and non-fluorescent components. Finally, we compare our results with prior work.

Recall that we do not use prior knowledge of appearance for $\mathbf{O}_{target}$ in the algorithm. For each dataset, we generate ground truth models by adding appearance information from our measurement process to $\mathbf{O}_{target}$, including per-vertex hue and saturation. We do not have ground truth for colorless datasets.

\subsection{Different Shape and Patterns from a Single Exemplar}

Tests $1$-$6$ evaluate color recovery when the object shape, object scale, pattern shape and pattern color are varied. Our dataset consists of $16$ avion eggshells: $13$ Coturnix quail, $1$ Valley Quail, $1$ Gambles quail, and $1$ peahen. For each experiment, we used the same source, a Coturnix quail eggshell (Figure~\ref{fig:teaser}a), but different targets to meet the experiment criteria (listed below). We did not incorporate any appearance information from the target, only material measurements. We then compared the reconstructed saturation and hue of the target with the ground truth. Our test criteria include:

\begin{enumerate}
  \setlength{\itemindent}{-0.7em}
\item[]Test 1: \textbf{similar} shape, scale, patterns and color.
\item[]Test 2: \textbf{similar} shape, scale, and color, \textbf{different} patterns.
\item[]Test 3: \textbf{similar} shape and scale, \textbf{different} patterns and color.
\item[]Test 4: \textbf{similar} patterns and color, \textbf{different} shape and scale.
\item[]Test 5: \textbf{different} shape, scale, patterns and  color.
\item[]Test 6: Varying shape, scale, patterns and color.
\end{enumerate}

\paragraph{\textbf{Qualitative Evaluation}} Figure~\ref{fig:similar_shape_and_pattern_test} shows results for Tests $1$-$5$ including material and appearance measurements. Figure~\ref{fig:all_eggs} shows results for Test $6$ (see supplemental for material measurements). Each vertex stores one set of ground truth measurements. Each eggshell had on average $177,231$ vertices. There were $2,835,700$ total vertices (ground truth samples).

Although close to the original our results were not perfect. Among our first five tests, our algorithm performed better for Tests $4$ and $5$. In both examples, the ground truth shows smooth distributions of hue and saturation within patch boundaries which we replicate well (See Figure~\ref{fig:similar_shape_and_pattern_test} Test $5$ GT and reconstructed). Complex pattern shapes and sharp changes in materials were harder to reconstruct. Test $3$ shows an example of inconsistencies we found due to over smoothing of saturation values. Test $2$ shows specular artifacts (Figure~\ref{fig:similar_shape_and_pattern_test} Test $2$ input) that produce inaccurate segmentations and reconstruction errors (See Section~\ref{sec:discussion}). Interestingly, the perceived orangish color in some hue maps suggests a high numbers of foreground patches (Figure~\ref{fig:similar_shape_and_pattern_test} Test $3$ GT hue compared to Figure~\ref{fig:similar_shape_and_pattern_test} Test $1$ GT hue).

\paragraph{\textbf{Quantitative Evaluation}} In Figure~\ref{fig:eval_graph} we computed the per-vertex absolute error between reconstructed hue and saturation and ground truth values, averaged over the number of vertices in the target for Tests $1$-$5$ and averaged over all vertices for Test $6$. We compute saturation with an $85\%$ accuracy and hue with a $92\%$ accuracy (Test 6). Lower saturation accuracy is due to greater variation in saturation values within a patch compared to hue (See Figure~\ref{fig:similar_shape_and_pattern_test} Test 3 GT). Over smoothing of high frequency saturation (Figure~\ref{fig:similar_shape_and_pattern_test} Test $3$ reconstructed) resulted in higher errors (See Section~\ref{sec:discussion}).

\input{fig-eval_graph}

\paragraph{\textbf{User Study}} Perceptual metrics are important for evaluating style transfer methods~\cite{nguyen2012}. We gathered qualitative metrics from a user study with $19$ participants. Most participants were novices, graduate students and professional colleagues randomly selected with no expertise in biology. Our goal is to evaluate a lay-person's perception. Only two participants worked in bio-related fields. Participants were randomly selected for one of two tasks to be completed remotely. No participants participated in both tasks. In both experiments eggshells (or eggshell pairs) were presented one at a time on a computer screen in a random order. Depending on the study task, participants were required to make a selection of \emph{real} or \emph{simulated}, or enter ratings from $1-10$ before pressing \emph{Next} to continue. Participants were not allowed to go back to make changes.

\noindent \emph{Task $1$: Real or Simulated.} One way to evaluate the results of Tests $1$-$5$ is to determine plausibility of the replica. We showed nine participants digital images of $16$ eggshells, half of which were real and half of which were simulated. We included the reconstruction results from Test $1$-$5$ to obtain additional perceptual data for our test criteria,  but otherwise chose simulated eggs randomly from the set (~Figure~\ref{fig:all_eggs}). Participants were told to examine the \hbox{3-D} model and select \emph{real} if the model was likely a \hbox{3-D} scan of a real object, or \emph{simulated} if it was likely generated by a computer algorithm.

\noindent \emph{Task $2$: Rate the Simulation.} Another question we wanted to answer was how visually similar the new texture was to the original. We showed another group of ten participants digital images of $16$ eggshell pairs clearly labeling the reconstruction and ground truth. Participants were asked to rate the quality of the reconstruction on a scale from $1$ (no similarity) to $10$ (identical) in three areas: \emph{color}, \emph{pattern} and \emph{overall}. Afterward, participants commented (in a text box) on what made the objects appear real, and what made the objects appear simulated.

Overall, $50\%$ of the eggs were identified correctly as real or simulated. More importantly, $48.61\%$ of the simulated eggshells were mistaken for real eggshells confirming our ability to make plausible replicas (Figure~\ref{fig:userstudy} \emph{left}). Only slightly more simulated eggshells were detected $51.39\%$. The average ratings per category were similar: $8.5$, $8.08$, $8.23$ for color, pattern and overall respectively (Figure~\ref{fig:userstudy} \emph{right}).

We gained insight by combining results from both tasks for eggshells from Tests $1$-$5$ (Figure~\ref{fig:userstudy_results}). We will refer to these eggshells by test number. Test $1$ was the most plausible, incorrectly identified as real by $77.78\%$ of the participants. Cross-checking with task two shows it had the highest overall rating ($8.96$) and highest pattern ratting ($9.6$). Test $4$ placed second, mistaken for real by $66.67\%$ of participants. It had the highest color rating ($9.2$). All participants recognized Test $2$ as simulated due to errors from specular artifacts. Test $2$, $3$ and $5$ had the lowest pattern ratings ($6.95$, $6.9$ and $4.9$ respectively) making them identifiable simulations (by $100\%$, $55.56\%$ and $66.67\%$ of participants respectively). Nine  ($9$) out of $10$ participants indicated that \textbf{accuracy of pattern shape and boundaries} was the most important factor for realism. The presence of realistic shading (lighting) effects drew mixed responses favoring both realism and simulation. Low image resolution and pixelation artifacts were the main indicators of simulated data (see comment summary in supplemental).

\input{fig-userstudy}
\input{fig-userstudy_results}
\input{fig-experiment_results}
\input{fig-comparison}

\input{fig-mineral}

\subsection{Evaluating Fluorescent and Non-Fluorescent Materials}
\label{sec:nonfluorescent}

\paragraph{Qualitative Evaluation} Next we evaluate our algorithm on a complex material. The source in Figure~\ref{fig:mineral} is composed of three materials: willemite (brick red), calcite (white) franklinite (black). The willemite fluoresces green and the calcite fluoresces red. Franklinite is non-fluorescent. This presents an interesting case as the target is almost entirely willemite and franklinite (see Figure~\ref{fig:mineral} GT). Brick red willemite dominates the reconstruction and it's ground truth. This confirms our patch matching algorithm pairs source and target patches with the most similar composition and concentration PDFs. Compared to the diffuse shading effects for franklinite in the ground truth, the reconstruction is a flat black in non-fluorescent regions as they are not detected. We successfully map across different topologies, adapting the exemplar pattern to the target structure.

\paragraph{Quantitative Evaluation} Using error metrics from Tests $1$-$5$, we reconstruct the hue of the target mineral sample with $82\%$ accuracy and saturation with $79\%$ accuracy. The low value for saturation is due to over smoothing of high frequency saturation values. There were $145,380$ vertices in our mineral sample. Results are summarized in Table~\ref{fig:experiment_results}.

\subsection{Comparison with Prior Work}
In this test, the source is a modern descendent of a colorless fossil target. We compare our approach (Figure~\ref{fig:comparison}a) to two mapping methods: texture mapping with \emph{Geomagic} (Figure~\ref{fig:comparison}c and M1 closeup), and conformal surface parameterization (Figure~\ref{fig:comparison}d and M2 closeup). Close comparison with the $\mathbf{R}_{bis}$ (contrast enhanced for this illustration) in Figure~\ref{fig:comparison}a  shows that other methods retain the pattern structure of the source. Our results change the shape, frequency, and color of patches on the surface to conform to target materials even when there is no color information on the target. We combine curvature matching with bidirectional property maps to match and reconstruct how the material concentration distribution changes with respect to shape. The results capture a randomized effect, as color changes with respect to saturation. We discuss this dataset and application further in Section~\ref{sec:applications}.

%% file: fig-similar_shape_and_pattern_test_v002.tex
\begin{figure*}[p]
\centering
\renewcommand{\arraystretch}{.75}
\begin{tabular}{@{\hspace{0.5\tabcolsep}} c @{\hspace{0.5\tabcolsep}} c @{\hspace{0.5\tabcolsep}} c @{\hspace{0.5\tabcolsep}} c @{\hspace{0.5\tabcolsep}} c @{\hspace{0.5\tabcolsep}} c @{\hspace{0.25\tabcolsep}} c @{\hspace{0.5\tabcolsep}} c @{\hspace{0.5\tabcolsep}} c @{\hspace{0.5\tabcolsep}} c @{\hspace{0.5\tabcolsep}} c }
    \multicolumn{11}{@{\hspace{0.5\tabcolsep}} l}{\footnotesize Test 1: \textbf{similar} shape, scale, patterns and color} \\
	
	\multirow{2}{*}{reconstructed} & \multirow{2}{*}{ GT} & & & & & & \multicolumn{2}{@{\hspace{0.5\tabcolsep}} c}{\footnotesize material properties} & \multicolumn{2}{@{\hspace{0.5\tabcolsep}} c}{\footnotesize appearance} \\
	 
	 & & & \footnotesize hue & \footnotesize saturation & & & \footnotesize concentration & \footnotesize composition & & \\
     
     %& & & & & & & \scriptsize (log scale) & & & \\
     
     \multirow{2}{*}[0.07\hsize]{\includegraphics[width=0.12\hsize]{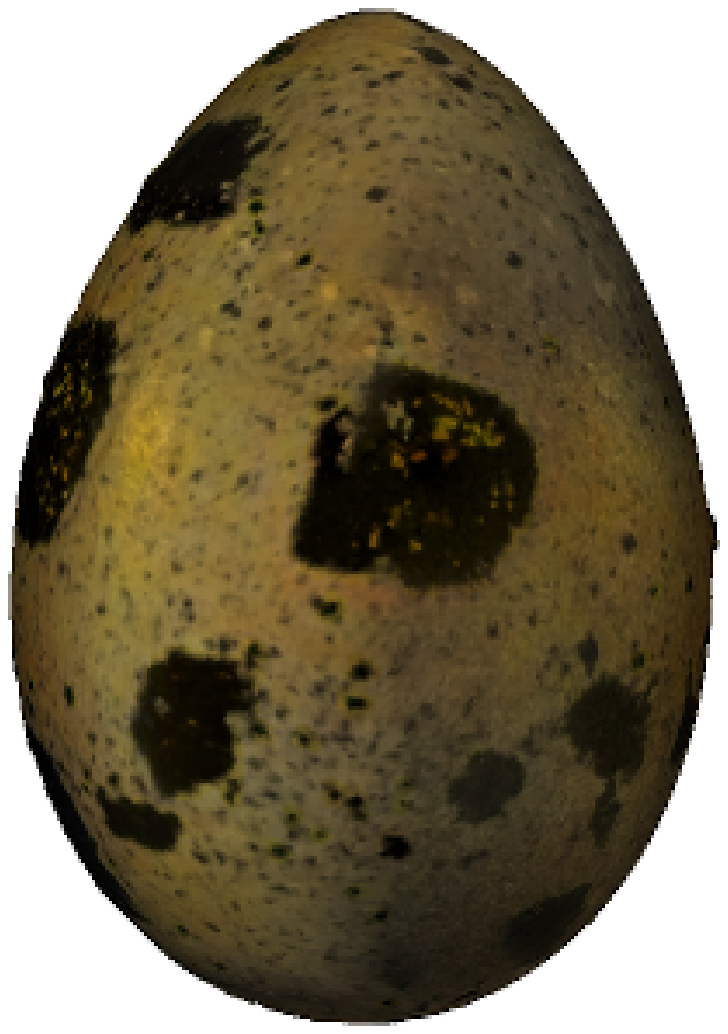}} & \multirow{2}{*}[0.07\hsize]{\includegraphics[width=0.12\hsize]{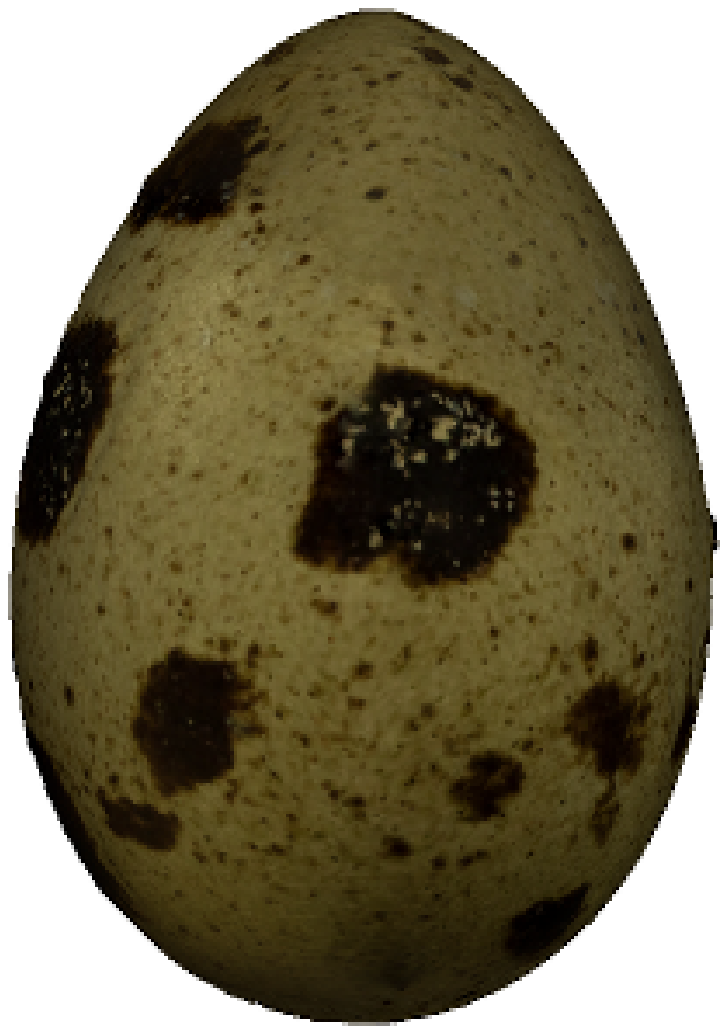}} & 
     \raisebox{1.25\normalbaselineskip}[0pt][0pt]{\rotatebox[origin=c]{90}{\footnotesize reconstructed}} &
     \includegraphics[width=0.06\hsize]{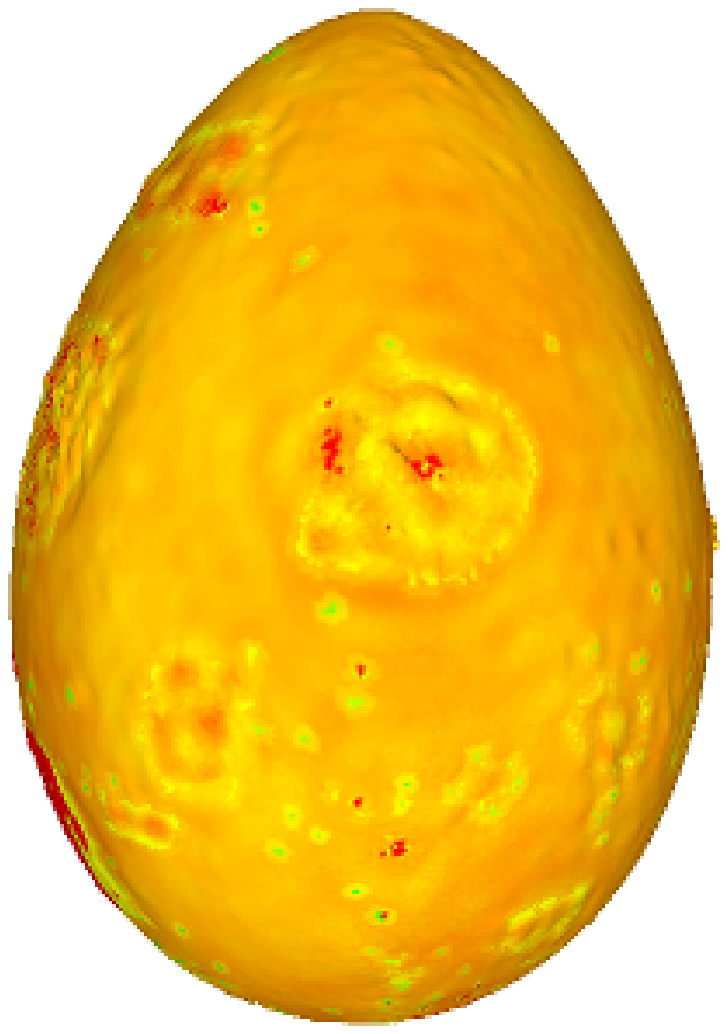} &
     \includegraphics[width=0.06\hsize]{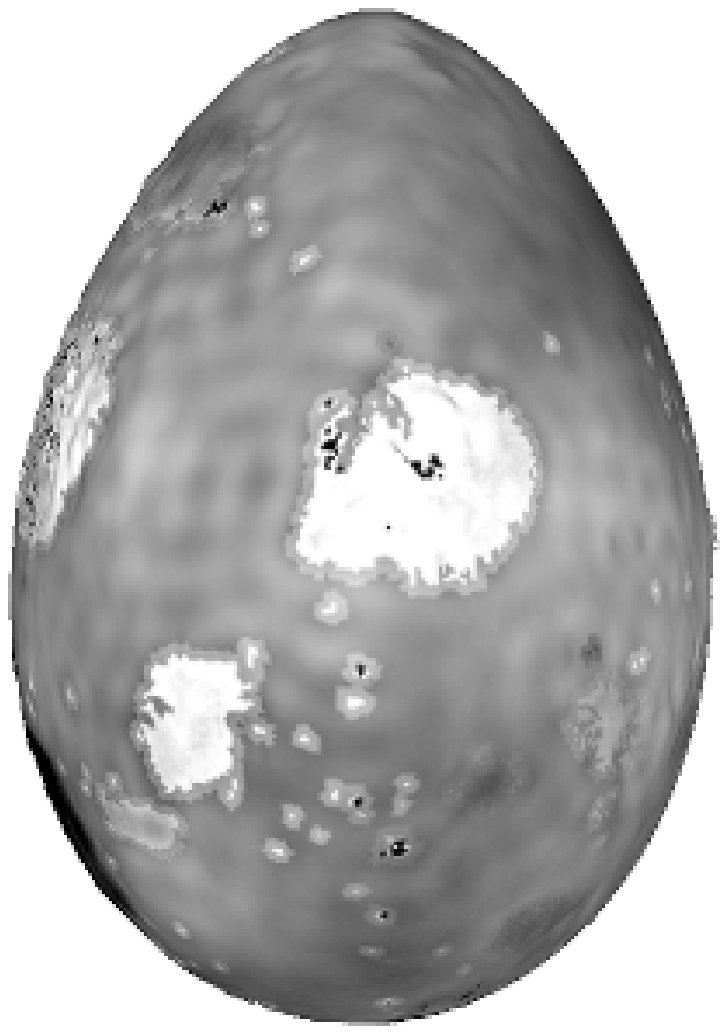} &
     \multirow{2}{*}{\rotatebox[origin=c]{90}{\footnotesize input}} & 
	 \raisebox{1.25\normalbaselineskip}[0pt][0pt]{\rotatebox[origin=c]{90}{\footnotesize target}} &
     \includegraphics[width=0.06\hsize]{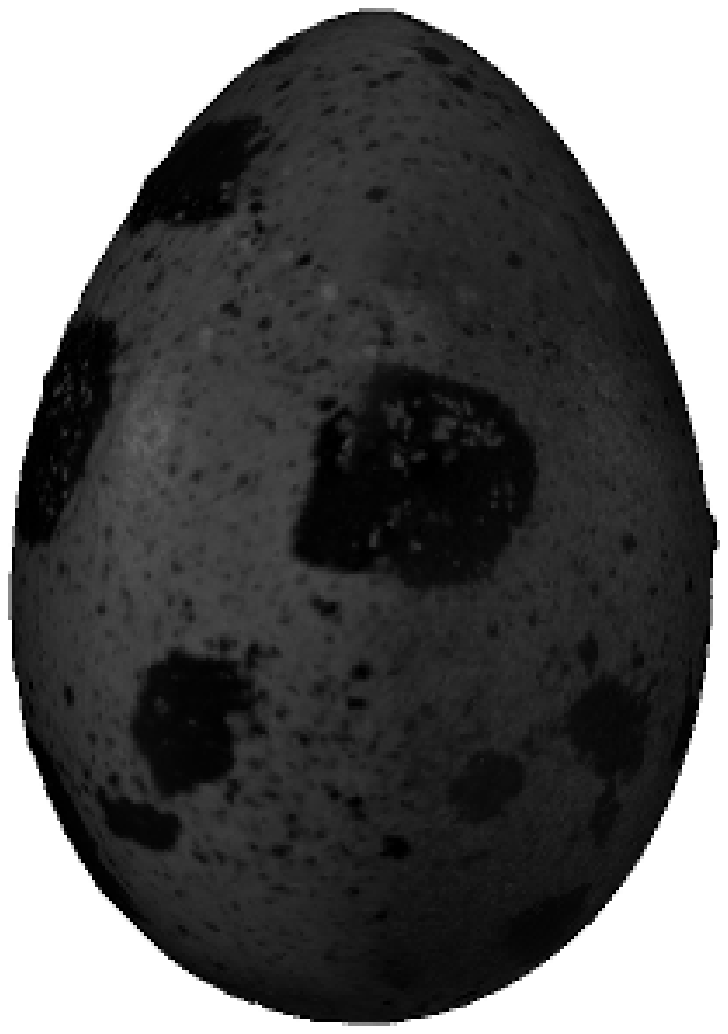} &
     \includegraphics[width=0.06\hsize]{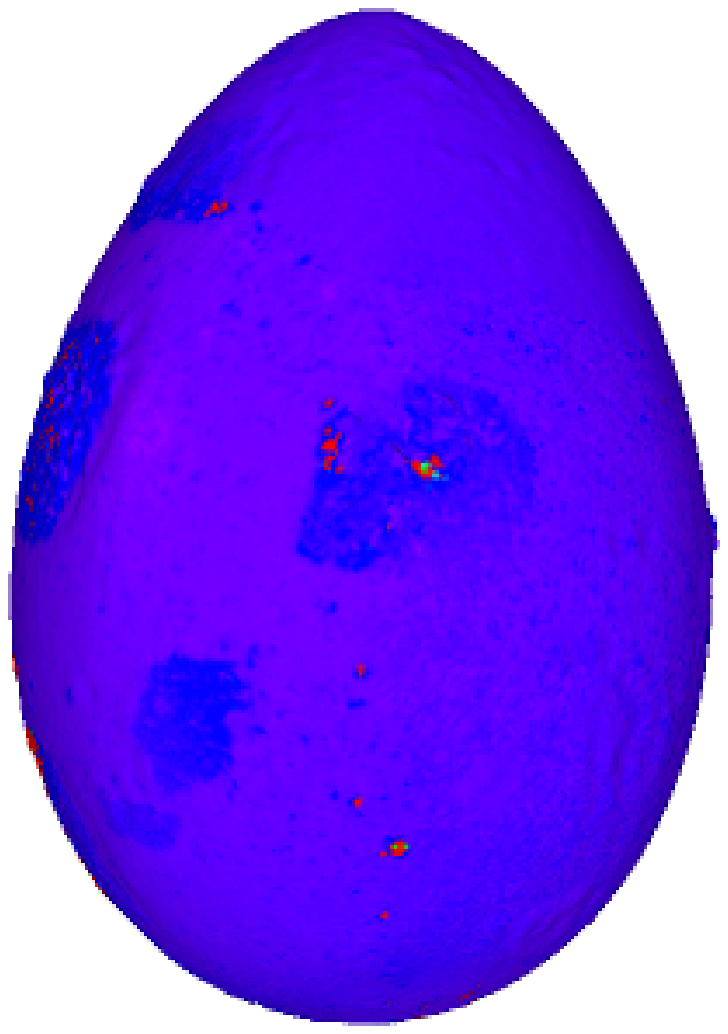} & 
     & \\
     
     & & \raisebox{1.25\normalbaselineskip}[0pt][0pt]{\rotatebox[origin=c]{90}{\footnotesize GT}} & 
     \includegraphics[width=0.06\hsize]{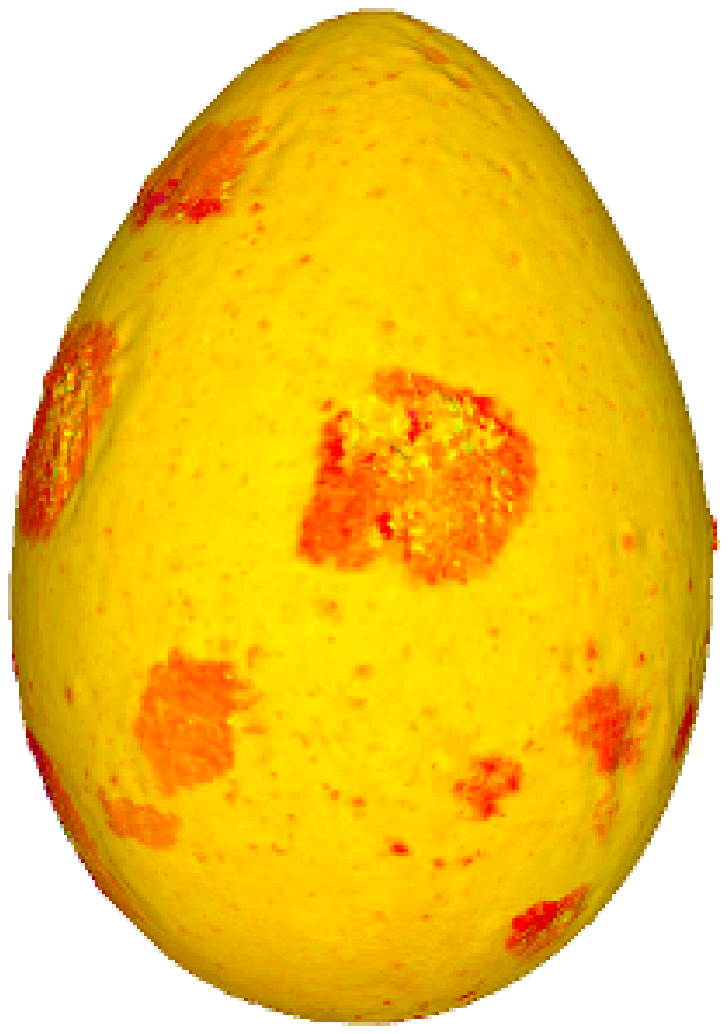} &
     \includegraphics[width=0.06\hsize]{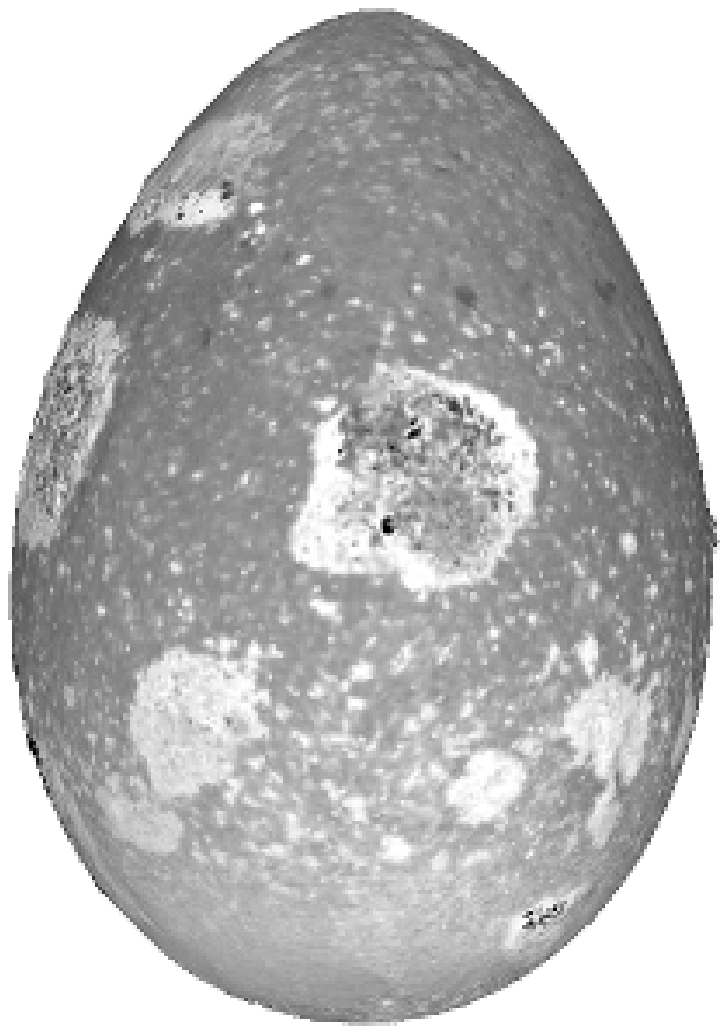} & &
     \raisebox{1.25\normalbaselineskip}[0pt][0pt]{\rotatebox[origin=c]{90}{\footnotesize source}} & 
     \includegraphics[width=0.06\hsize]{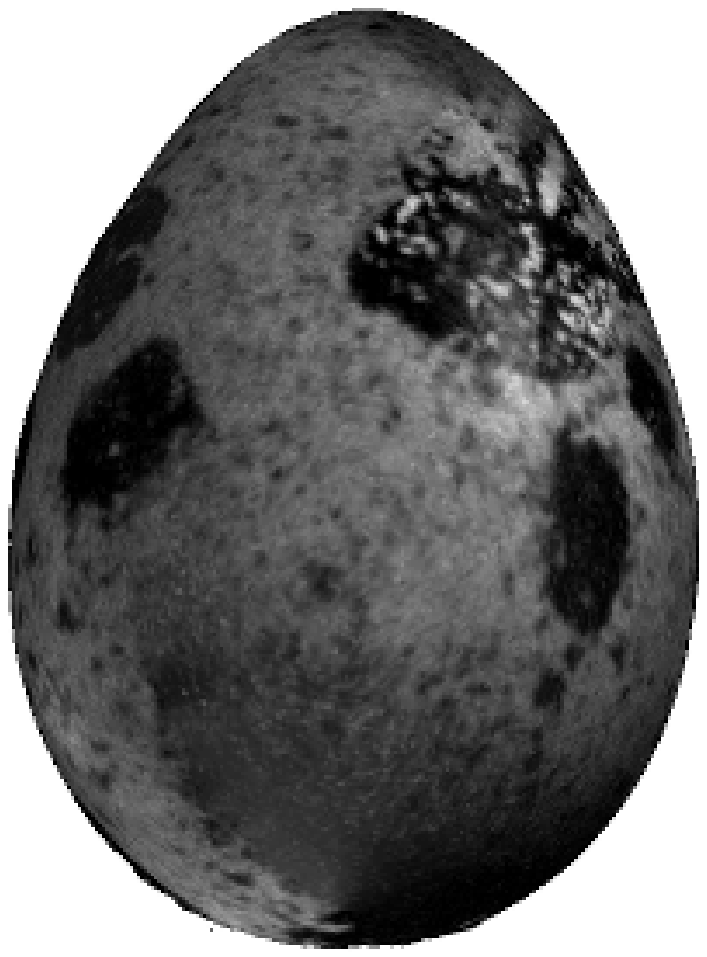} &
     \includegraphics[width=0.06\hsize]{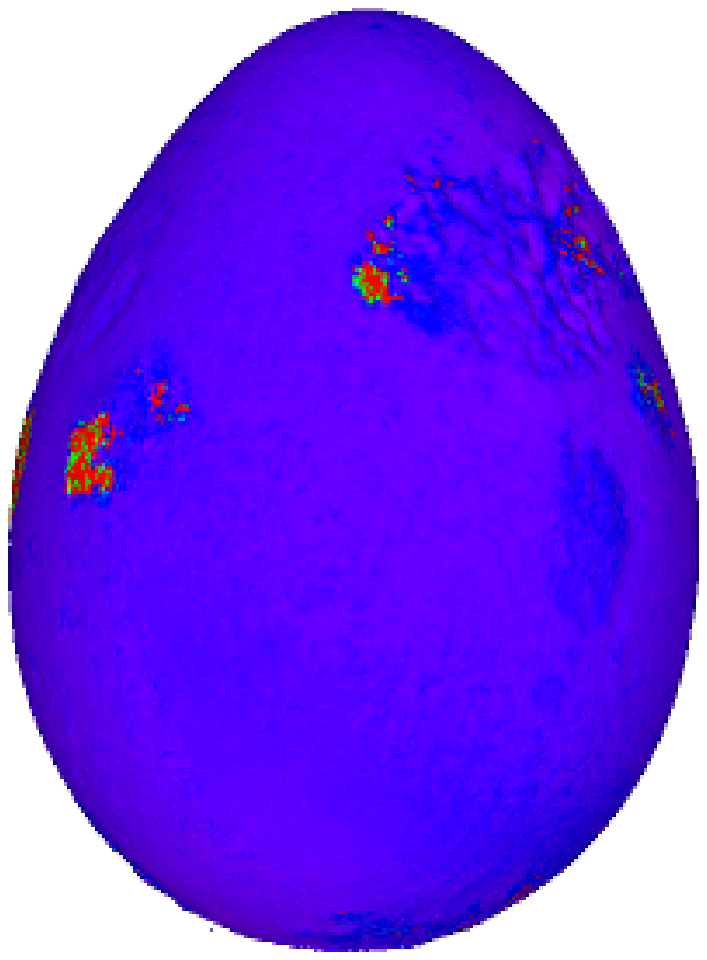} & 
     \includegraphics[width=0.06\hsize]{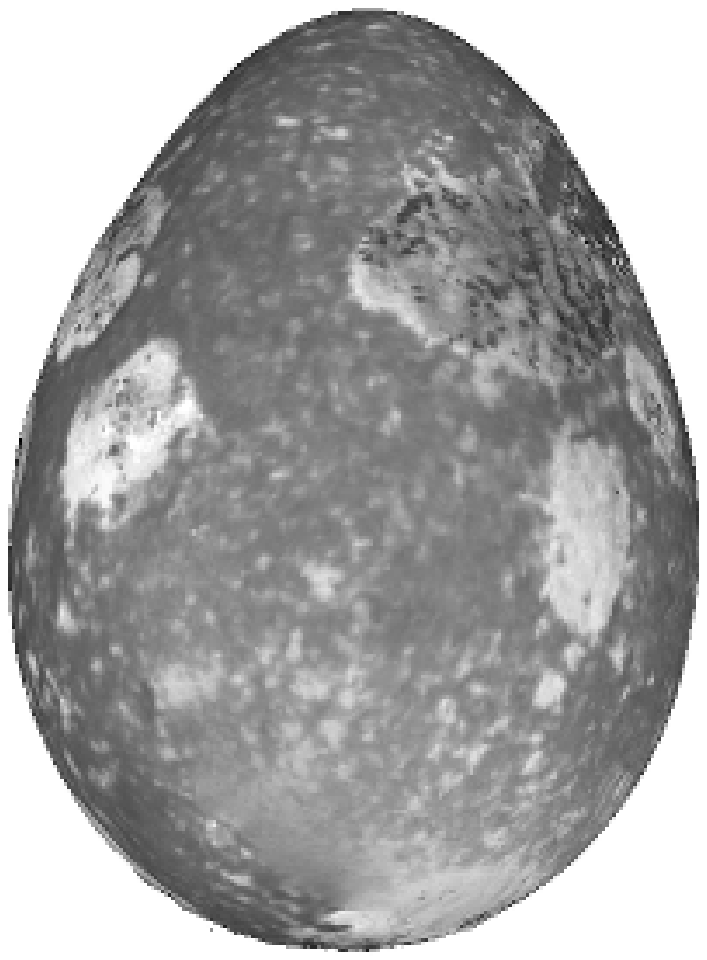} &
     \includegraphics[width=0.06\hsize]{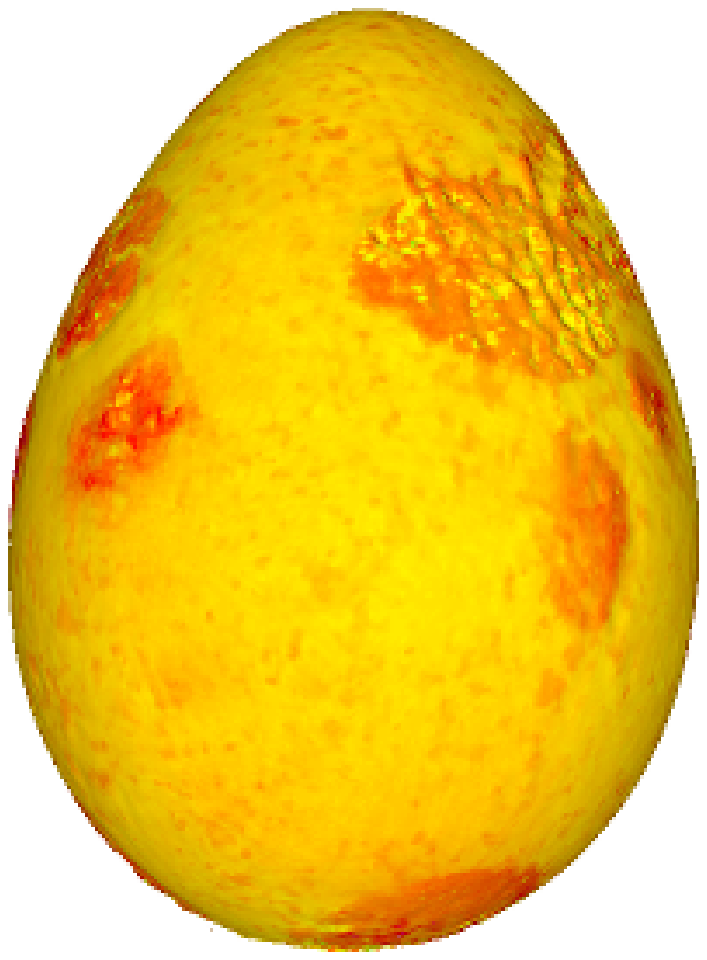} \\
     
     \multicolumn{2}{c}{\scriptsize Coturnix Quail} & & & & & & \footnotesize intensity & \footnotesize hue & \footnotesize saturation & \footnotesize hue \\
     
     & & & & & & & \multicolumn{2}{c}{\scriptsize UV (365nm)} & \multicolumn{2}{c}{\scriptsize visible (400nm-700nm)} \\
\end{tabular}

%\vspace{1mm}

\begin{tabular}{@{\hspace{0.5\tabcolsep}} c @{\hspace{0.5\tabcolsep}} c @{\hspace{0.5\tabcolsep}} c @{\hspace{0.5\tabcolsep}} c @{\hspace{0.5\tabcolsep}} c @{\hspace{0.5\tabcolsep}} c @{\hspace{0.25\tabcolsep}} c @{\hspace{0.5\tabcolsep}} c @{\hspace{0.5\tabcolsep}} c @{\hspace{0.5\tabcolsep}} c @{\hspace{0.5\tabcolsep}} c }
    \multicolumn{11}{@{\hspace{0.5\tabcolsep}} l}{\footnotesize Test 2: \textbf{similar} shape, scale, and color, \textbf{different} patterns} \\
	
	\multirow{2}{*}{reconstructed} & \multirow{2}{*}{GT} & & & & & & \multicolumn{2}{@{\hspace{0.5\tabcolsep}} c}{\footnotesize material properties} & \multicolumn{2}{@{\hspace{0.5\tabcolsep}} c}{\footnotesize appearance} \\
	 
	 & & & \footnotesize hue & \footnotesize saturation & & & \footnotesize concentration & \footnotesize composition & & \\
     
     %& & & & & & & \scriptsize (log scale) & & & \\
     
     \multirow{2}{*}[0.07\hsize]{\includegraphics[width=0.12\hsize]{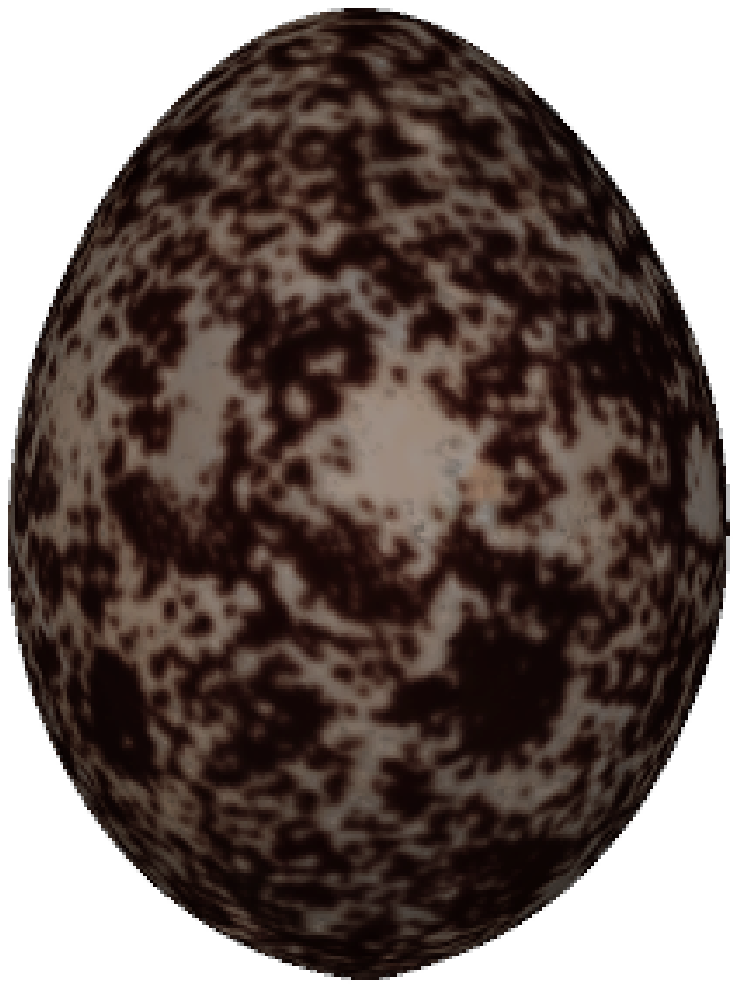}} & \multirow{2}{*}[0.07\hsize]{\includegraphics[width=0.12\hsize]{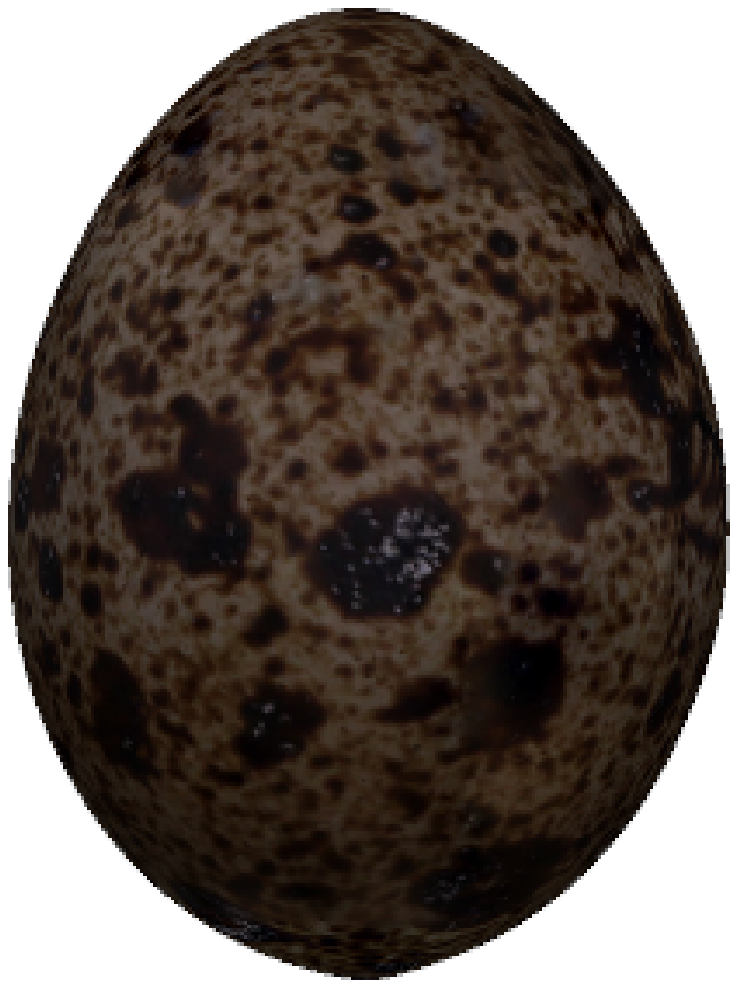}} & 
     \raisebox{1.25\normalbaselineskip}[0pt][0pt]{\rotatebox[origin=c]{90}{\footnotesize reconstructed}} &
     \includegraphics[width=0.06\hsize]{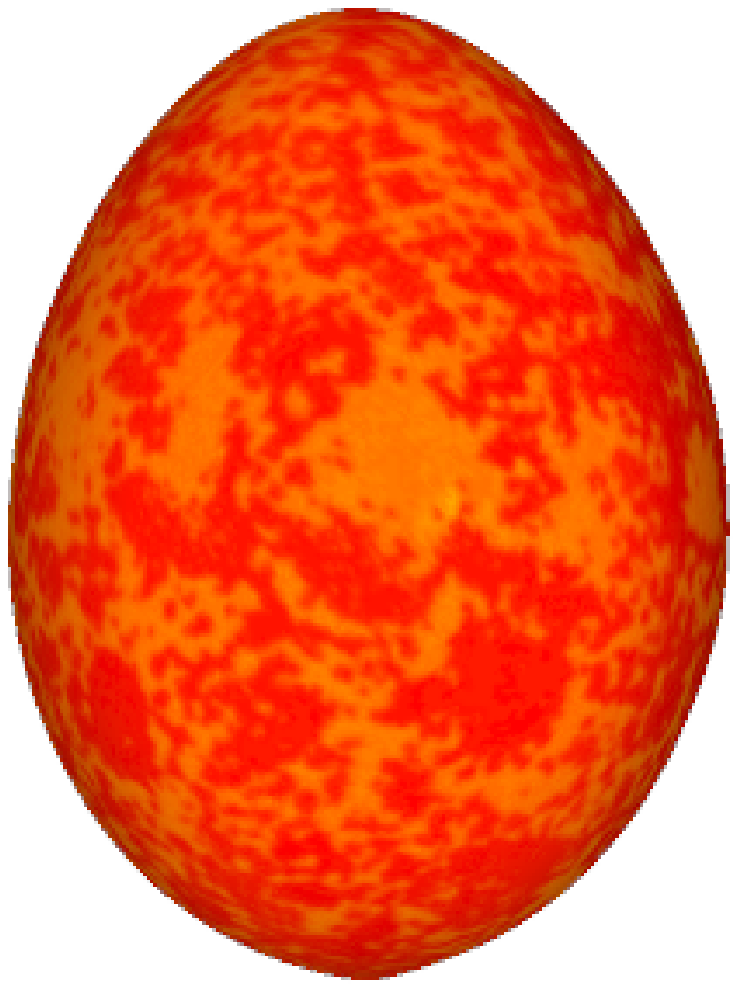} &
     \includegraphics[width=0.06\hsize]{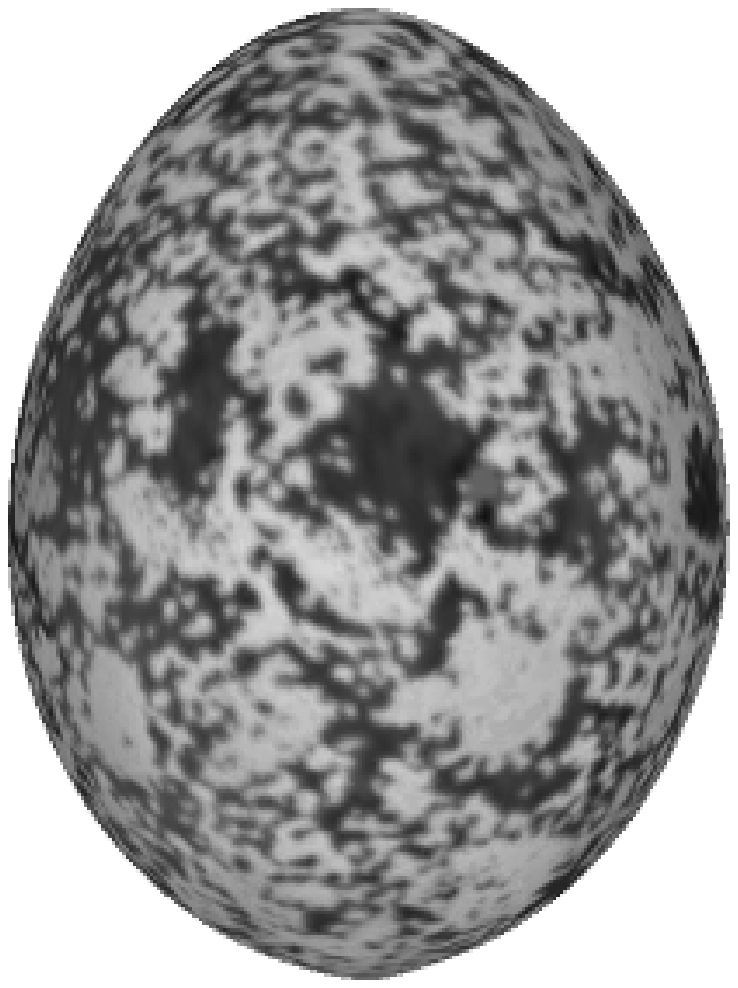} &
     \multirow{2}{*}{\rotatebox[origin=c]{90}{\footnotesize input}} & 
	 \raisebox{1.25\normalbaselineskip}[0pt][0pt]{\rotatebox[origin=c]{90}{\footnotesize target}} &
     \includegraphics[width=0.06\hsize]{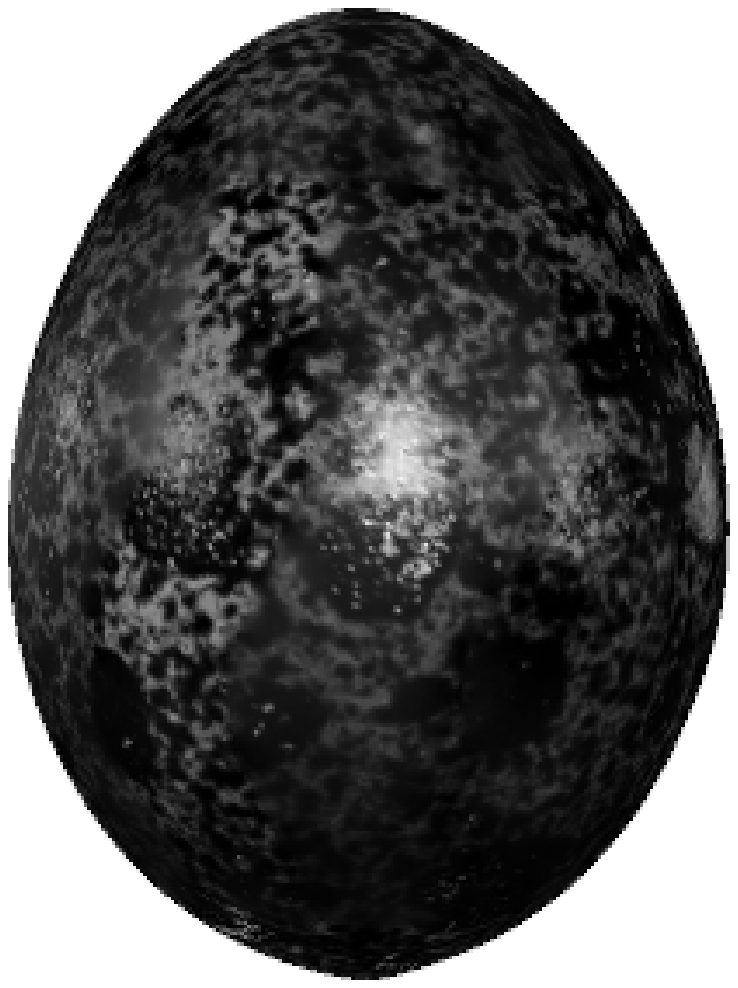} &
     \includegraphics[width=0.06\hsize]{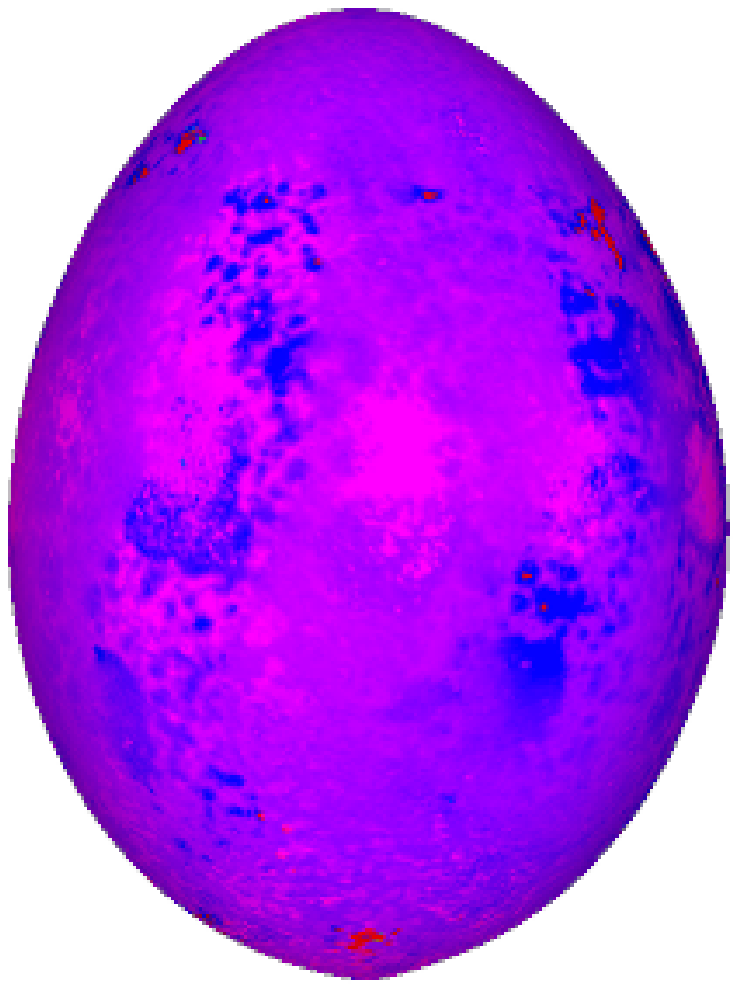} & 
     & \\
     
     & & \raisebox{1.25\normalbaselineskip}[0pt][0pt]{\rotatebox[origin=c]{90}{\footnotesize GT}} & 
     \includegraphics[width=0.06\hsize]{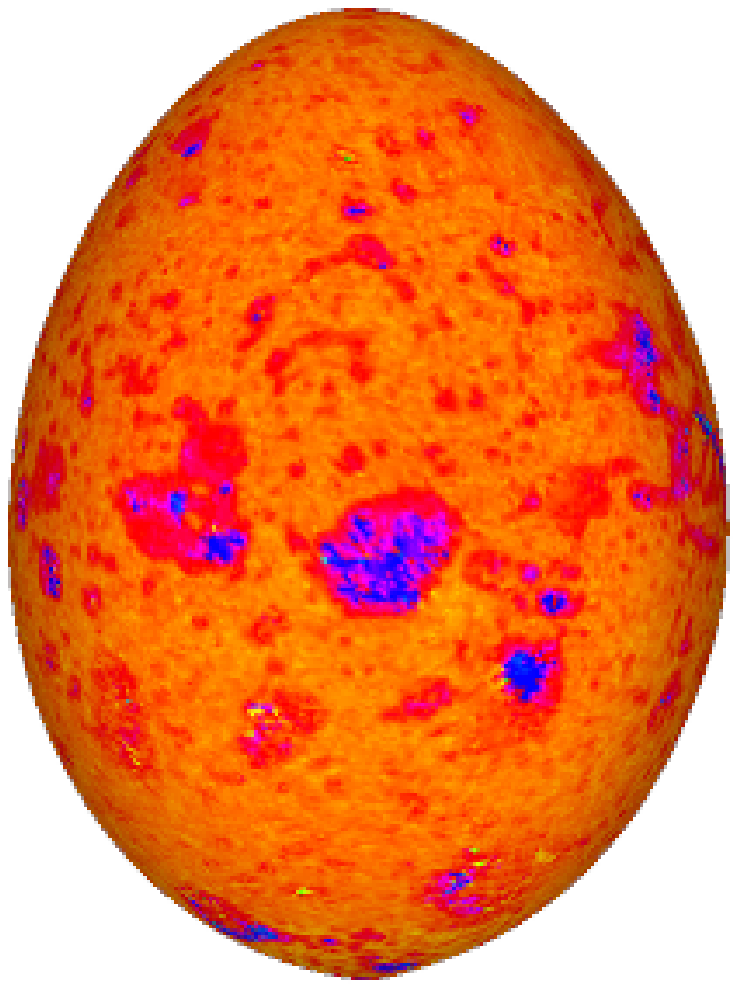} &
     \includegraphics[width=0.06\hsize]{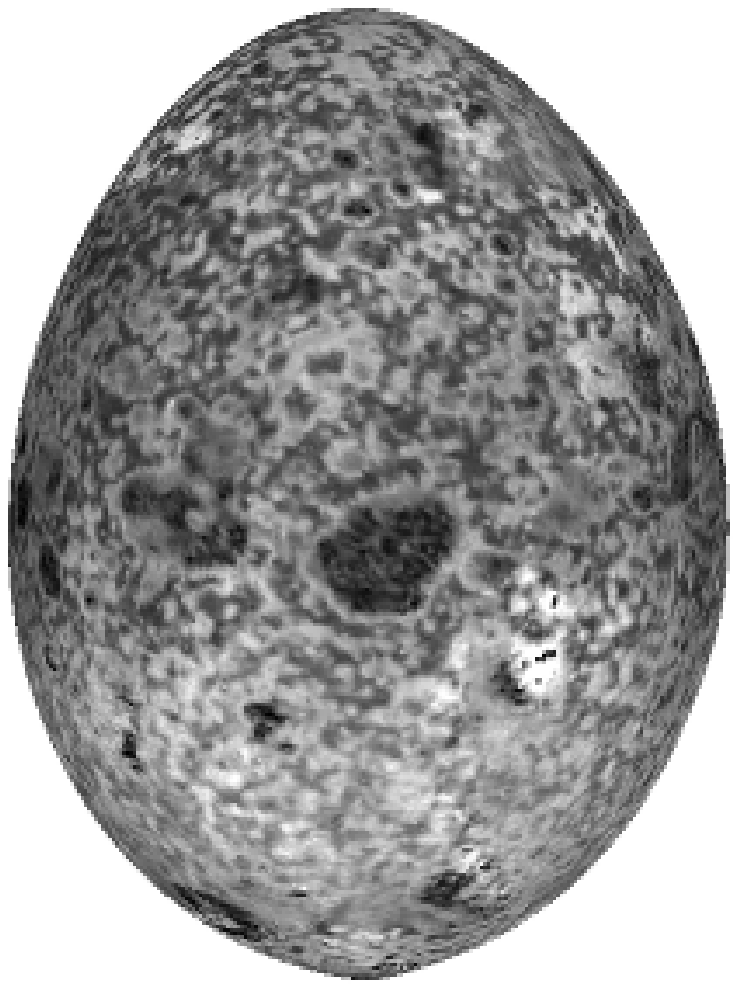} & &
     \raisebox{1.25\normalbaselineskip}[0pt][0pt]{\rotatebox[origin=c]{90}{\footnotesize source}} & 
     \includegraphics[width=0.06\hsize]{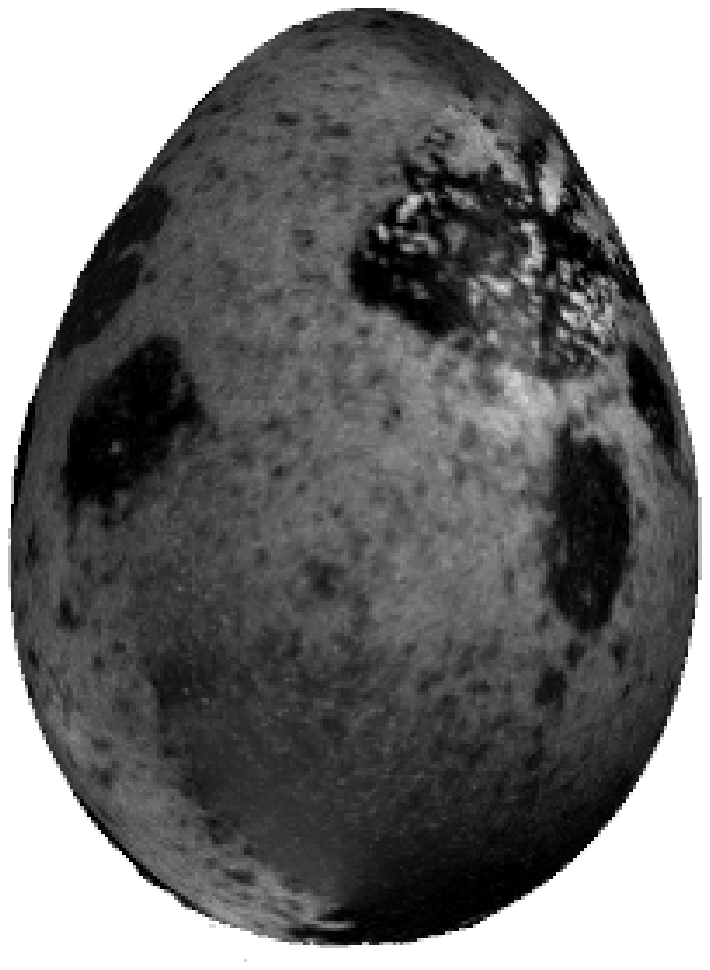} &
     \includegraphics[width=0.06\hsize]{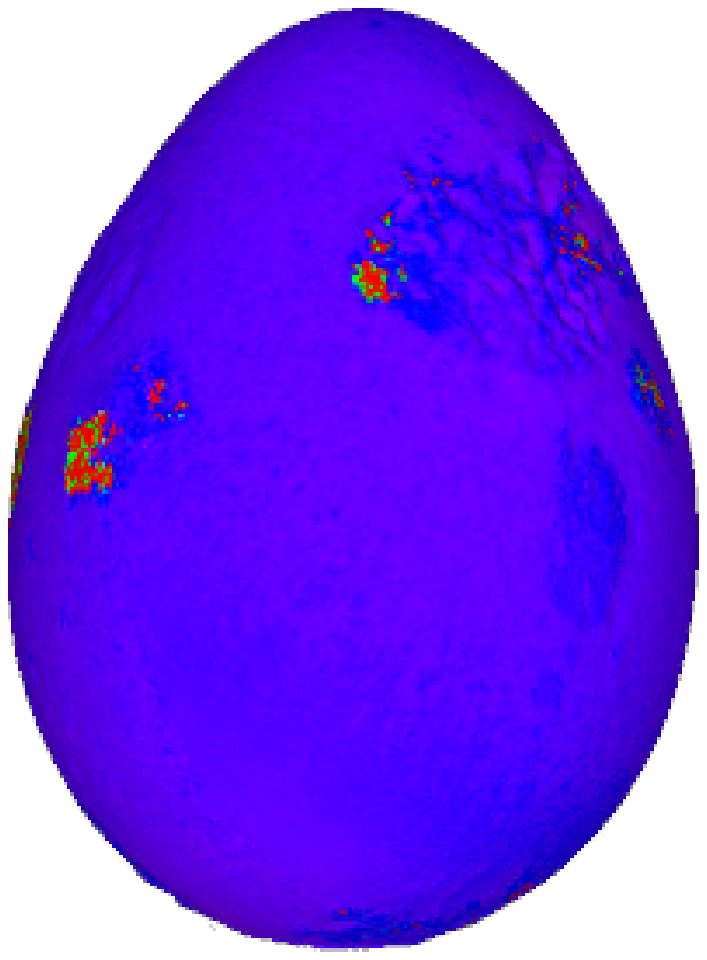} & 
     \includegraphics[width=0.06\hsize]{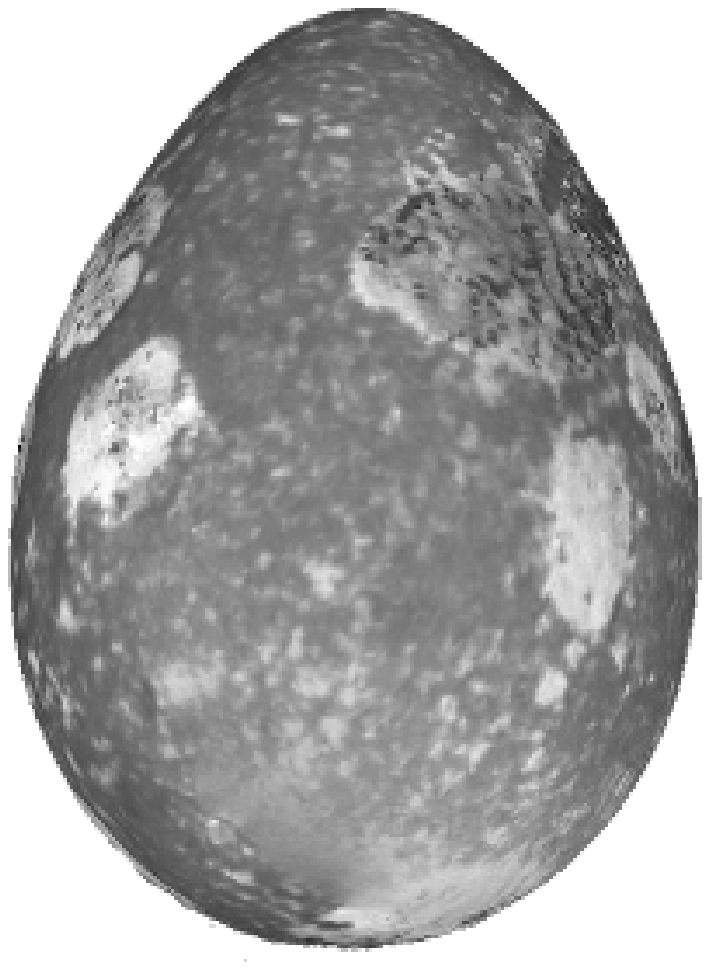} &
     \includegraphics[width=0.06\hsize]{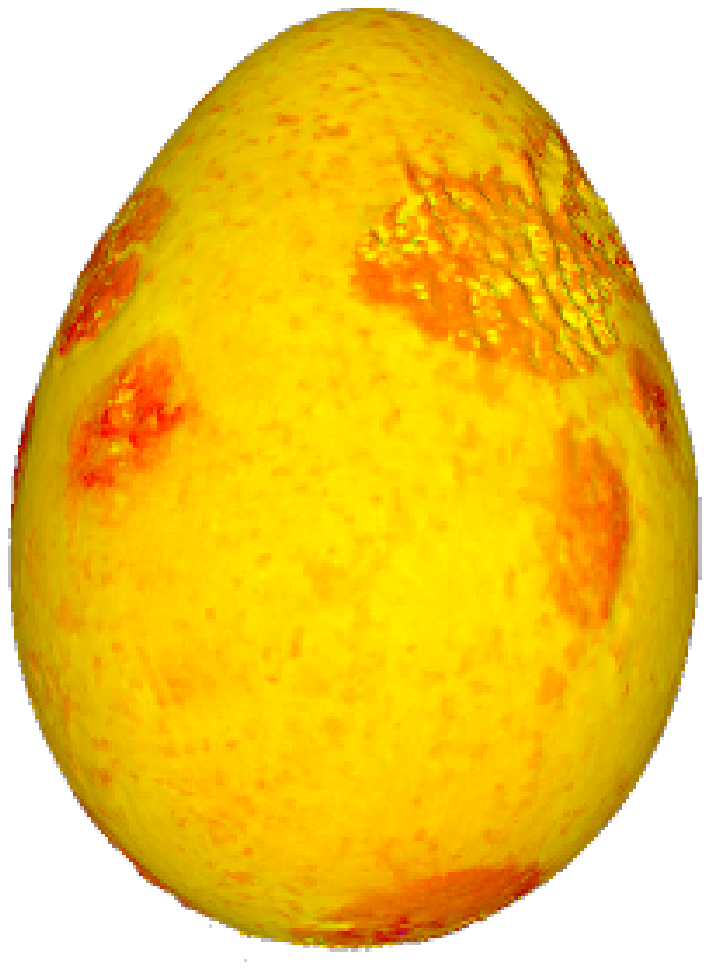} \\
     
     \multicolumn{2}{c}{\scriptsize Coturnix Quail} & & & & & & \footnotesize intensity & \footnotesize hue & \footnotesize saturation & \footnotesize hue \\
     
     & & & & & & & \multicolumn{2}{c}{\scriptsize UV (365nm)} & \multicolumn{2}{c}{\scriptsize visible (400nm-700nm)} \\
\end{tabular}

%\vspace{1mm}

\begin{tabular}{@{\hspace{0.5\tabcolsep}} c @{\hspace{0.5\tabcolsep}} c @{\hspace{0.5\tabcolsep}} c @{\hspace{0.5\tabcolsep}} c @{\hspace{0.5\tabcolsep}} c @{\hspace{0.5\tabcolsep}} c @{\hspace{0.25\tabcolsep}} c @{\hspace{0.5\tabcolsep}} c @{\hspace{0.5\tabcolsep}} c @{\hspace{0.5\tabcolsep}} c @{\hspace{0.5\tabcolsep}} c }
    \multicolumn{11}{@{\hspace{0.5\tabcolsep}} l}{\footnotesize Test 3: \textbf{similar} shape and scale, \textbf{different} patterns and color} \\
	
	\multirow{2}{*}{reconstructed} & \multirow{2}{*}{GT} & & & & & & \multicolumn{2}{@{\hspace{0.5\tabcolsep}} c}{\footnotesize material properties} & \multicolumn{2}{@{\hspace{0.5\tabcolsep}} c}{\footnotesize appearance} \\
	 
	 & & & \footnotesize hue & \footnotesize saturation & & & \footnotesize concentration & \footnotesize composition & & \\
     
     %& & & & & & & \scriptsize (log scale) & & & \\
     
     \multirow{2}{*}[0.07\hsize]{\includegraphics[width=0.12\hsize]{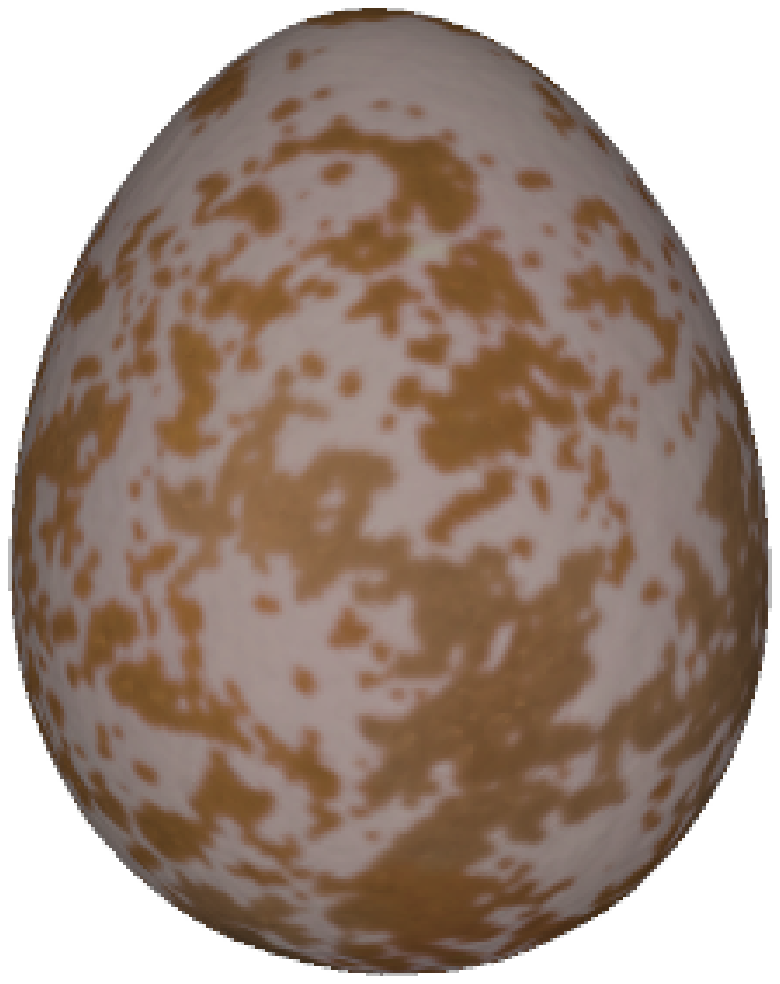}} & \multirow{2}{*}[0.07\hsize]{\includegraphics[width=0.12\hsize]{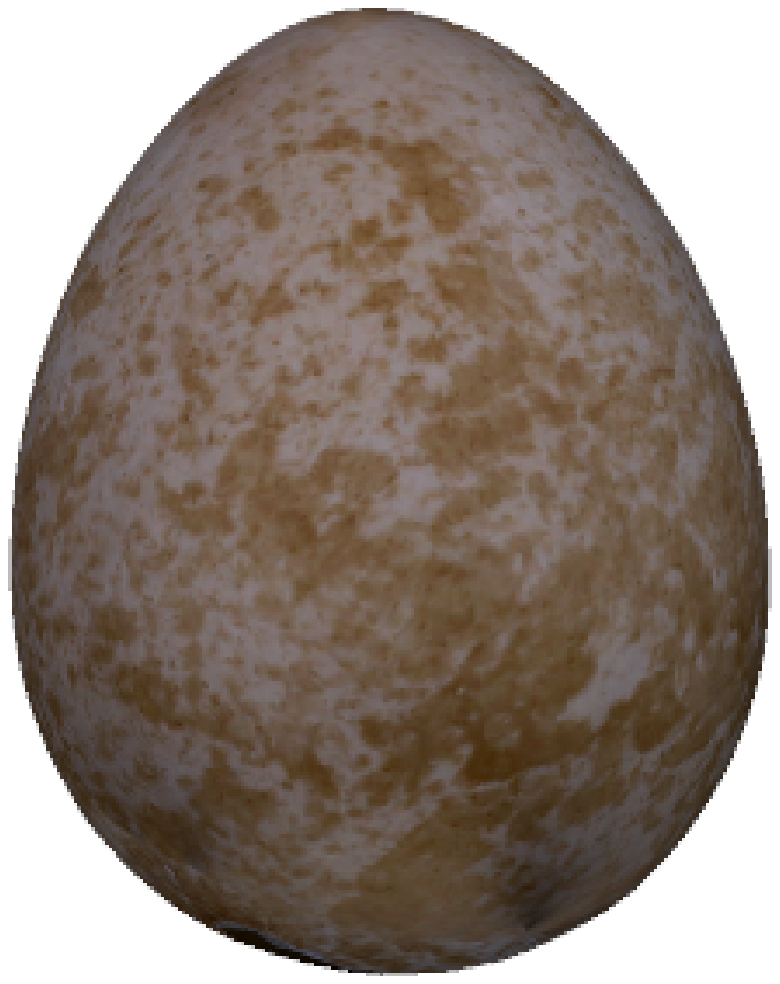}} & 
     \raisebox{1.25\normalbaselineskip}[0pt][0pt]{\rotatebox[origin=c]{90}{\footnotesize reconstructed}} &
     \includegraphics[width=0.06\hsize]{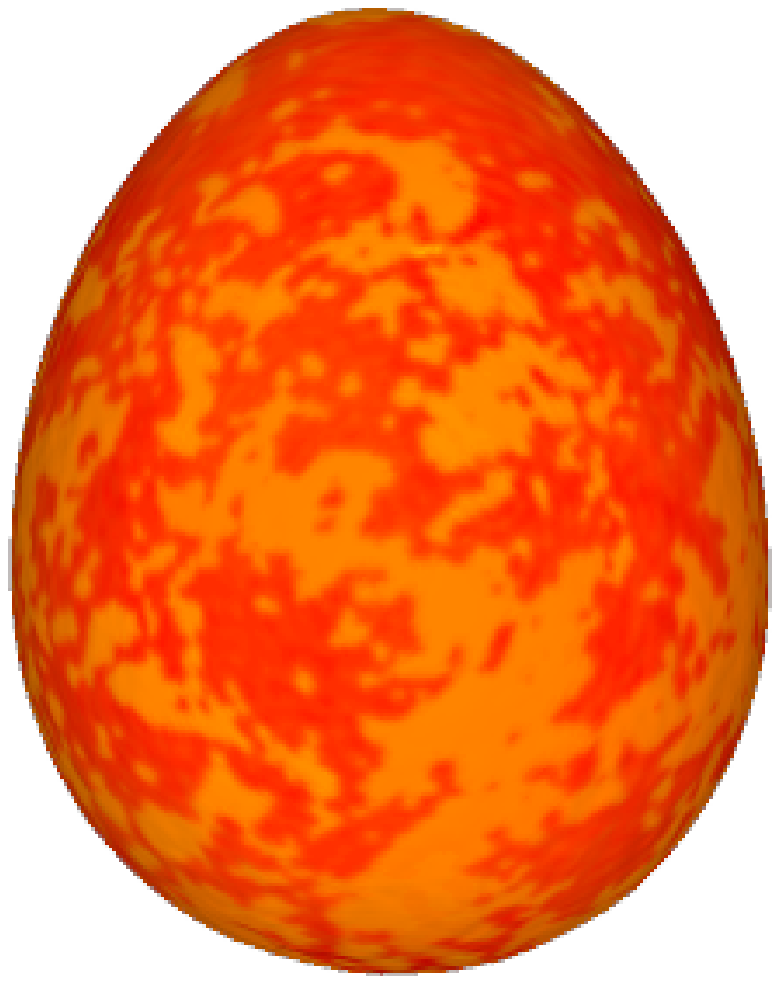} &
     \includegraphics[width=0.06\hsize]{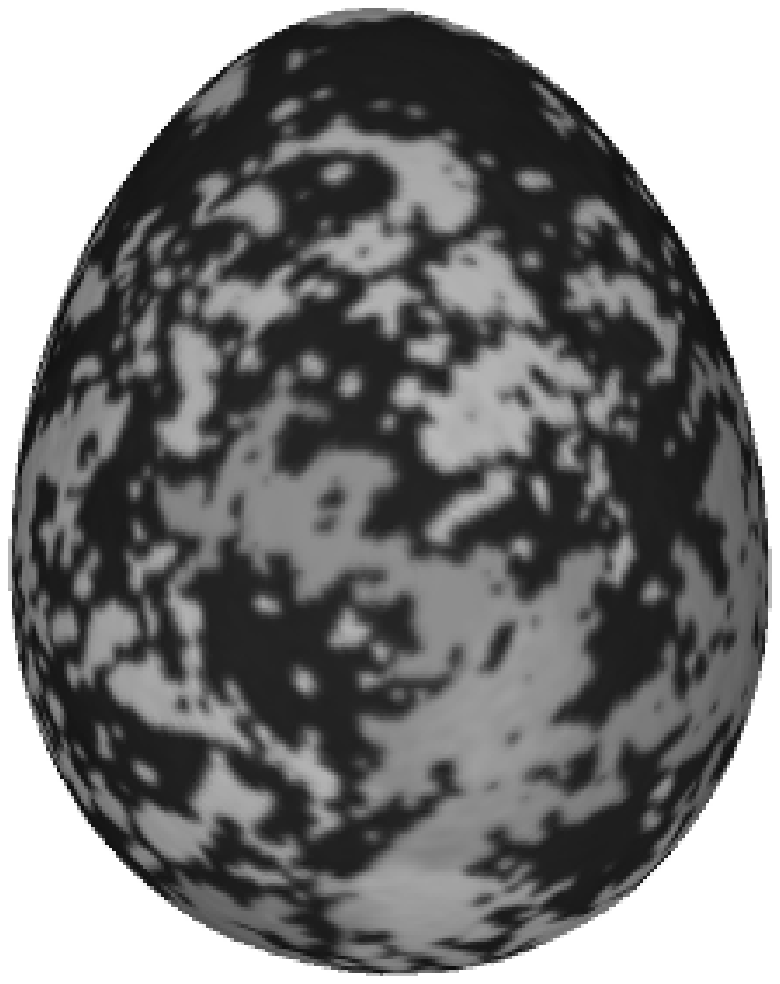} &
     \multirow{2}{*}{\rotatebox[origin=c]{90}{\footnotesize input}} & 
	 \raisebox{1.25\normalbaselineskip}[0pt][0pt]{\rotatebox[origin=c]{90}{\footnotesize target}} &
     \includegraphics[width=0.06\hsize]{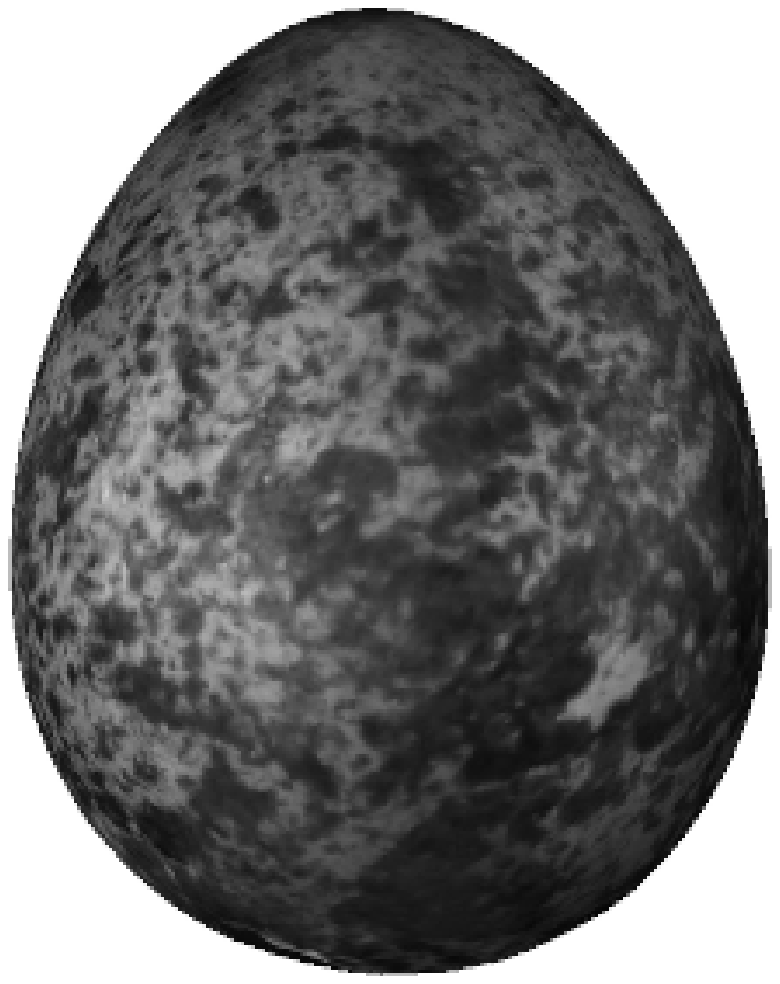} &
     \includegraphics[width=0.06\hsize]{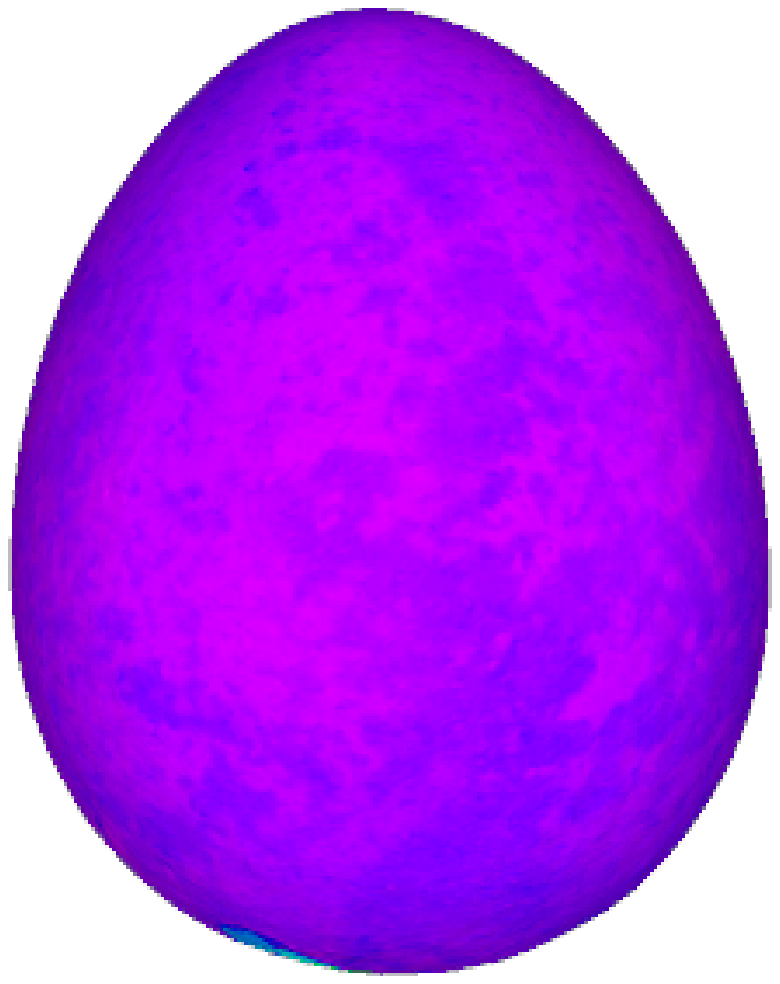} & 
     & \\
     
     & & \raisebox{1.25\normalbaselineskip}[0pt][0pt]{\rotatebox[origin=c]{90}{\footnotesize GT}} & 
     \includegraphics[width=0.06\hsize]{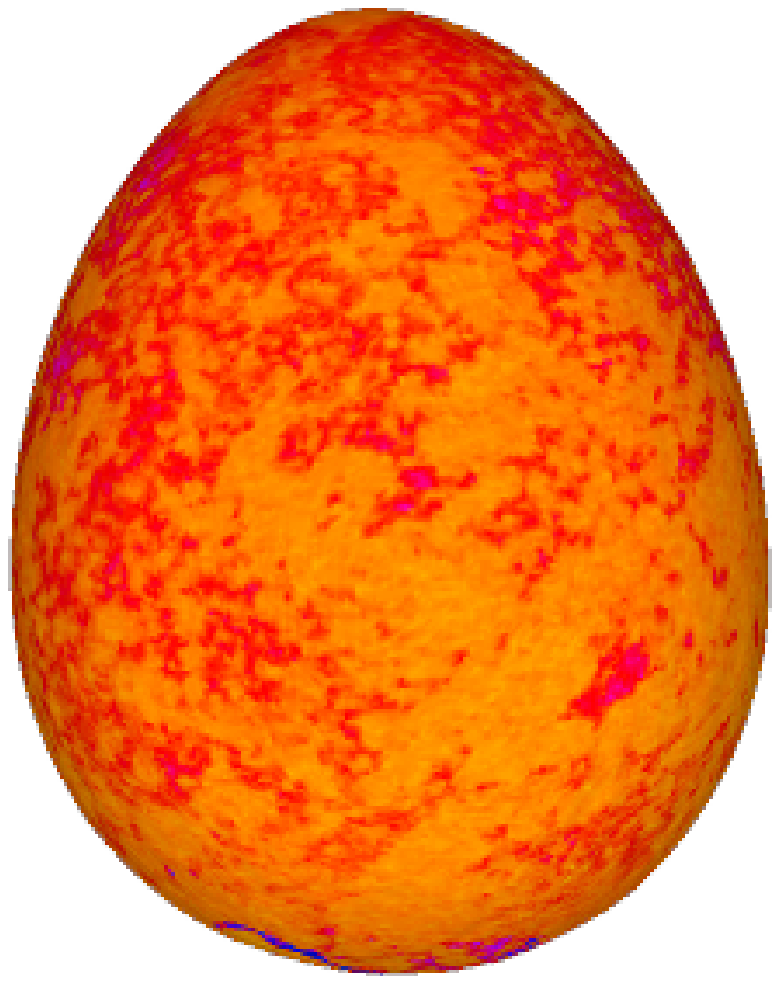} &
     \includegraphics[width=0.06\hsize]{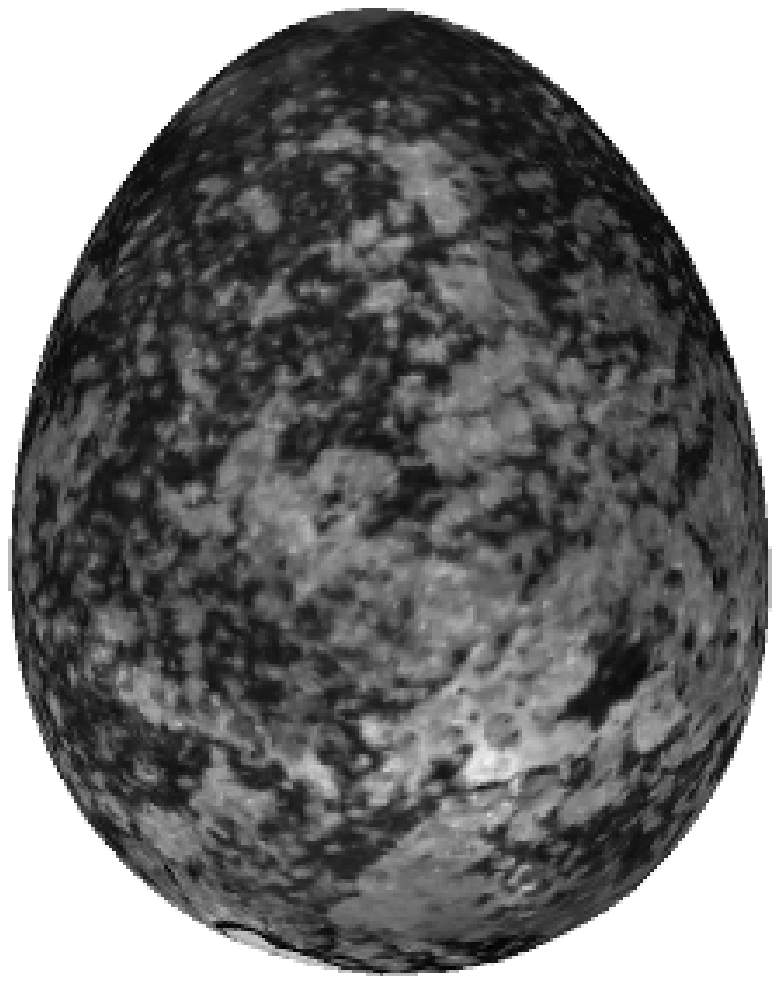} & &
     \raisebox{1.25\normalbaselineskip}[0pt][0pt]{\rotatebox[origin=c]{90}{\footnotesize source}} & 
     \includegraphics[width=0.06\hsize]{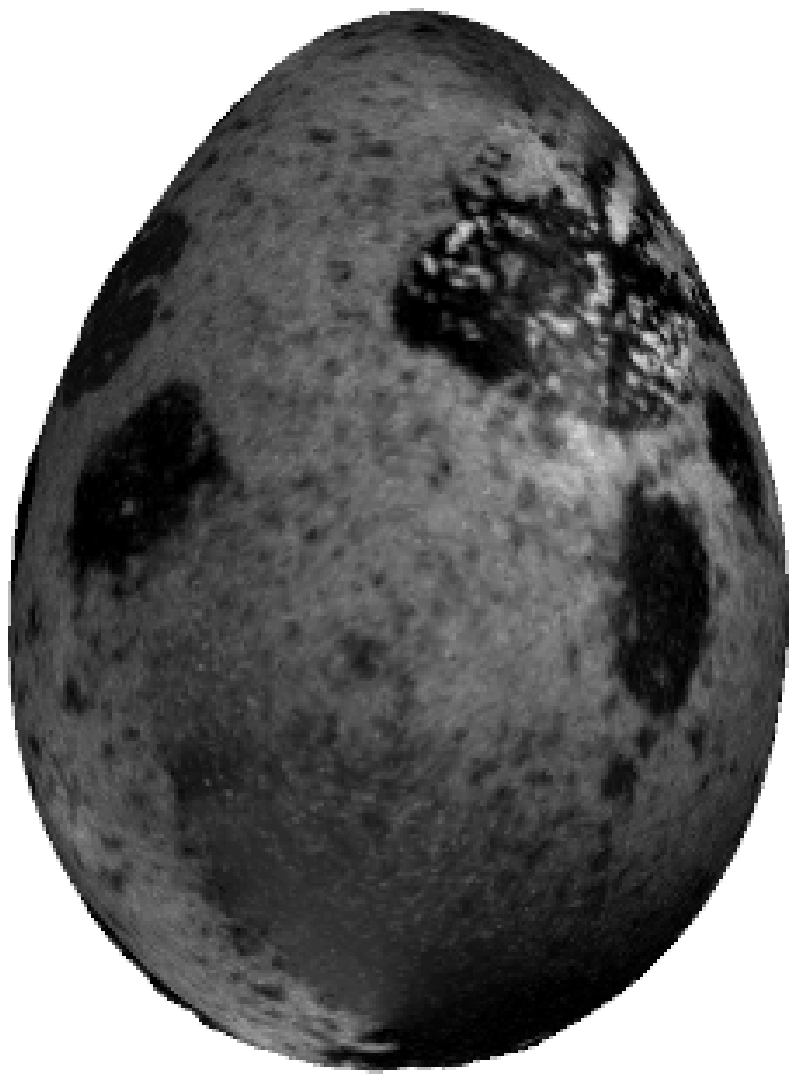} &
     \includegraphics[width=0.06\hsize]{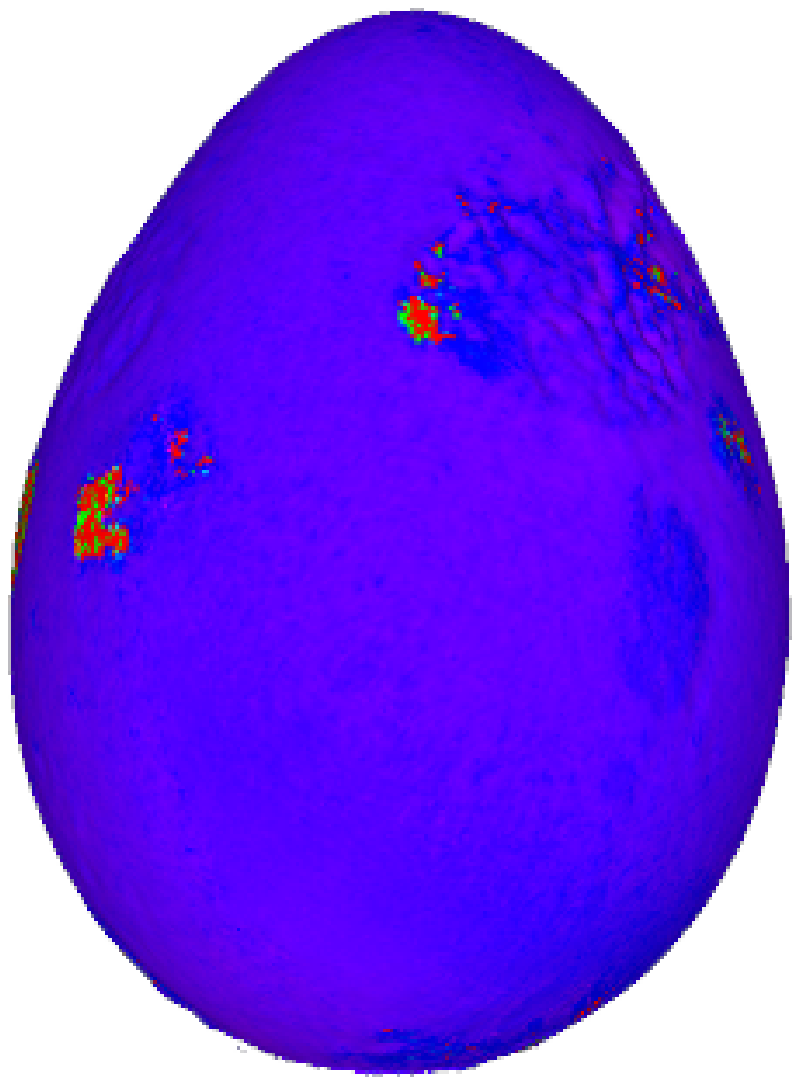} & 
     \includegraphics[width=0.06\hsize]{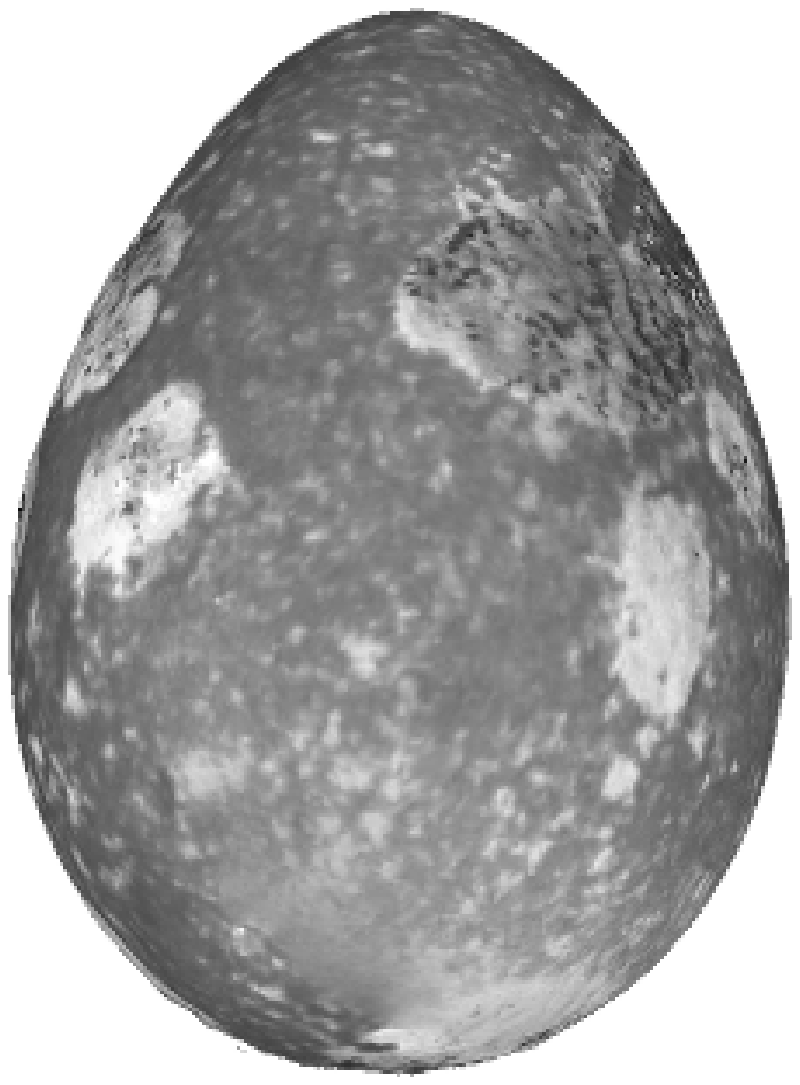} &
     \includegraphics[width=0.06\hsize]{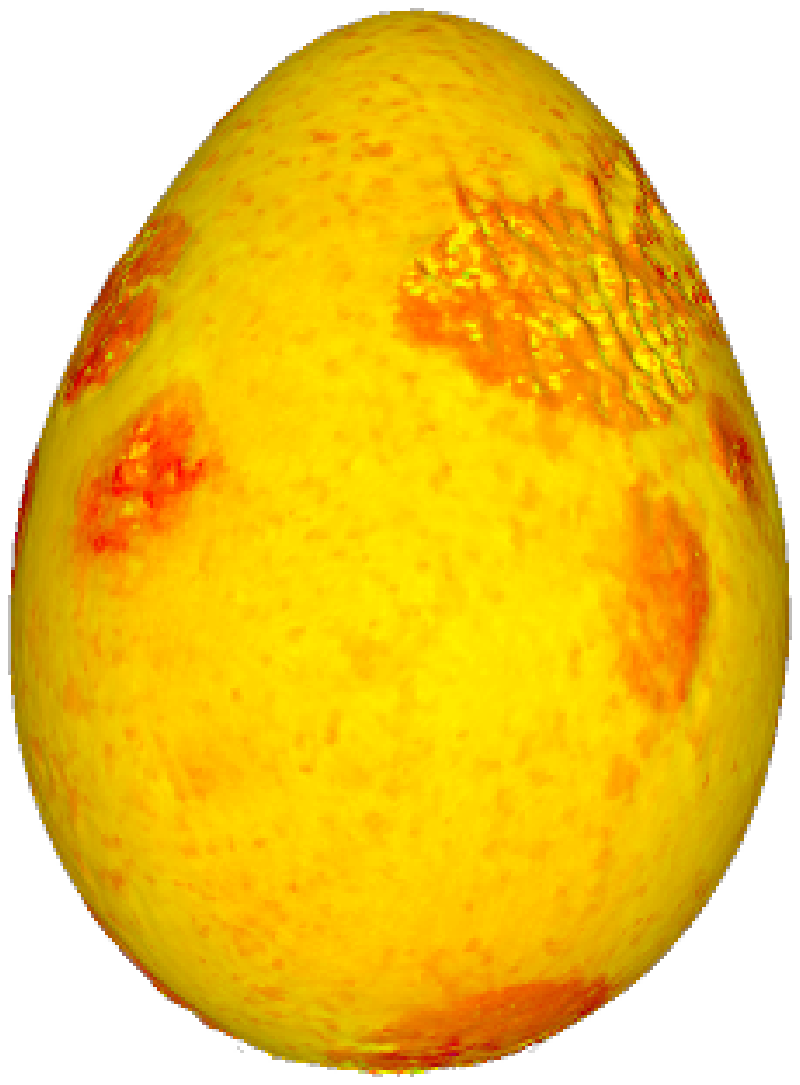} \\
     
     \multicolumn{2}{c}{\scriptsize Valley Quail} & & & & & & \footnotesize intensity & \footnotesize hue & \footnotesize saturation & \footnotesize hue \\
     
     & & & & & & & \multicolumn{2}{c}{\scriptsize UV (365nm)} & \multicolumn{2}{c}{\scriptsize visible (400nm-700nm)} \\
\end{tabular}

%\vspace{1mm}

\begin{tabular}{@{\hspace{0.5\tabcolsep}} c @{\hspace{0.5\tabcolsep}} c @{\hspace{0.5\tabcolsep}} c @{\hspace{0.5\tabcolsep}} c @{\hspace{0.5\tabcolsep}} c @{\hspace{0.5\tabcolsep}} c @{\hspace{0.25\tabcolsep}} c @{\hspace{0.5\tabcolsep}} c @{\hspace{0.5\tabcolsep}} c @{\hspace{0.5\tabcolsep}} c @{\hspace{0.5\tabcolsep}} c }
    \multicolumn{11}{@{\hspace{0.5\tabcolsep}} l}{\footnotesize Test 4: \textbf{similar} patterns and color, \textbf{different} shape and scale} \\
	
	\multirow{2}{*}{reconstructed} & \multirow{2}{*}{GT} & & & & & & \multicolumn{2}{@{\hspace{0.5\tabcolsep}} c}{\footnotesize material properties} & \multicolumn{2}{@{\hspace{0.5\tabcolsep}} c}{\footnotesize appearance} \\
	 
	 & & & \footnotesize hue & \footnotesize saturation & & & \footnotesize concentration & \footnotesize composition & & \\
     
     %& & & & & & & \scriptsize (log scale) & & & \\
     
     \multirow{2}{*}[0.07\hsize]{\includegraphics[width=0.12\hsize]{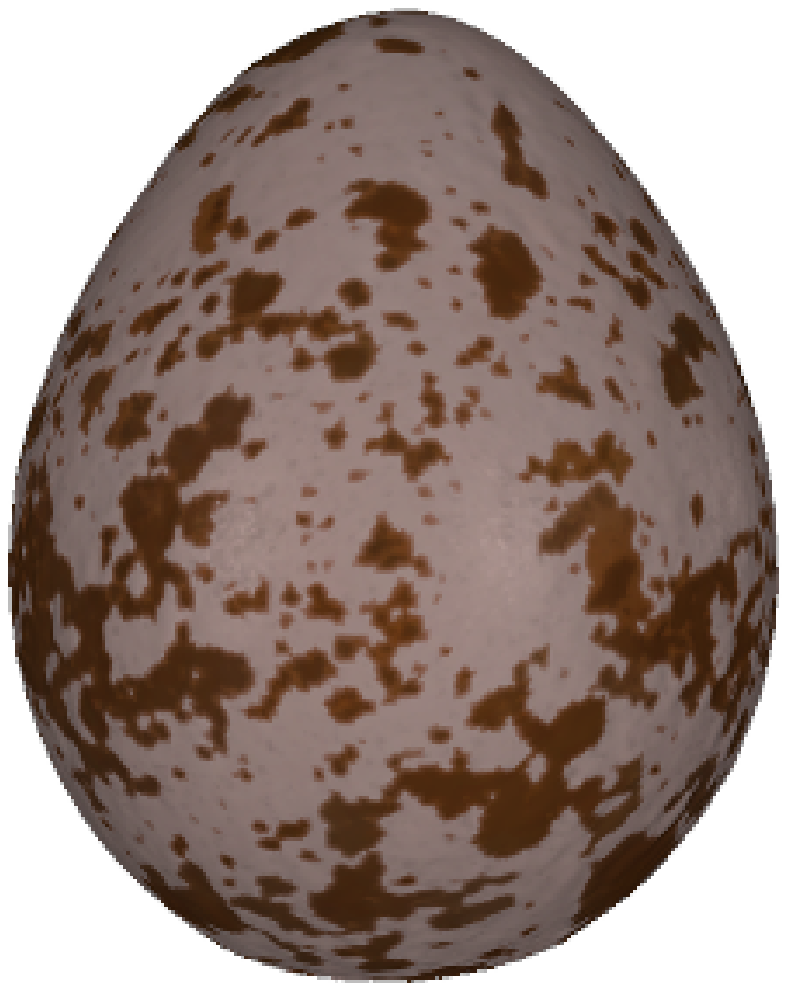}} & \multirow{2}{*}[0.07\hsize]{\includegraphics[width=0.12\hsize]{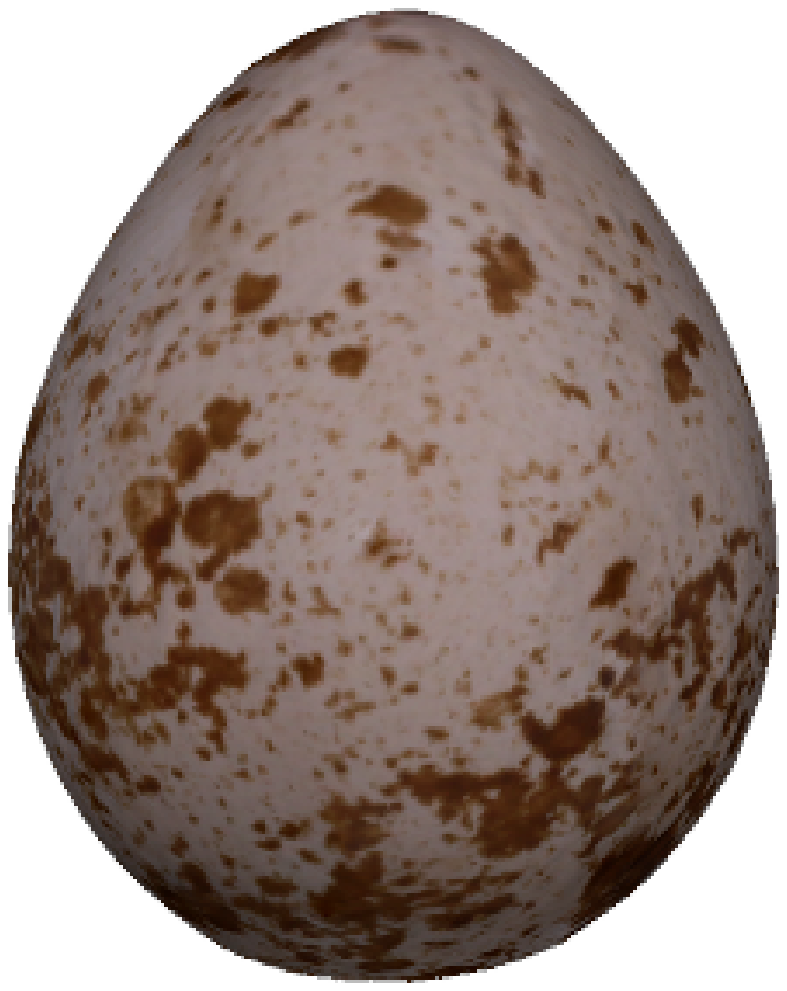}} & 
     \raisebox{1.25\normalbaselineskip}[0pt][0pt]{\rotatebox[origin=c]{90}{\footnotesize reconstructed}} &
     \includegraphics[width=0.06\hsize]{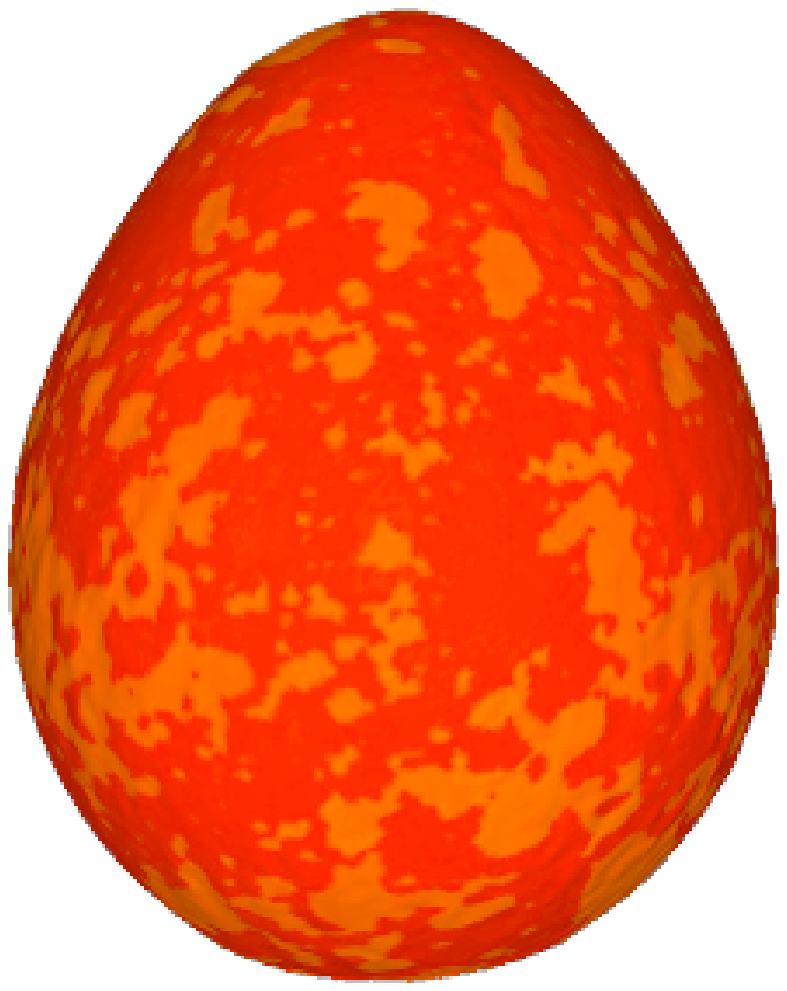} &
     \includegraphics[width=0.06\hsize]{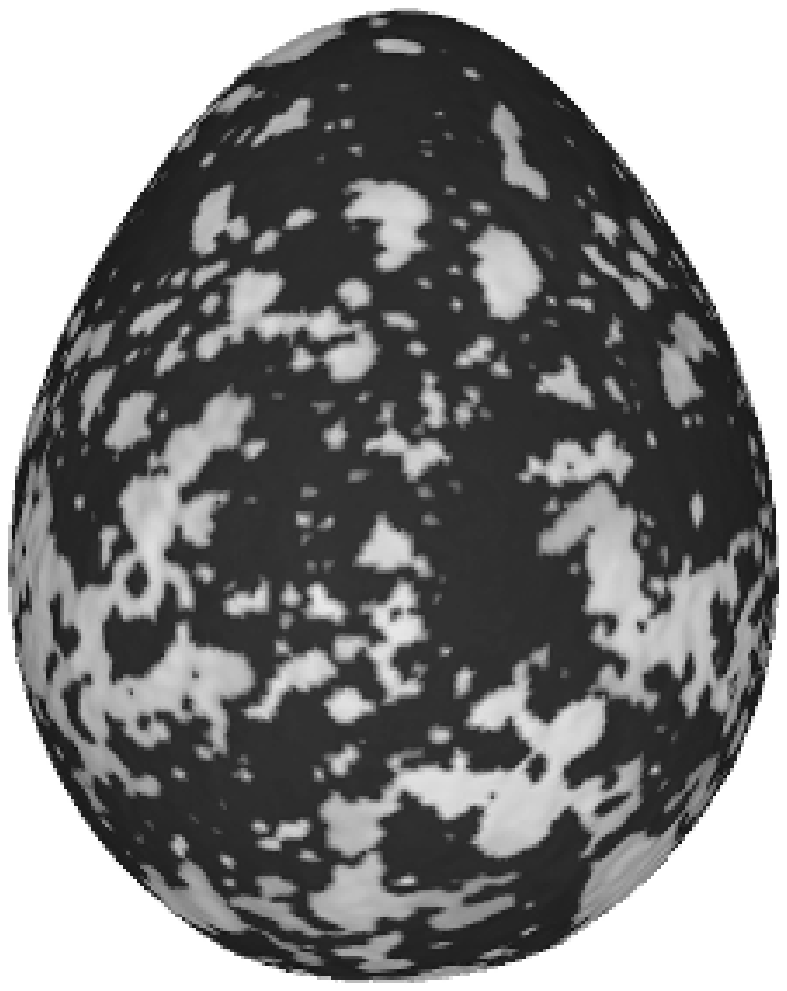} &
     \multirow{2}{*}{\rotatebox[origin=c]{90}{\footnotesize input}} & 
	 \raisebox{1.25\normalbaselineskip}[0pt][0pt]{\rotatebox[origin=c]{90}{\footnotesize target}} &
     \includegraphics[width=0.06\hsize]{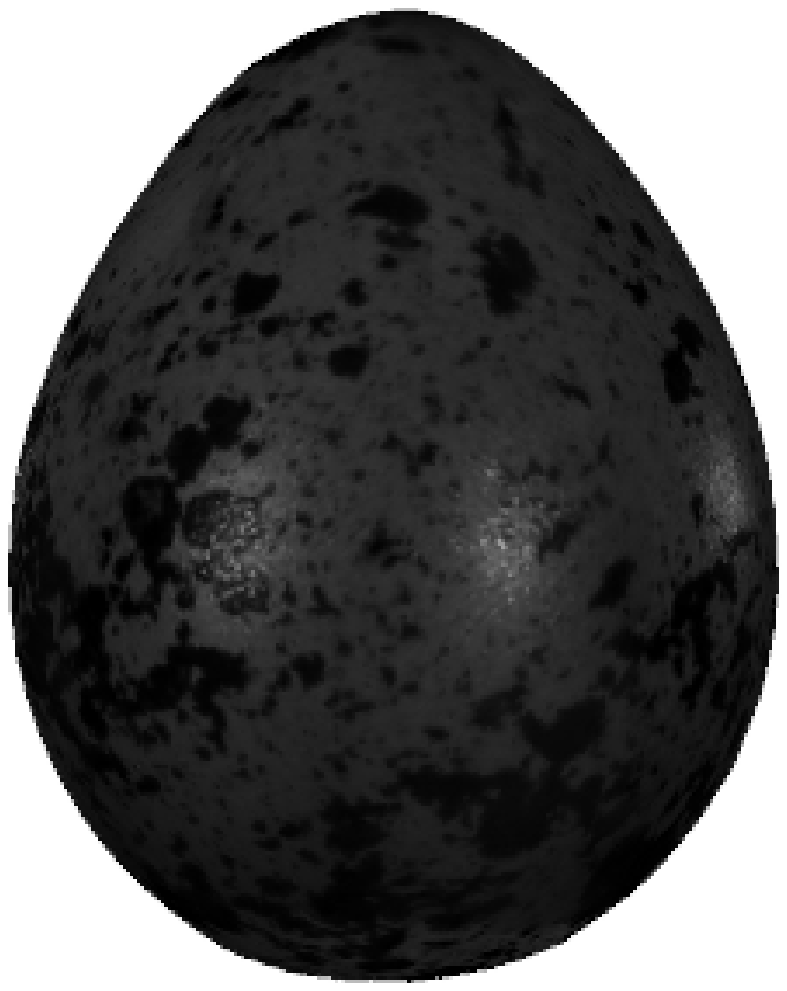} &
     \includegraphics[width=0.06\hsize]{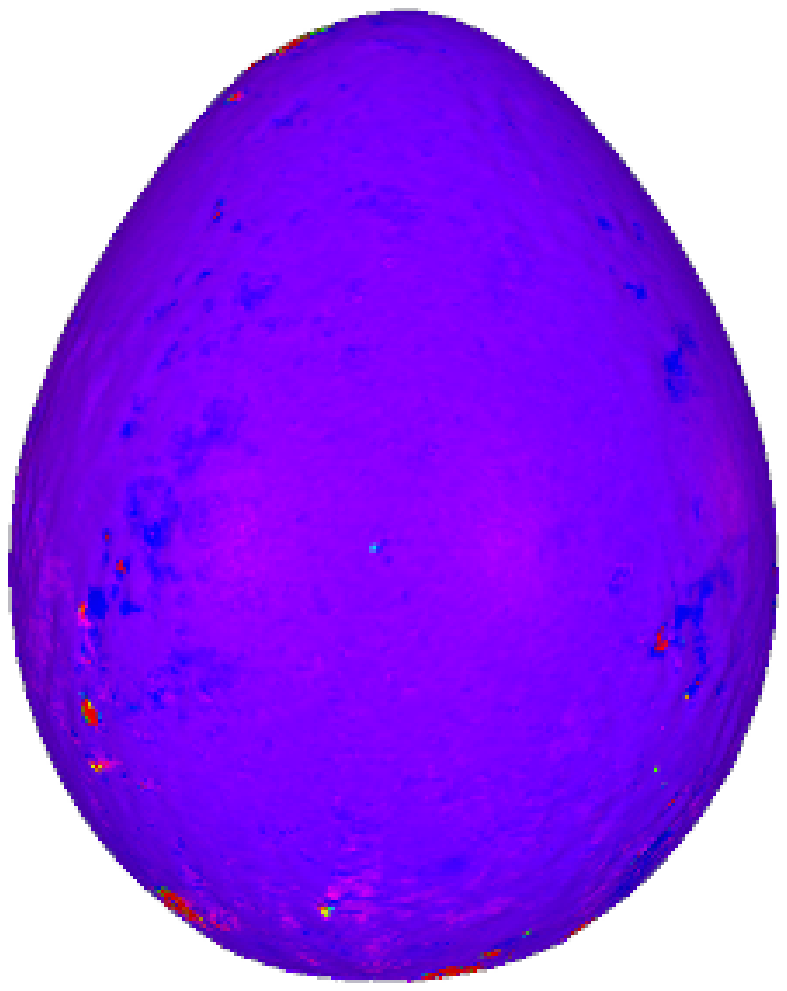} & 
     & \\
     
     & & \raisebox{1.25\normalbaselineskip}[0pt][0pt]{\rotatebox[origin=c]{90}{\footnotesize GT}} & 
     \includegraphics[width=0.06\hsize]{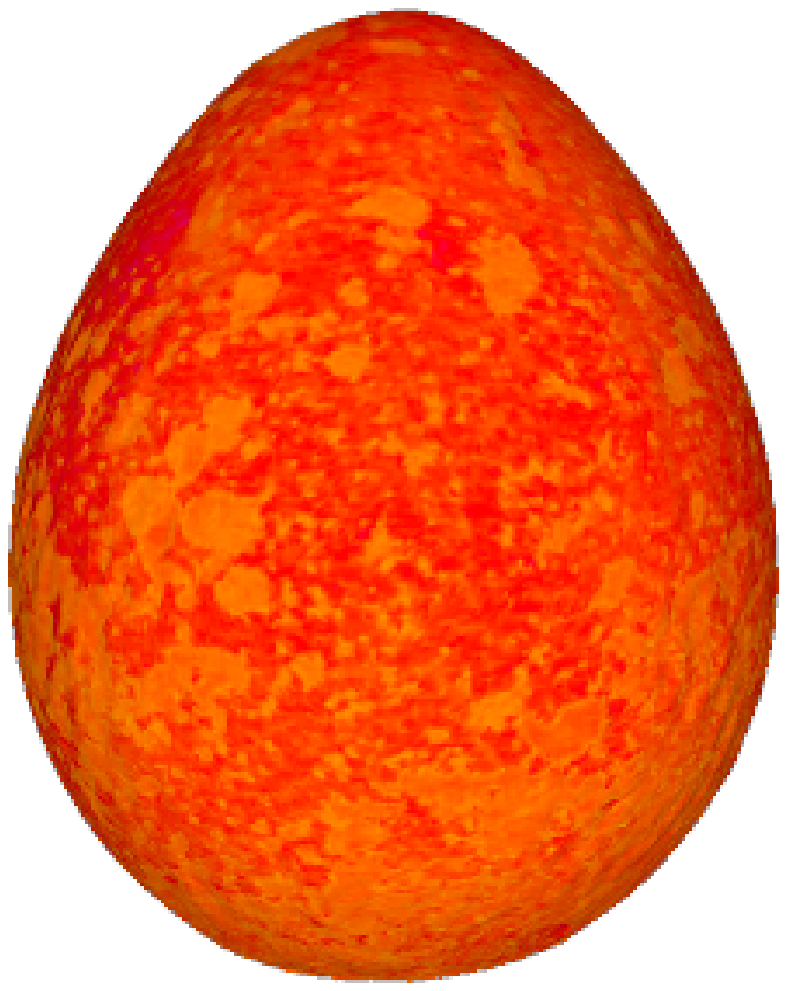} &
     \includegraphics[width=0.06\hsize]{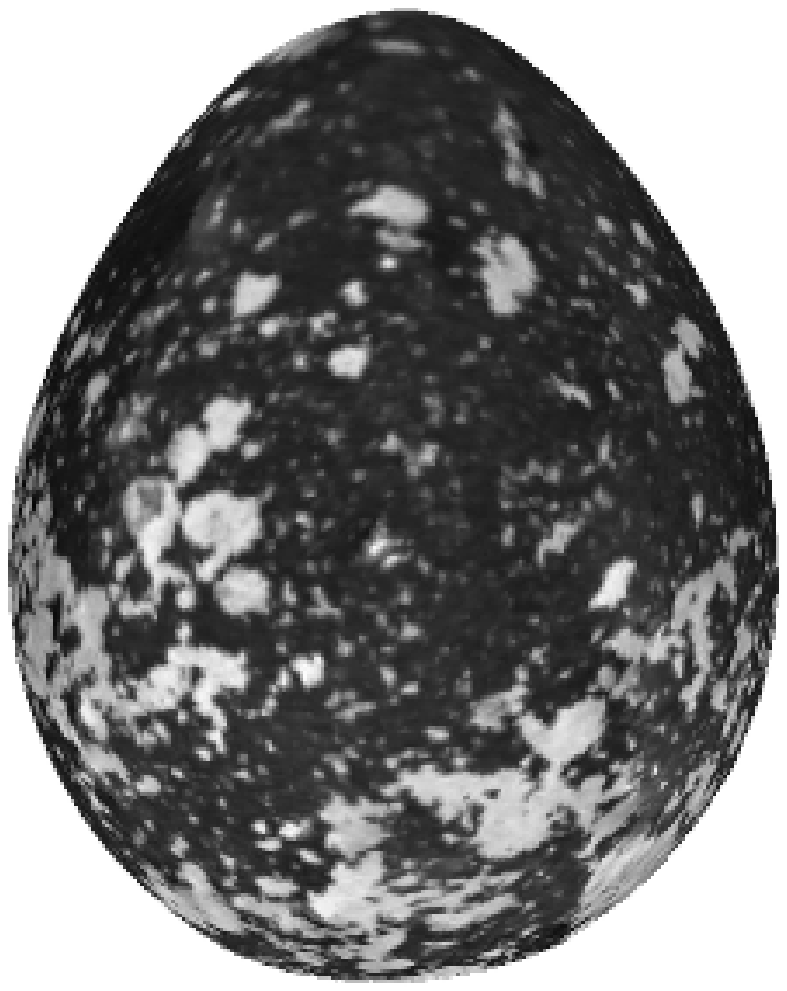} & &
     \raisebox{1.25\normalbaselineskip}[0pt][0pt]{\rotatebox[origin=c]{90}{\footnotesize source}} & 
     \includegraphics[width=0.06\hsize]{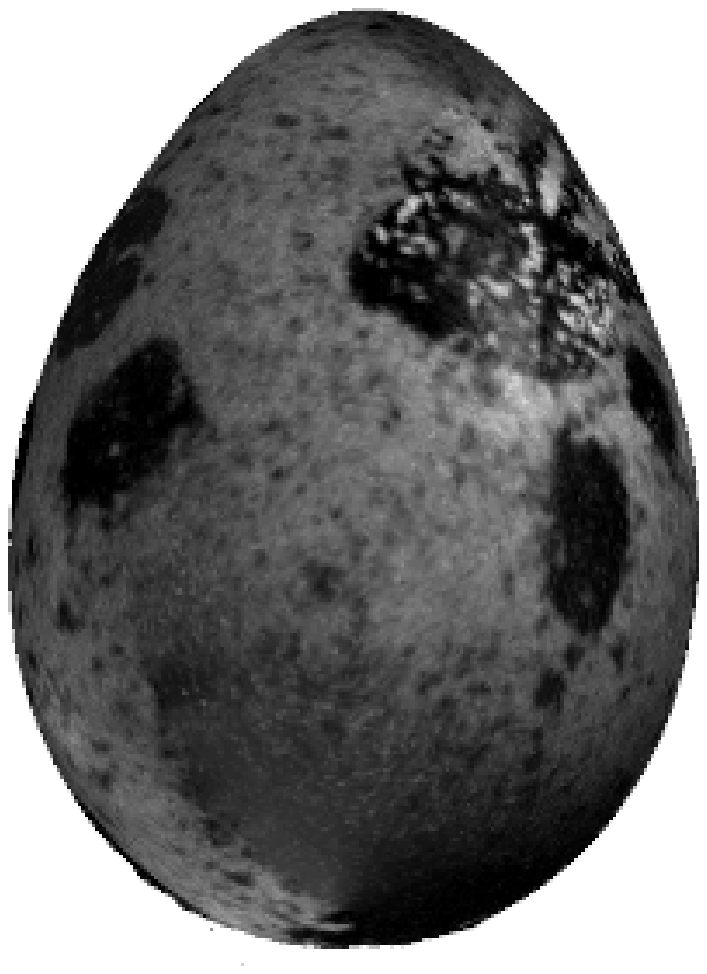} &
     \includegraphics[width=0.06\hsize]{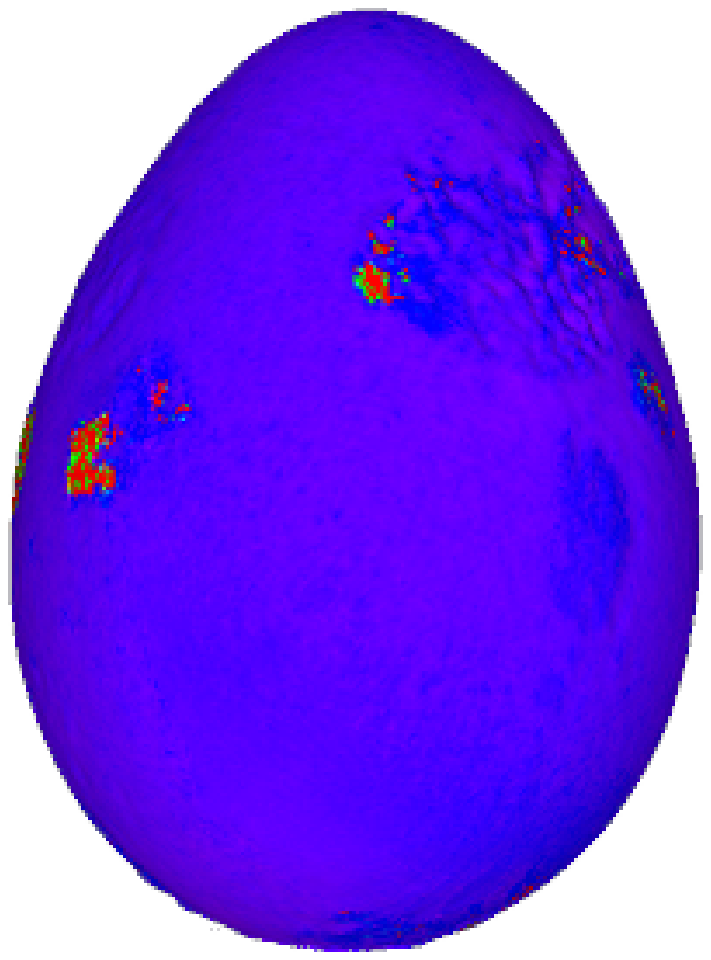} & 
     \includegraphics[width=0.06\hsize]{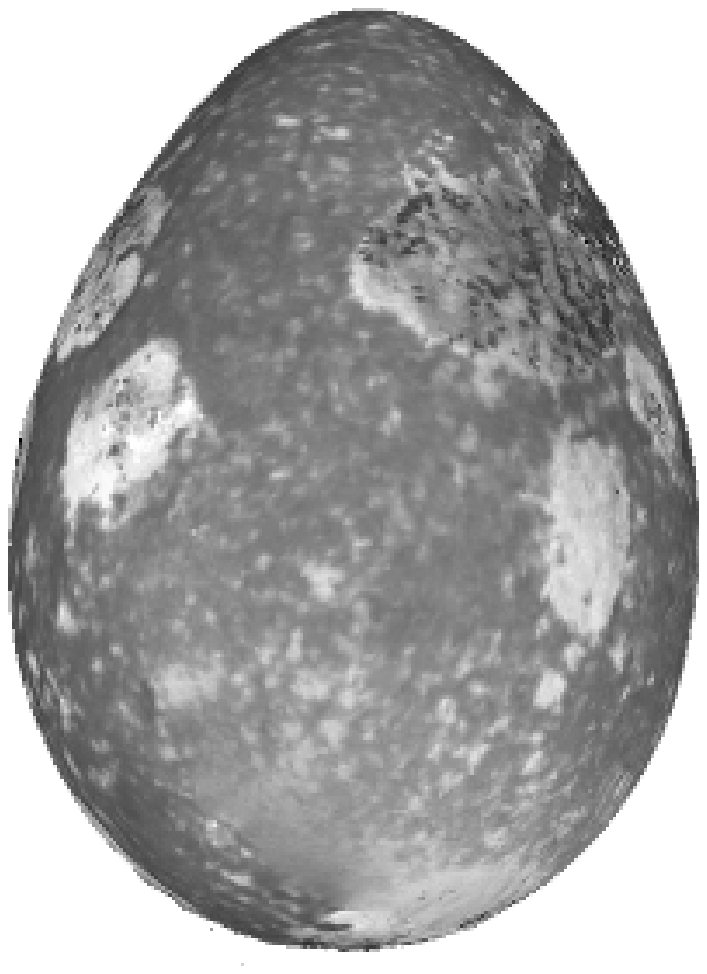} &
     \includegraphics[width=0.06\hsize]{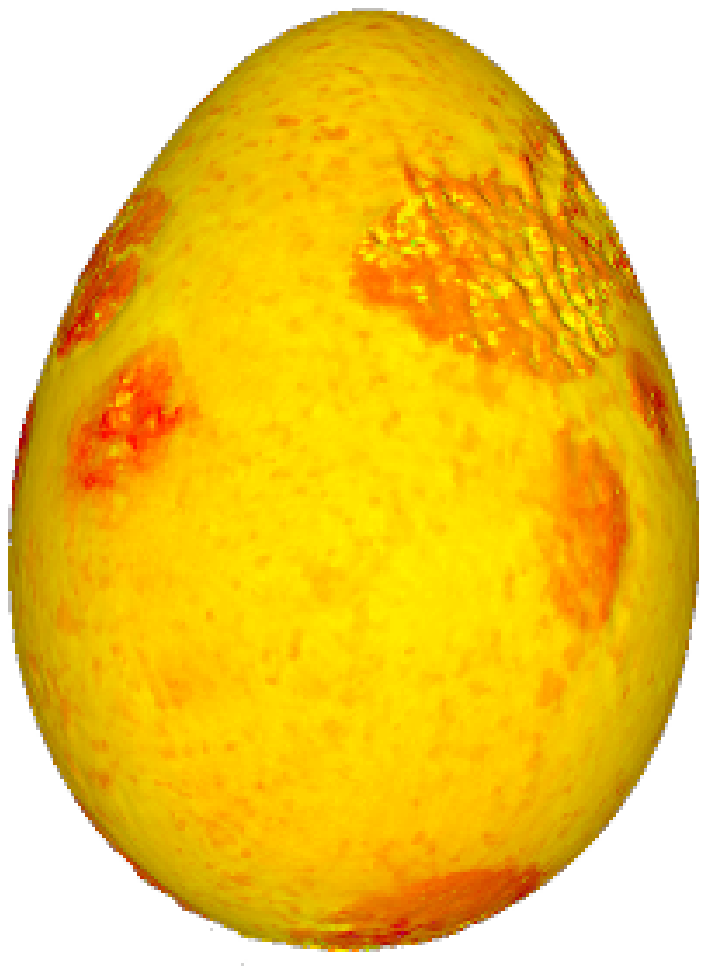} \\
     
      \multicolumn{2}{c}{\scriptsize Gambles Quail} & & & & & & \footnotesize intensity & \footnotesize hue & \footnotesize saturation & \footnotesize hue \\
     
     & & & & & & & \multicolumn{2}{c}{\scriptsize UV (365nm)} & \multicolumn{2}{c}{\scriptsize visible (400nm-700nm)} \\
\end{tabular}

%\vspace{1mm}

\begin{tabular}{@{\hspace{0.5\tabcolsep}} c @{\hspace{0.5\tabcolsep}} c @{\hspace{0.5\tabcolsep}} c @{\hspace{0.5\tabcolsep}} c @{\hspace{0.5\tabcolsep}} c @{\hspace{0.5\tabcolsep}} c @{\hspace{0.25\tabcolsep}} c @{\hspace{0.5\tabcolsep}} c @{\hspace{0.5\tabcolsep}} c @{\hspace{0.5\tabcolsep}} c @{\hspace{0.5\tabcolsep}} c }
    \multicolumn{11}{@{\hspace{0.5\tabcolsep}} l}{\footnotesize Test 5: \textbf{different} shape, scale, patterns and  color} \\
	
	\multirow{2}{*}{reconstructed} & \multirow{2}{*}{GT} & & & & & & \multicolumn{2}{@{\hspace{0.5\tabcolsep}} c}{\footnotesize material properties} & \multicolumn{2}{@{\hspace{0.5\tabcolsep}} c}{\footnotesize appearance} \\
	 
	 & & & \footnotesize hue & \footnotesize saturation & & & \footnotesize concentration & \footnotesize composition & & \\
     
     %& & & & & & & \scriptsize (log scale) & & & \\
     
     \multirow{2}{*}[0.07\hsize]{\includegraphics[width=0.12\hsize]{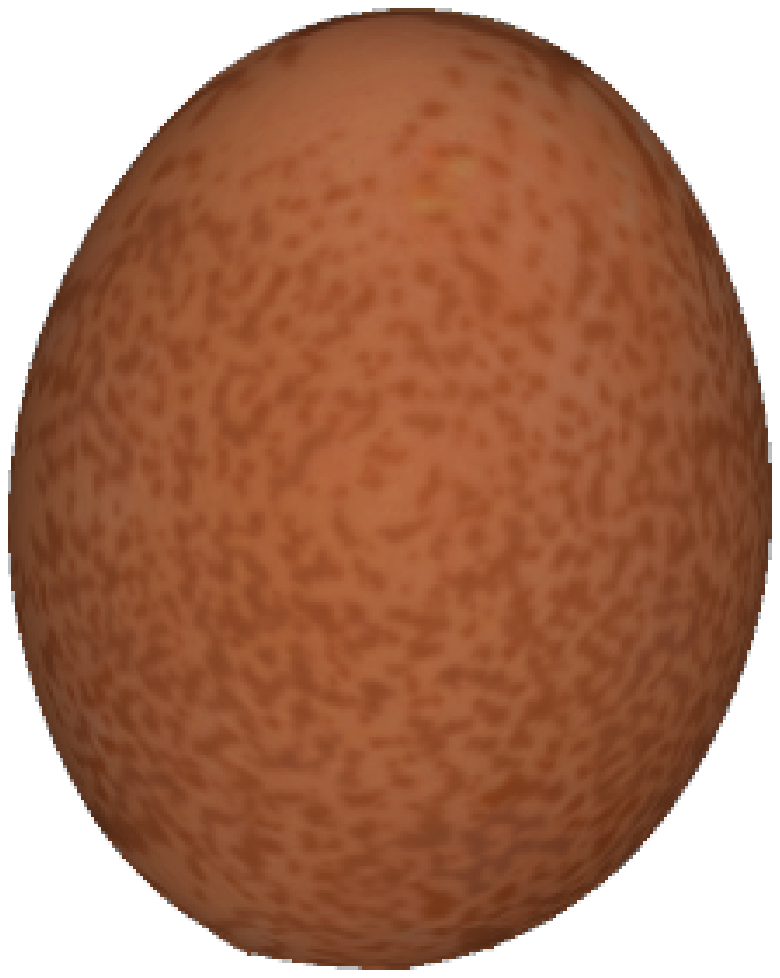}} & \multirow{2}{*}[0.07\hsize]{\includegraphics[width=0.12\hsize]{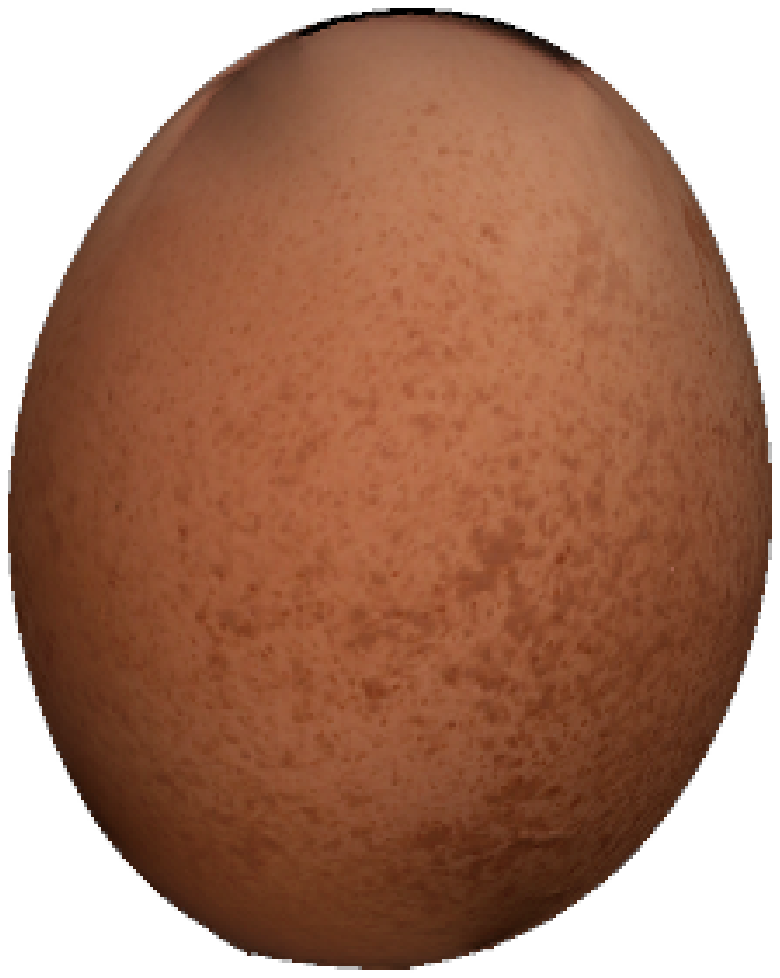}} & 
     \raisebox{1.25\normalbaselineskip}[0pt][0pt]{\rotatebox[origin=c]{90}{\footnotesize reconstructed}} &
     \includegraphics[width=0.06\hsize]{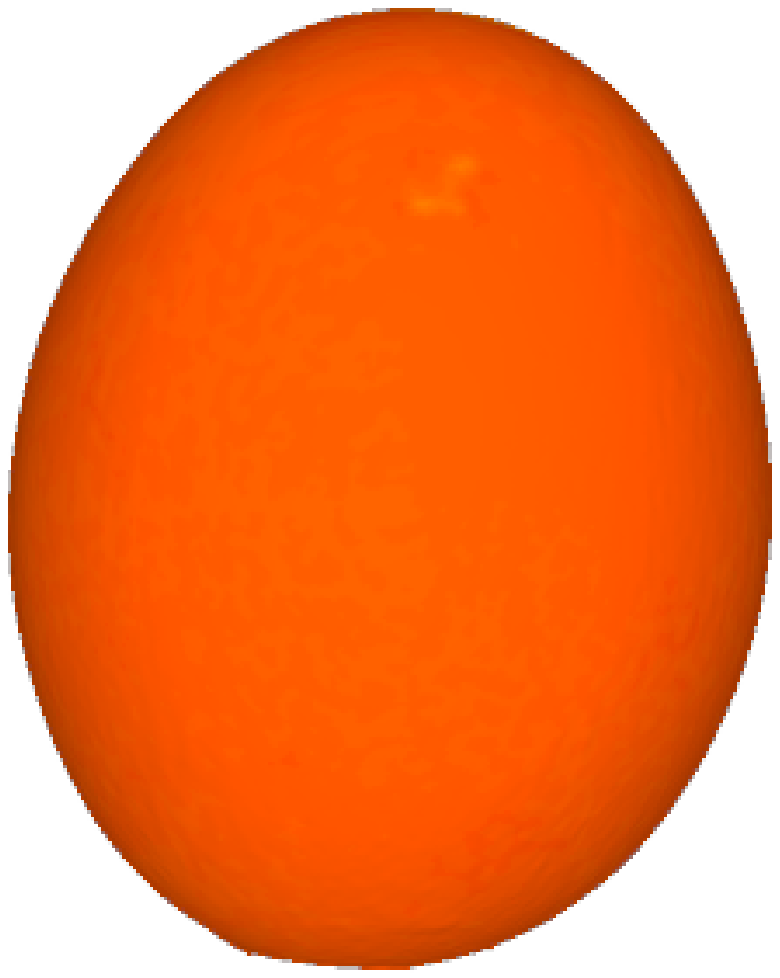} &
     \includegraphics[width=0.06\hsize]{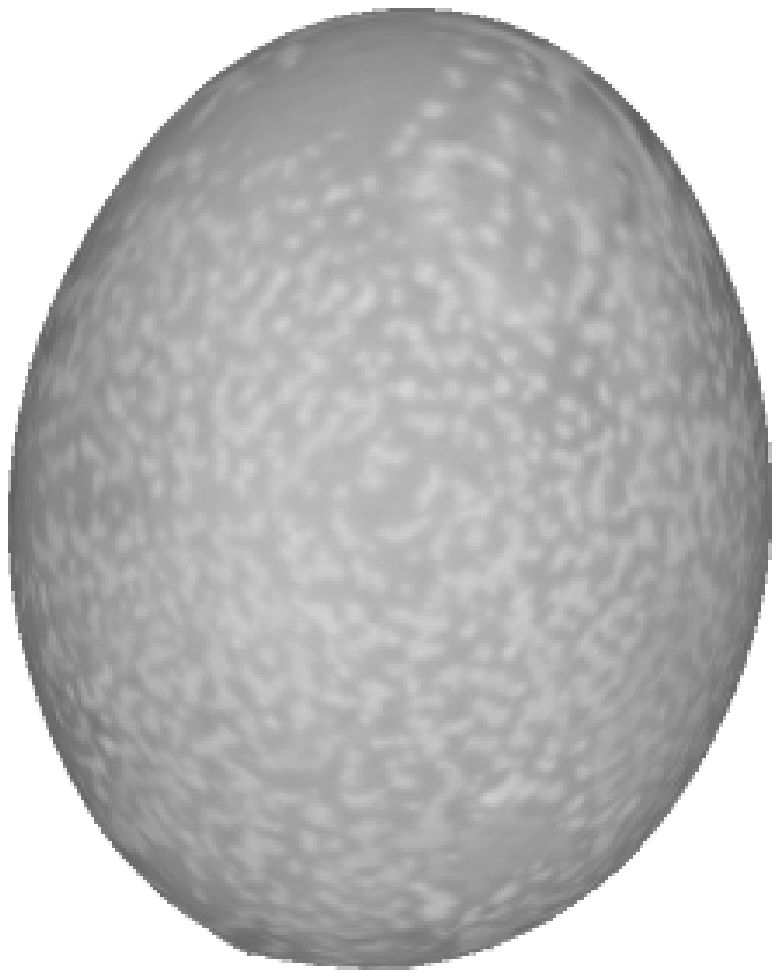} &
     \multirow{2}{*}{\rotatebox[origin=c]{90}{\footnotesize input}} & 
	 \raisebox{1.25\normalbaselineskip}[0pt][0pt]{\rotatebox[origin=c]{90}{\footnotesize target}} &
     \includegraphics[width=0.06\hsize]{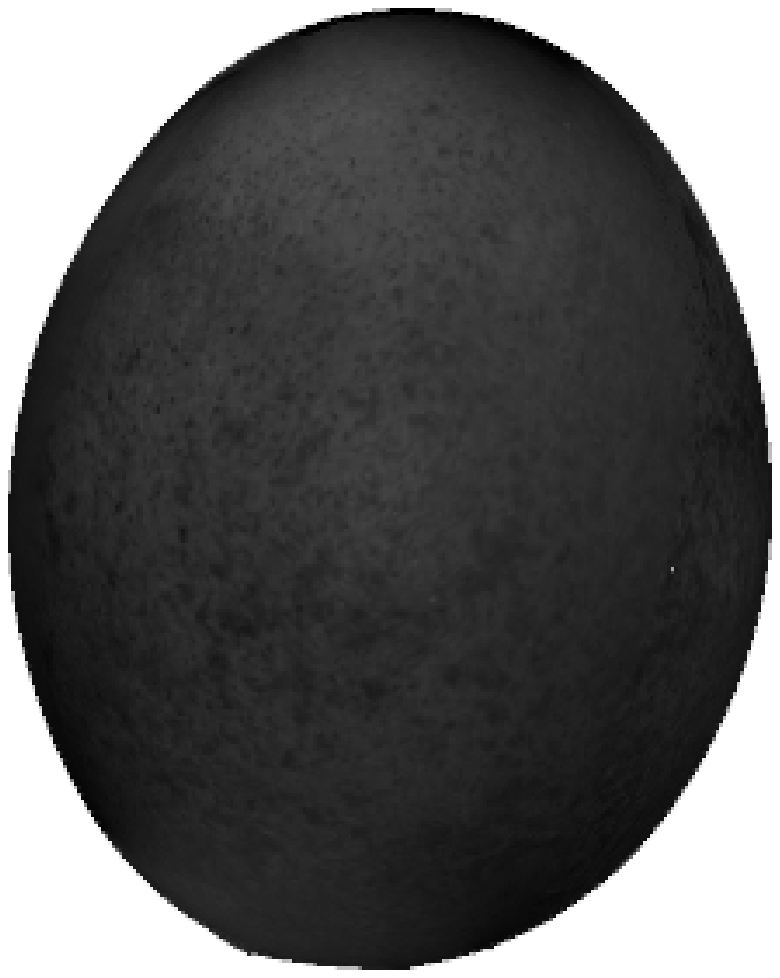} &
     \includegraphics[width=0.06\hsize]{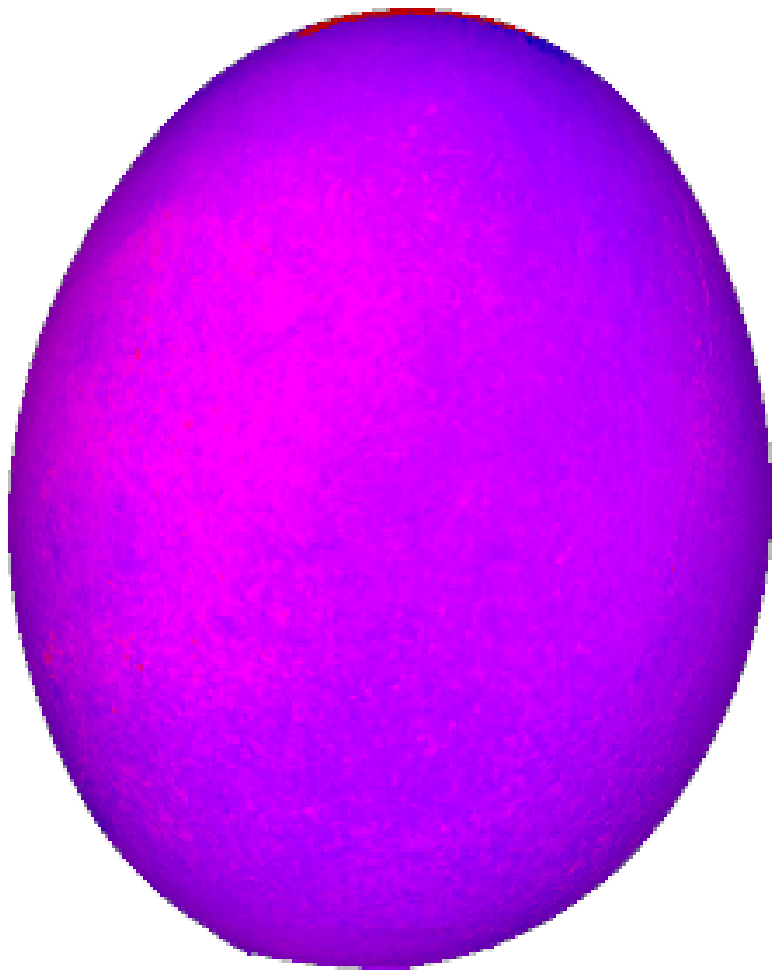} & 
     & \\
     
     & & \raisebox{1.25\normalbaselineskip}[0pt][0pt]{\rotatebox[origin=c]{90}{\footnotesize GT}} & 
     \includegraphics[width=0.06\hsize]{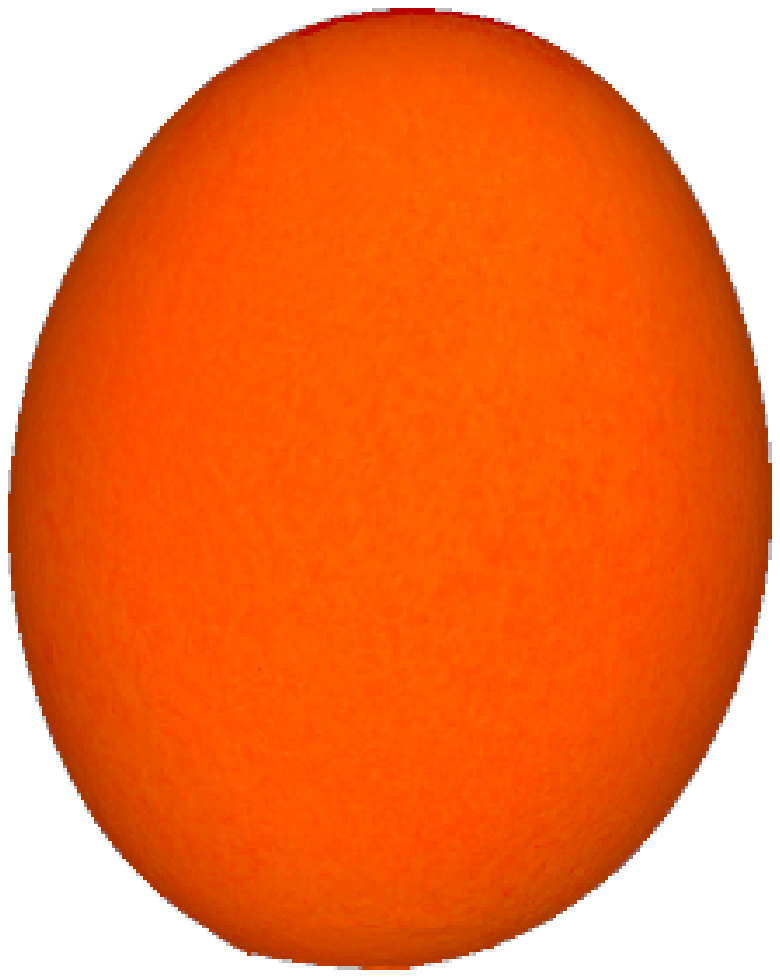} &
     \includegraphics[width=0.06\hsize]{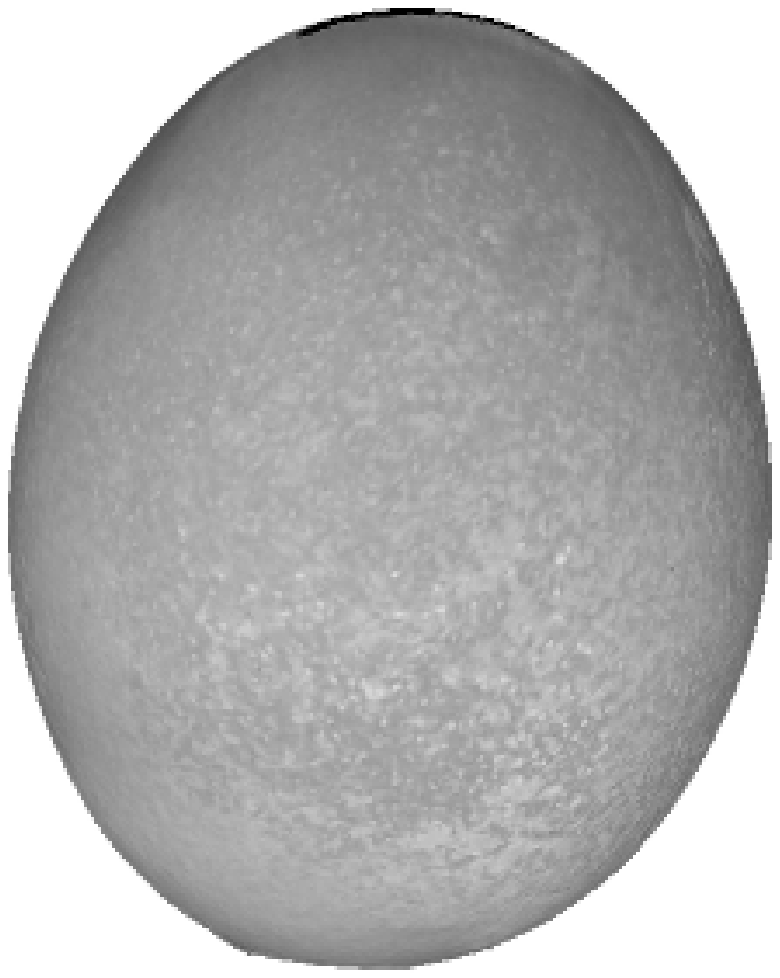} & &
     \raisebox{1.25\normalbaselineskip}[0pt][0pt]{\rotatebox[origin=c]{90}{\footnotesize source}} & 
     \includegraphics[width=0.06\hsize]{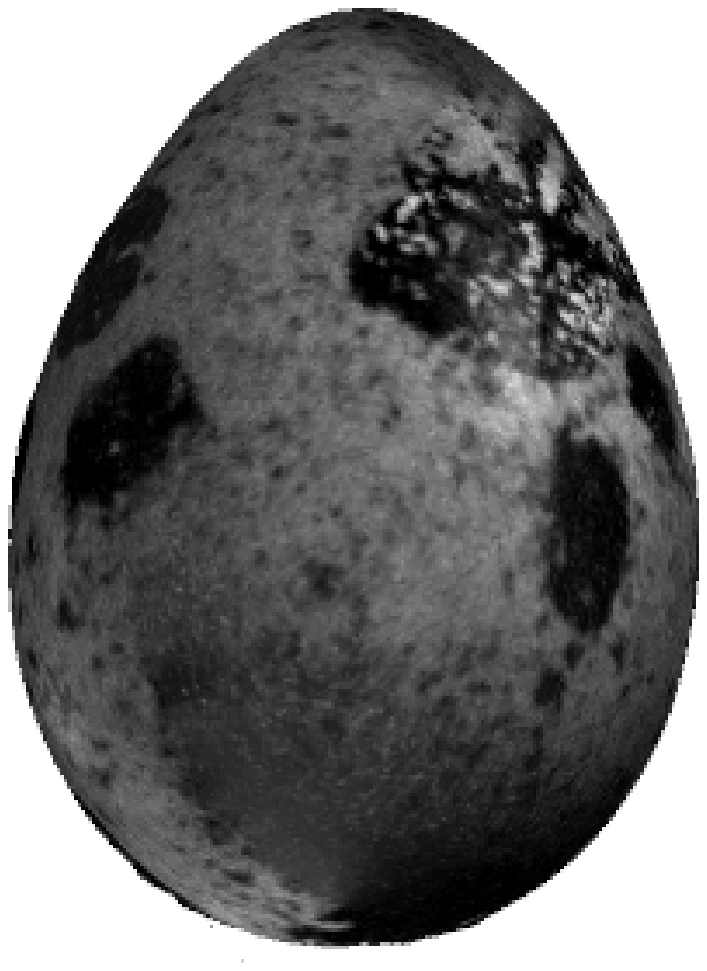} &
     \includegraphics[width=0.06\hsize]{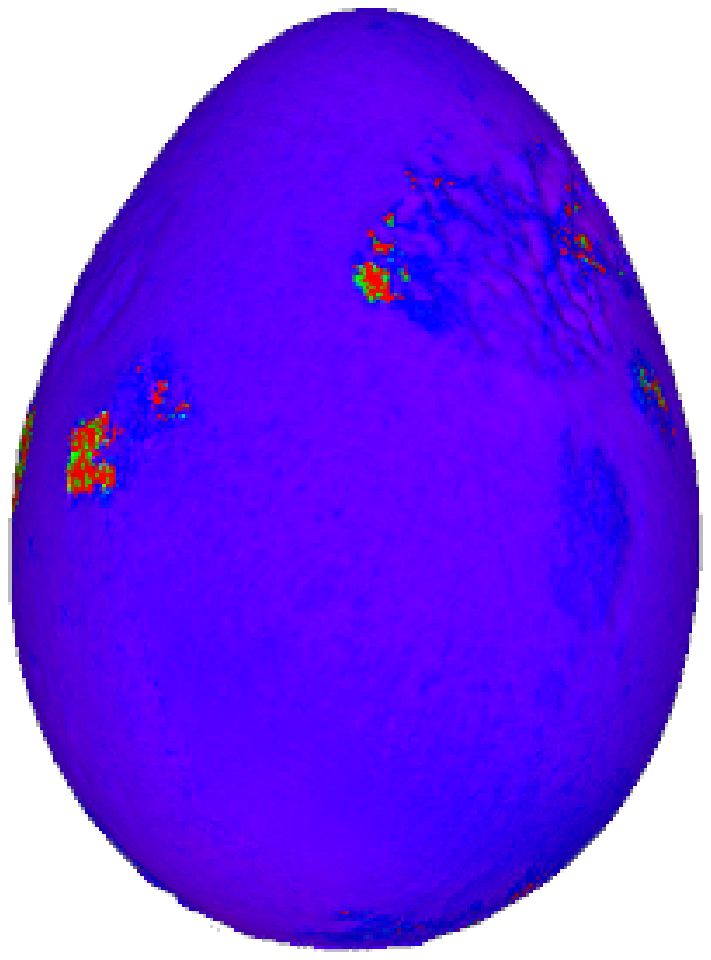} & 
     \includegraphics[width=0.06\hsize]{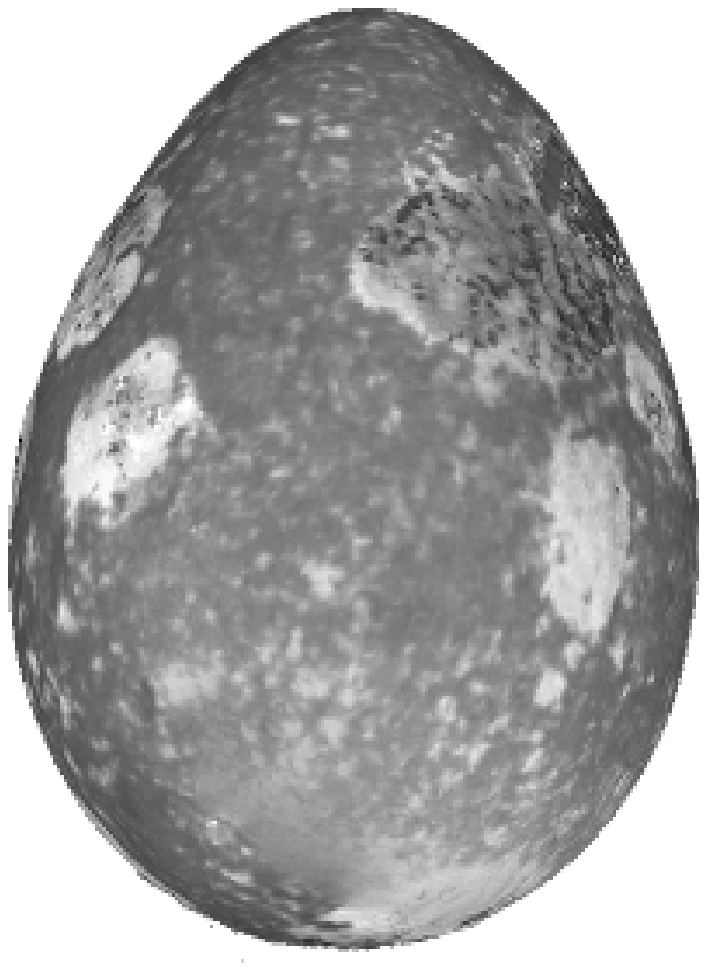} &
     \includegraphics[width=0.06\hsize]{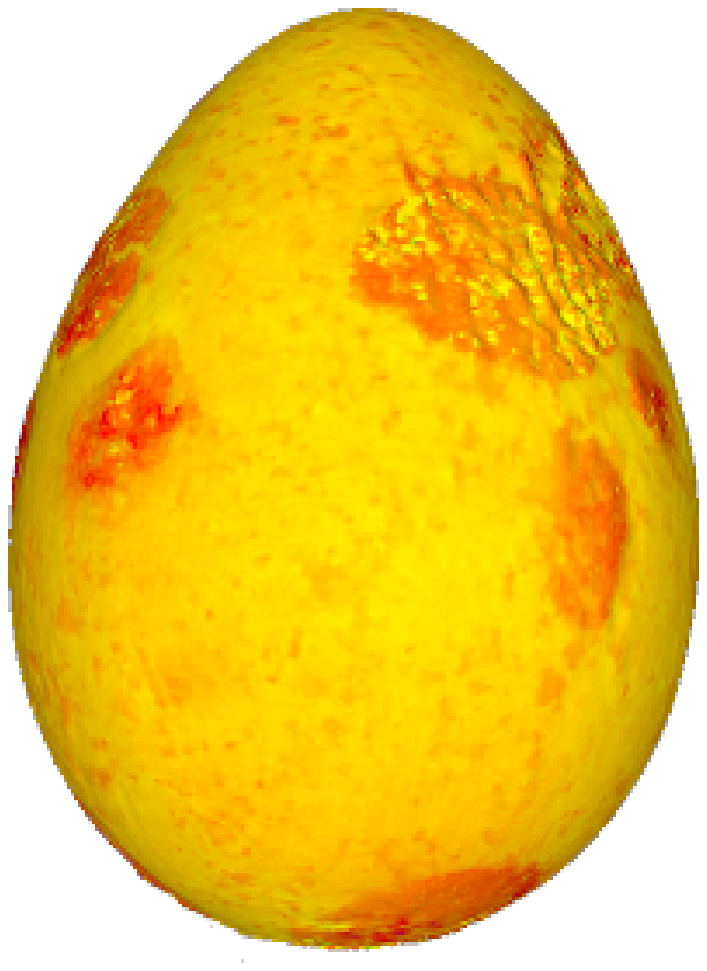} \\
     
     \multicolumn{2}{c}{\scriptsize Peahen} & & & & & & \footnotesize intensity & \footnotesize hue & \footnotesize saturation & \footnotesize hue \\
     
     & & & & & & & \multicolumn{2}{c}{\scriptsize UV (365nm)} & \multicolumn{2}{c}{\scriptsize visible (400nm-700nm)} \\
\end{tabular}

\caption{\label{fig:similar_shape_and_pattern_test}%
Evaluating different shape and patterns from a single exemplar (Test 1-5).}
\end{figure*}

%% file: fig-alleggs.tex
\begin{figure*}[h]
\centering
\begin{tabular}{c c c c c c c c c c c c}
\raisebox{1.5\normalbaselineskip}[0pt][0pt]{\rotatebox[origin=c]{90}{\footnotesize reconstructed}} 
& \includegraphics[width=0.06\hsize]{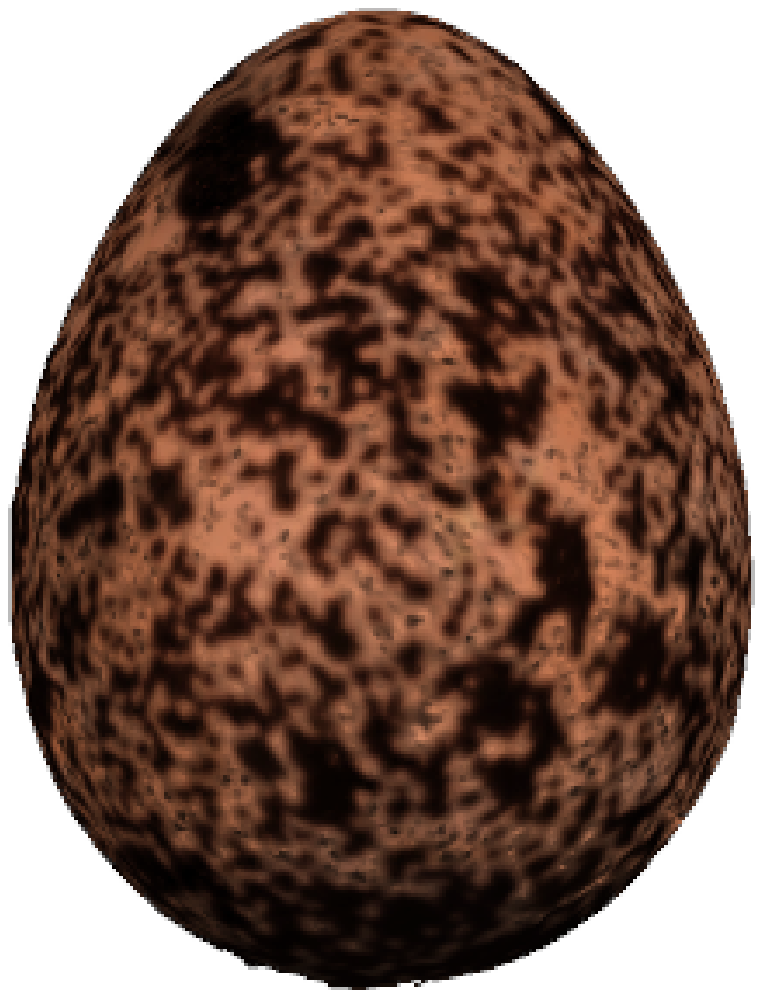} 
& \includegraphics[width=0.06\hsize]{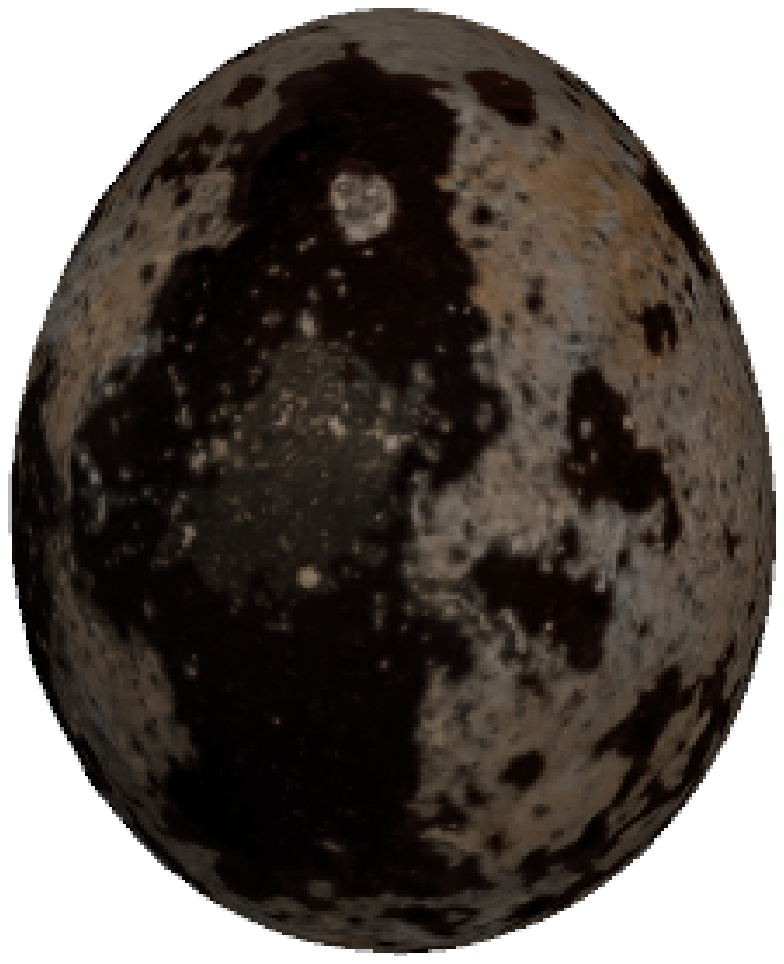} 
& \includegraphics[width=0.06\hsize]{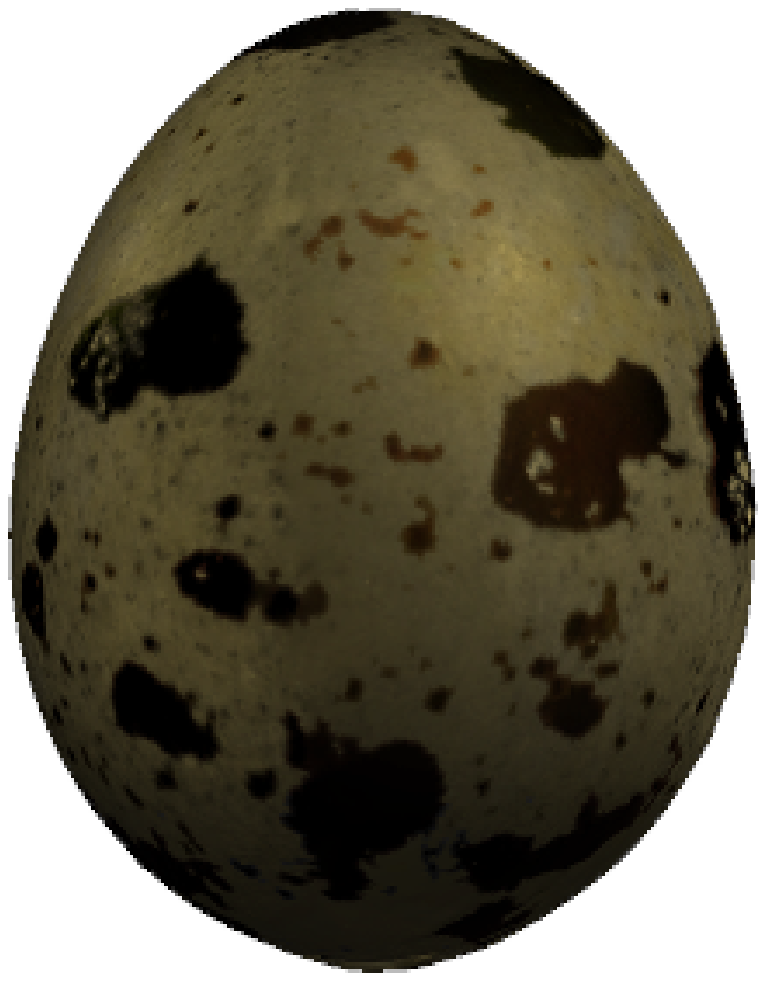} 
& \includegraphics[width=0.06\hsize]{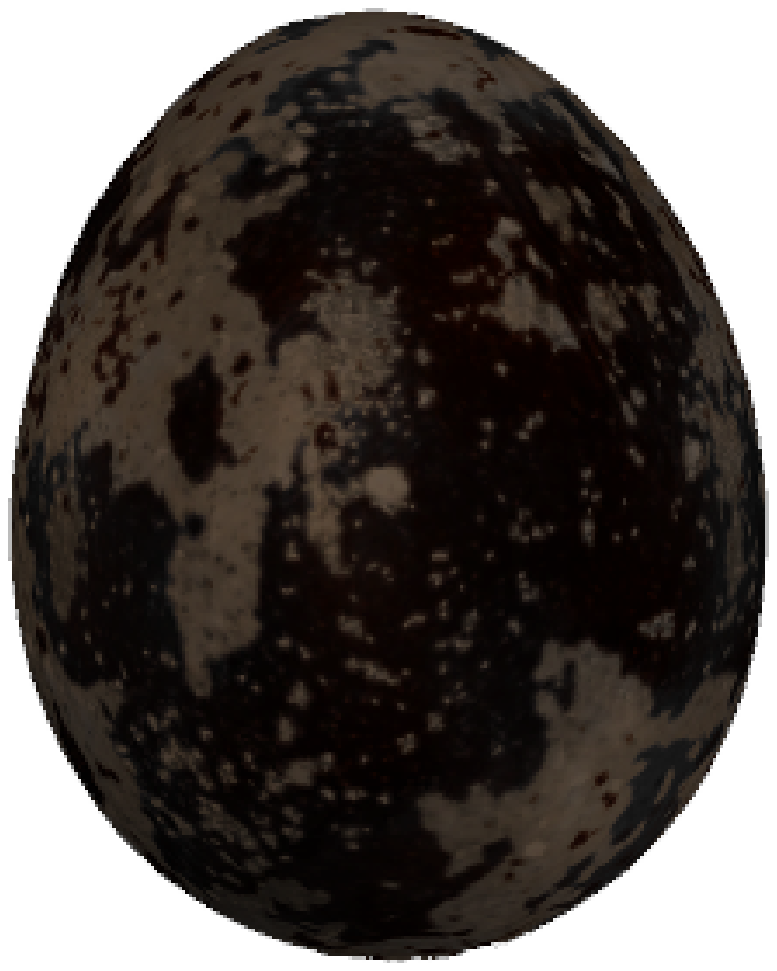} 
& \includegraphics[width=0.06\hsize]{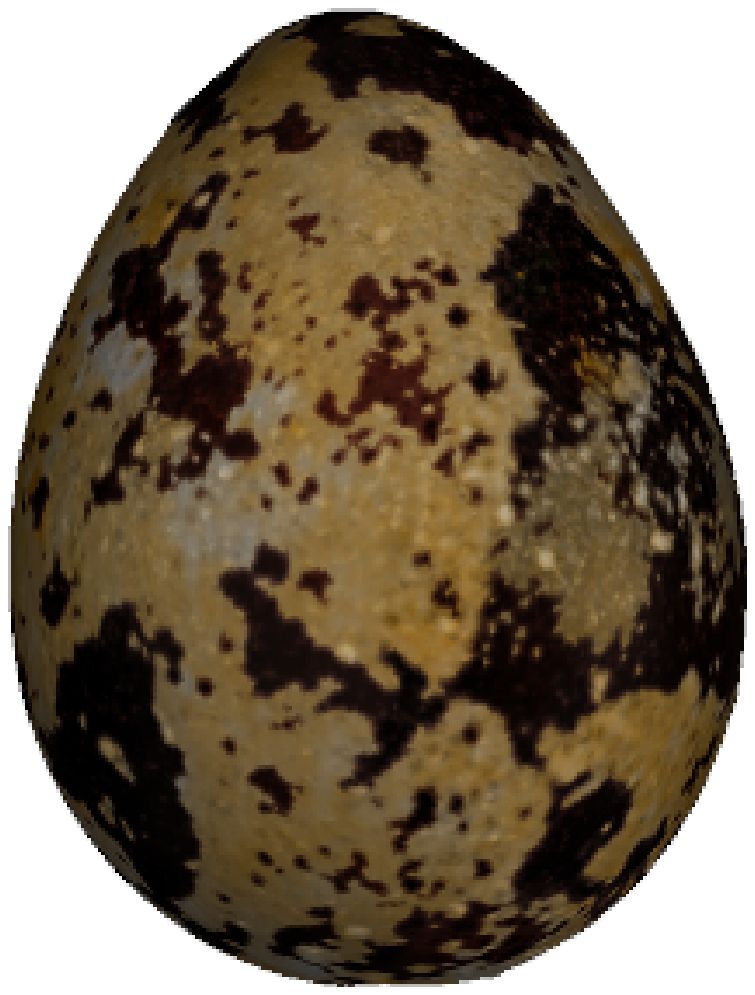}
& \includegraphics[width=0.06\hsize]{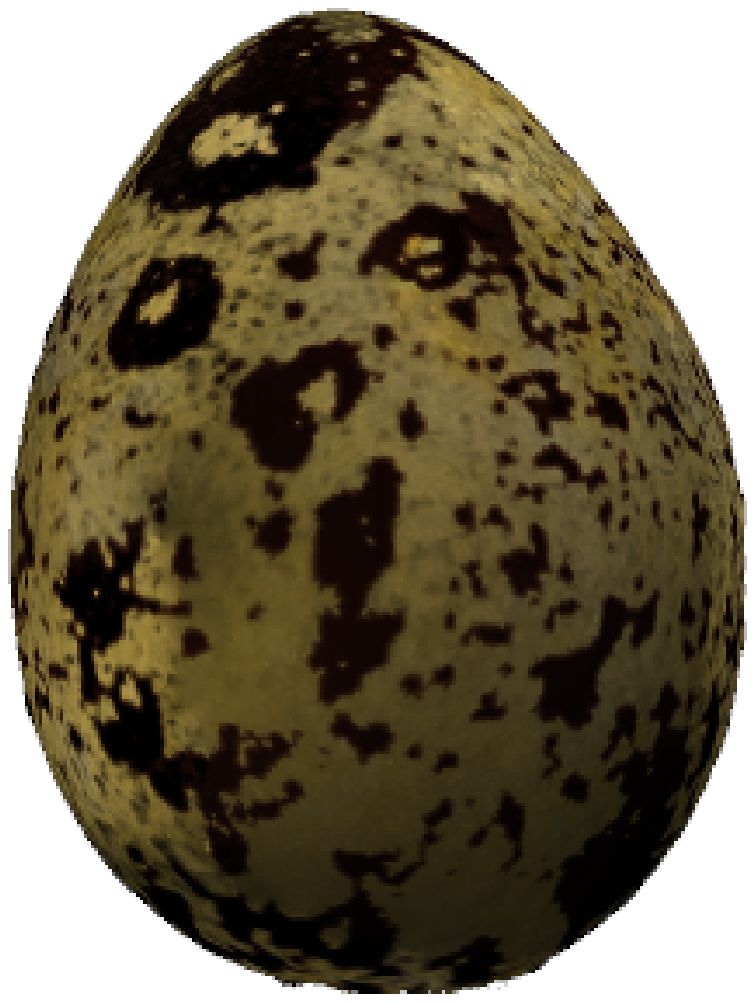}
& \includegraphics[width=0.06\hsize]{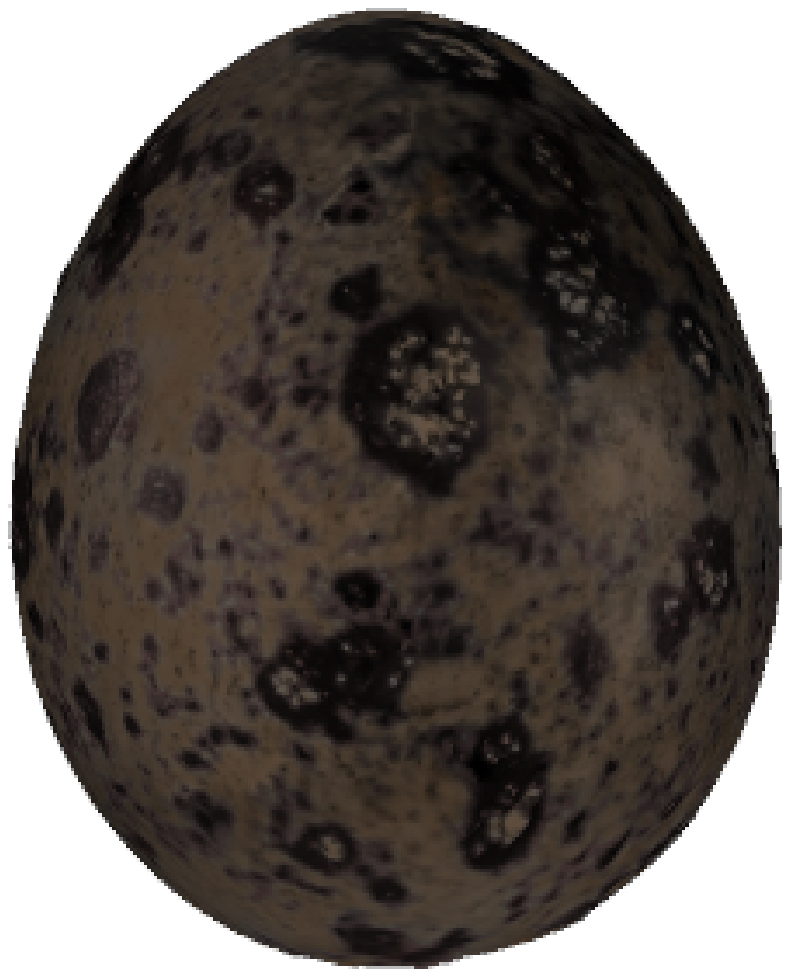}
& \includegraphics[width=0.06\hsize]{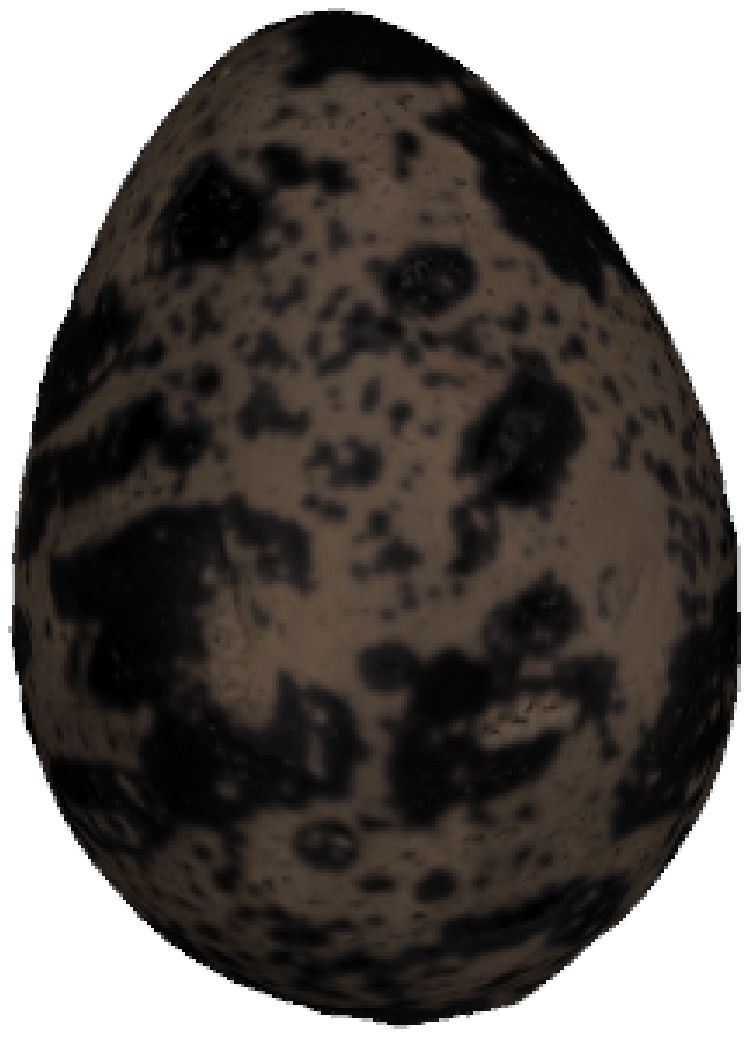}
& \includegraphics[width=0.06\hsize]{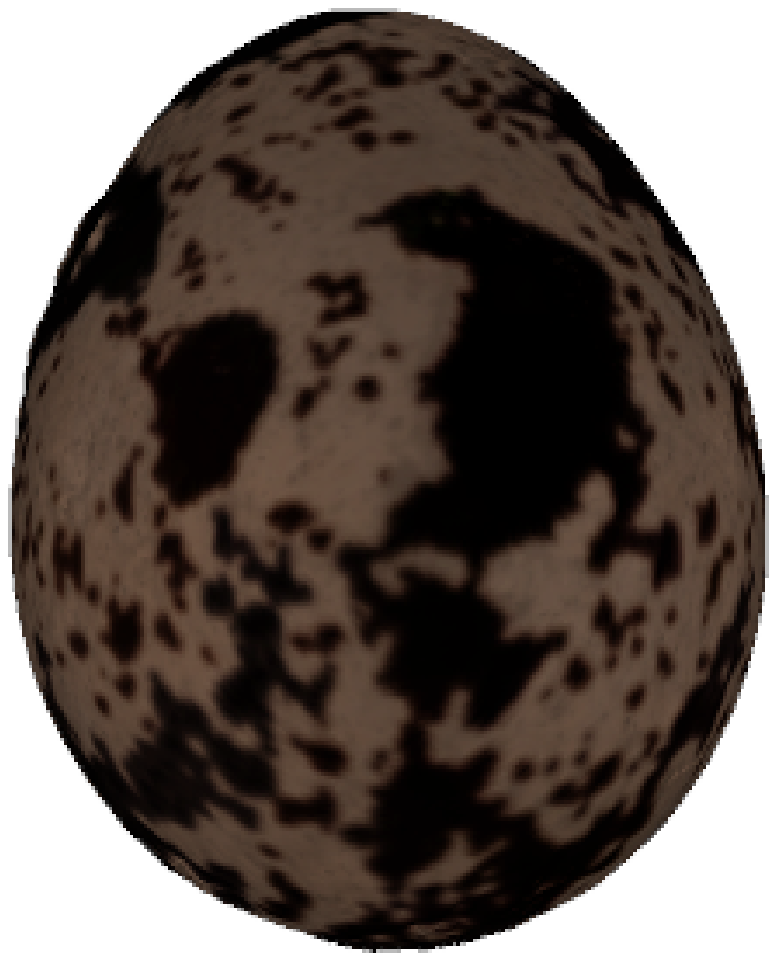}
& \includegraphics[width=0.06\hsize]{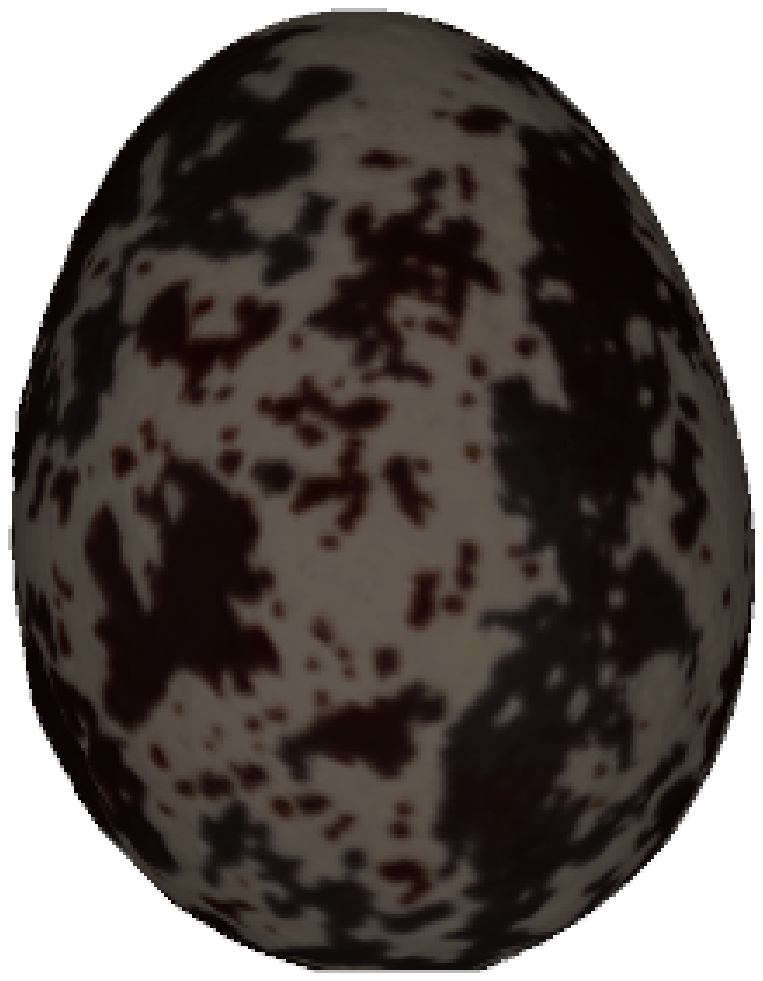}
& \includegraphics[width=0.06\hsize]{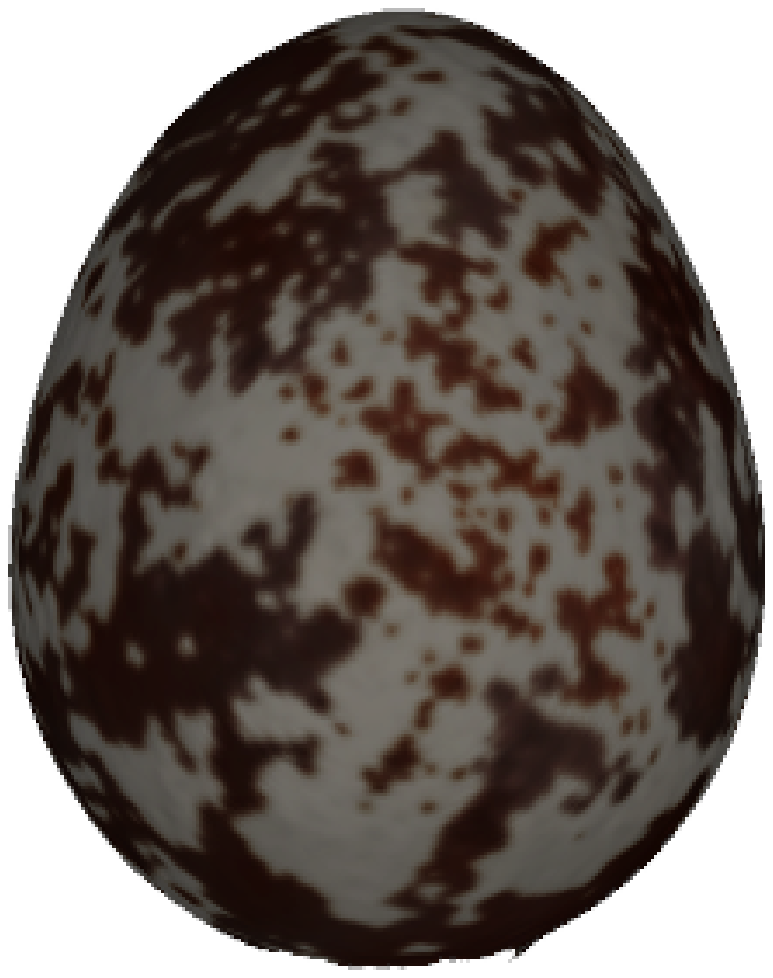}
\\

\raisebox{1.5\normalbaselineskip}[0pt][0pt]{\rotatebox[origin=c]{90}{\footnotesize GT}} 
& \includegraphics[width=0.06\hsize]{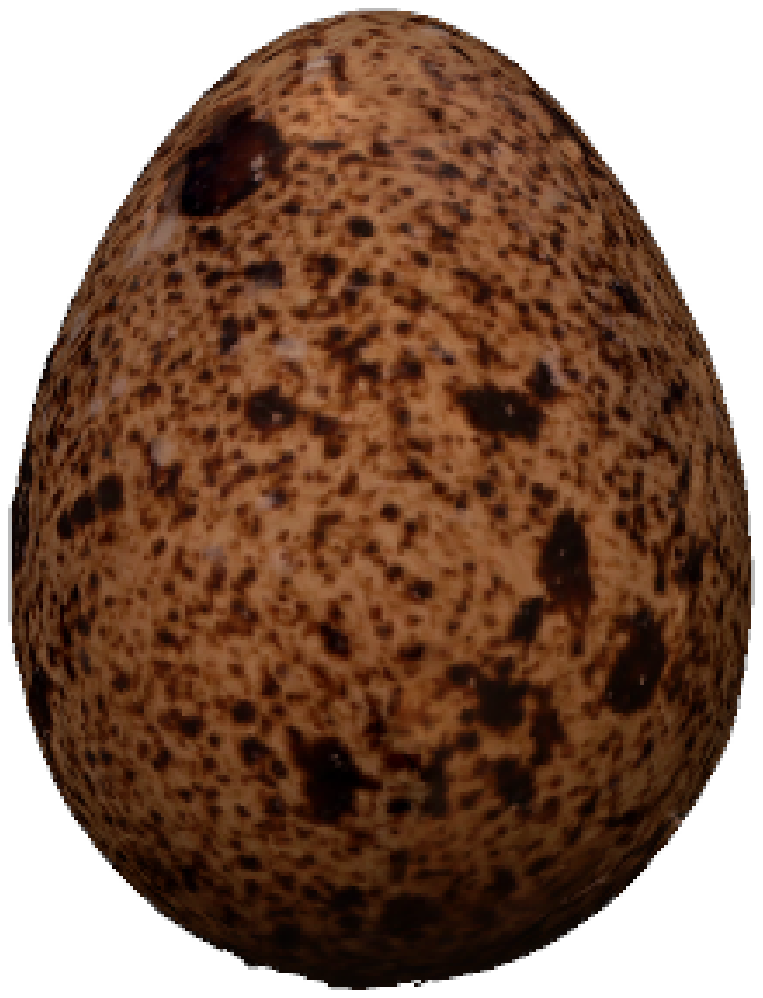} 
& \includegraphics[width=0.06\hsize]{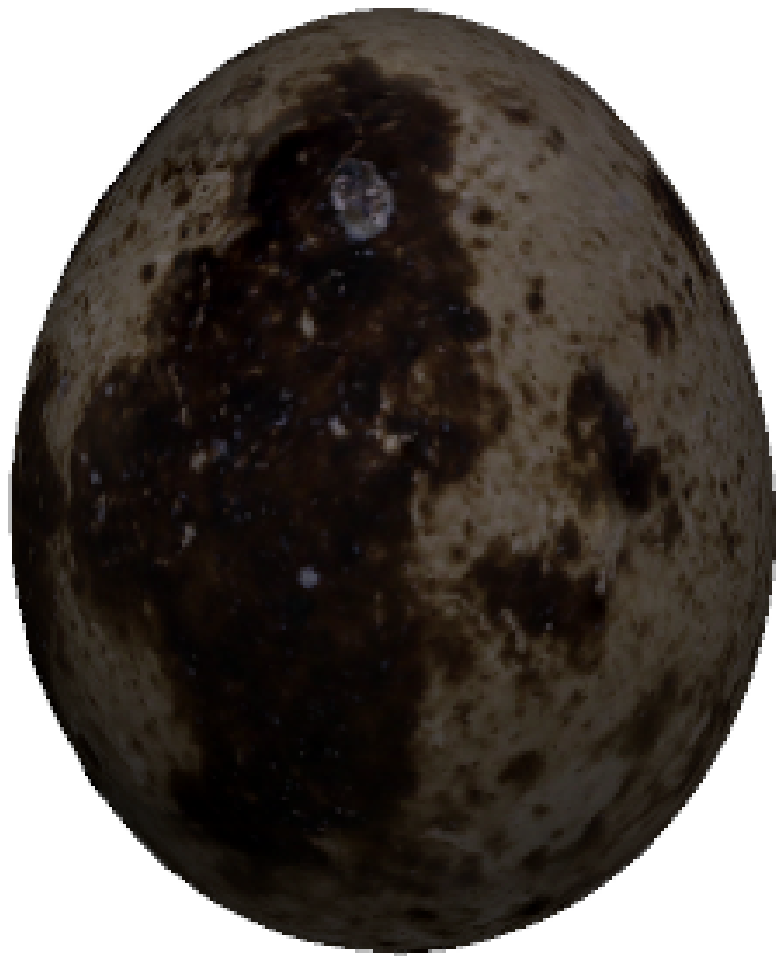} 
& \includegraphics[width=0.06\hsize]{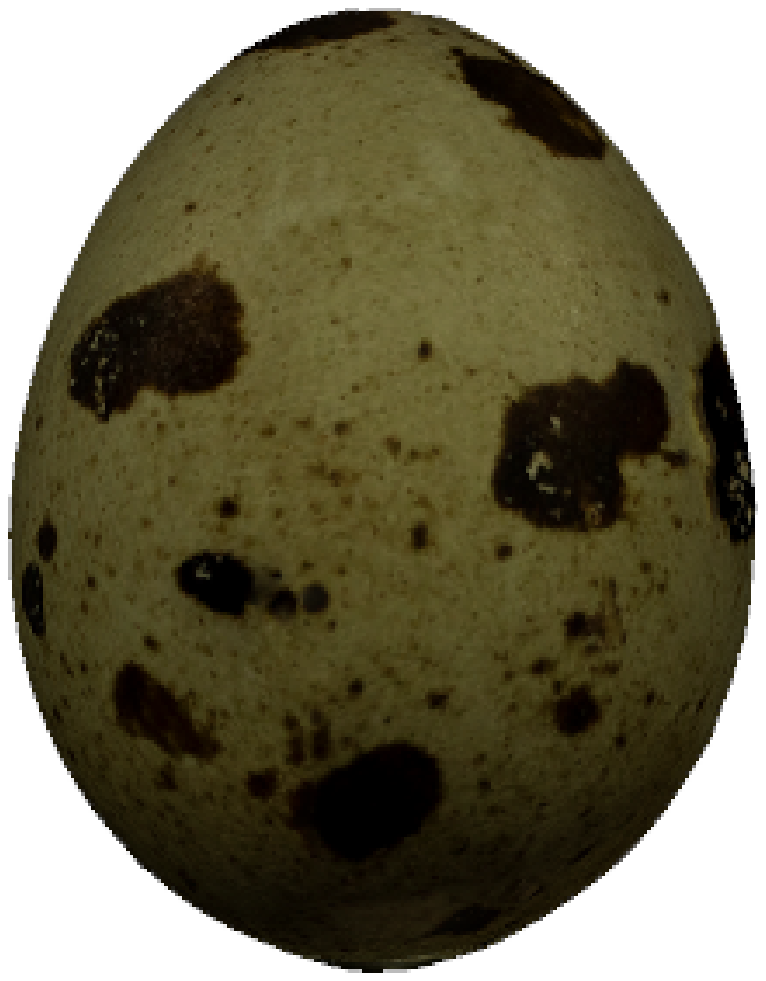} 
& \includegraphics[width=0.06\hsize]{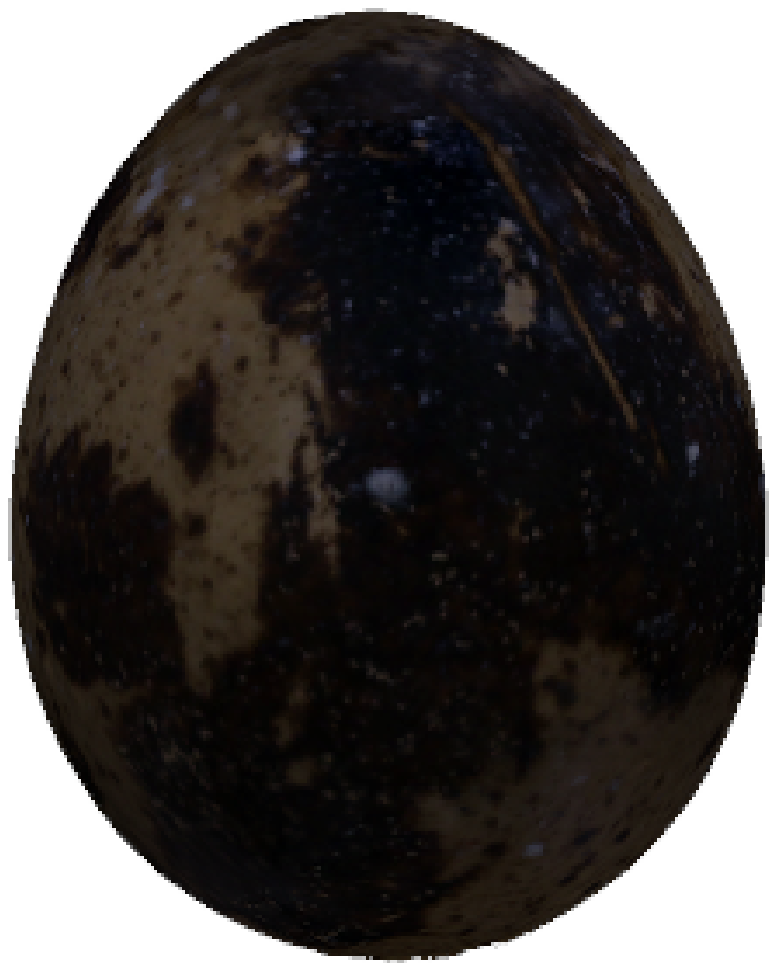} 
& \includegraphics[width=0.06\hsize]{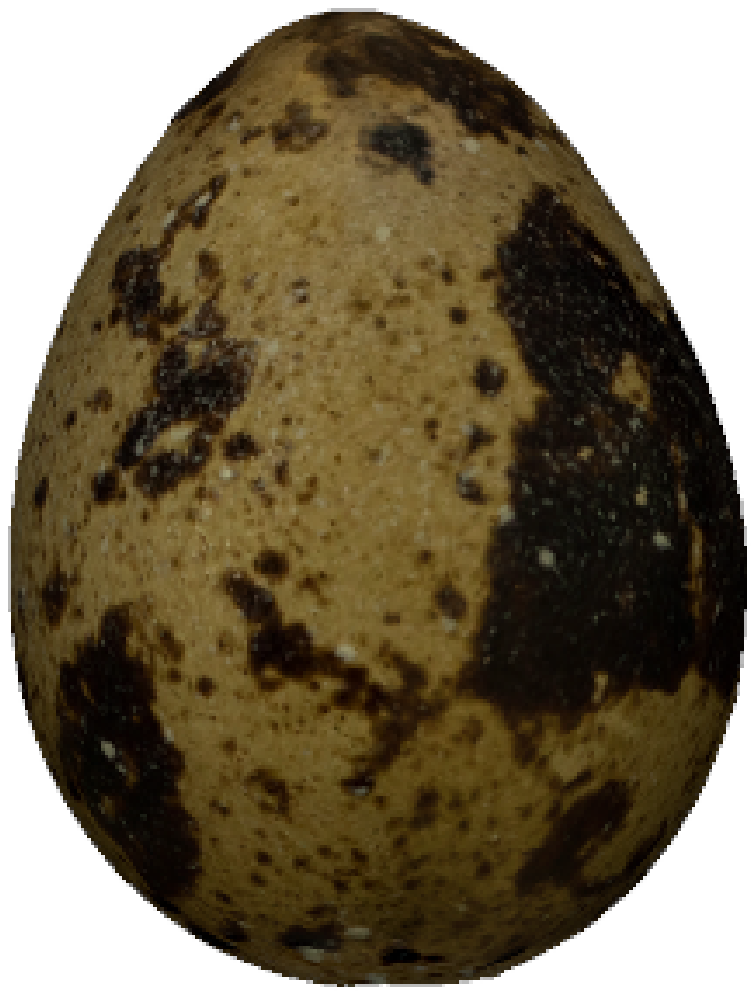}
& \includegraphics[width=0.06\hsize]{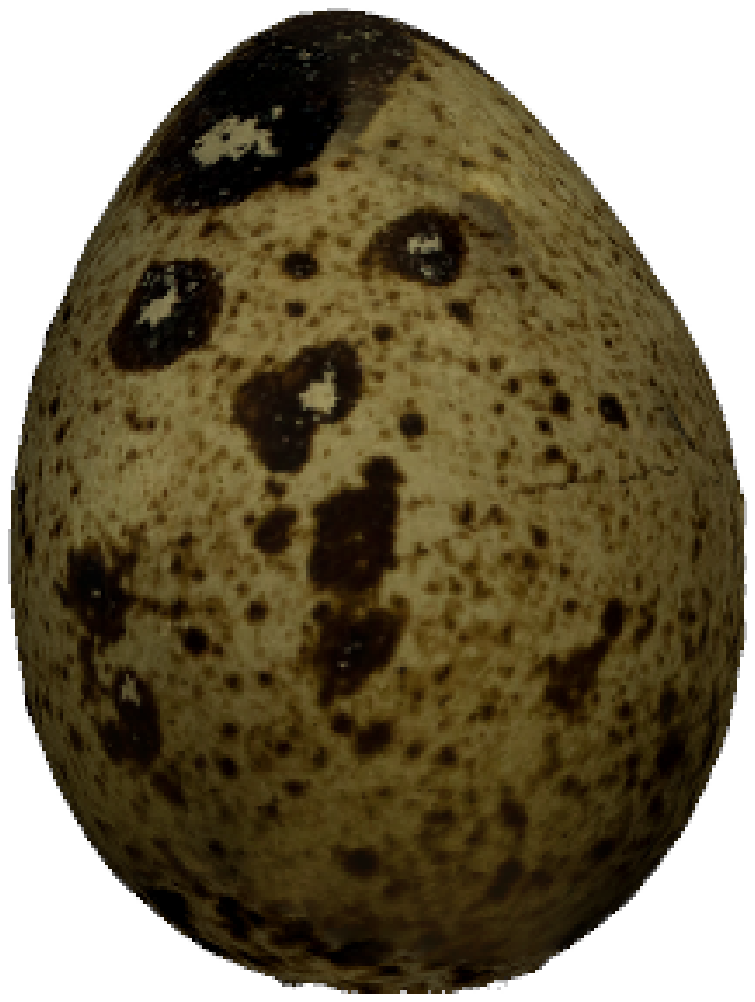}
& \includegraphics[width=0.06\hsize]{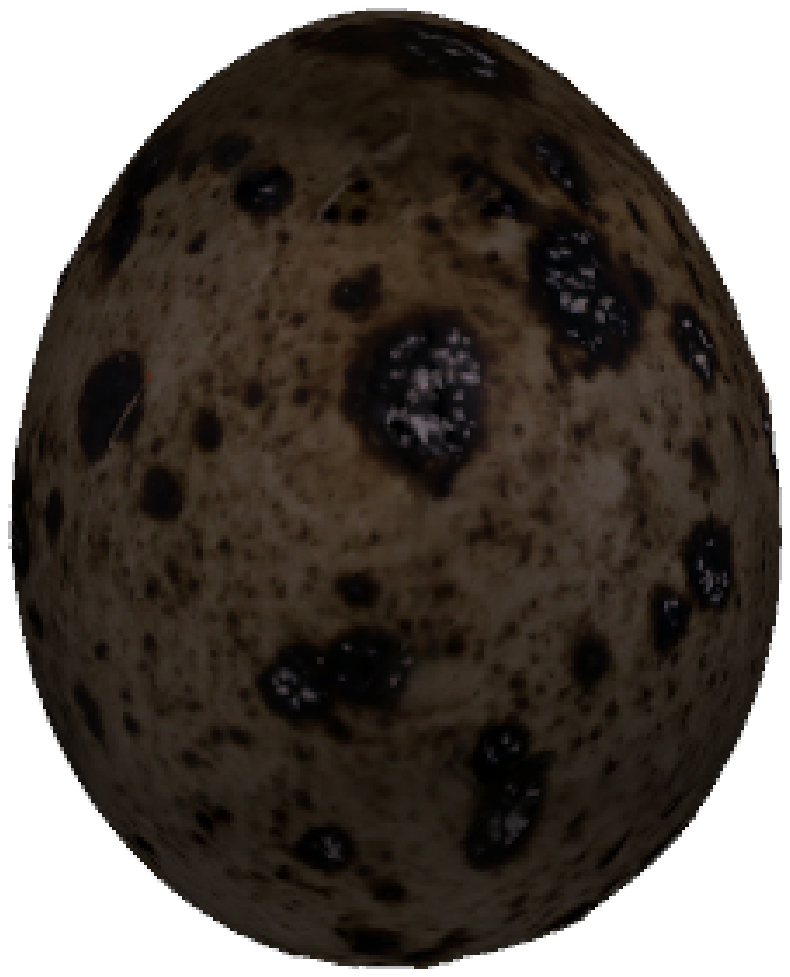}
& \includegraphics[width=0.06\hsize]{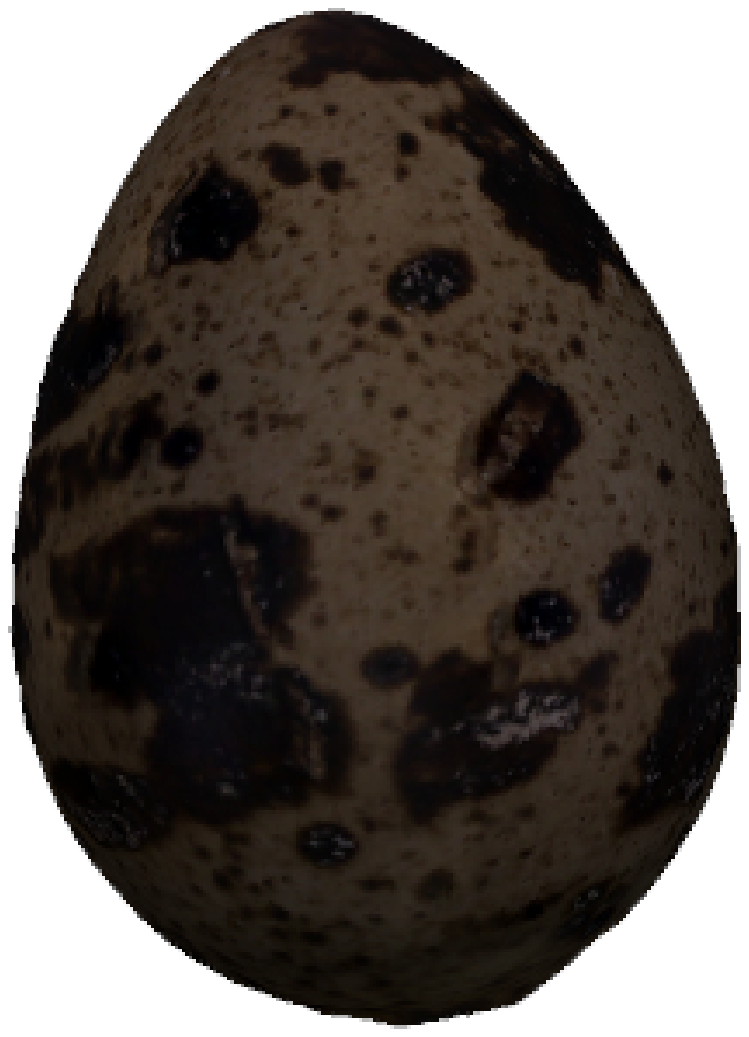}
& \includegraphics[width=0.06\hsize]{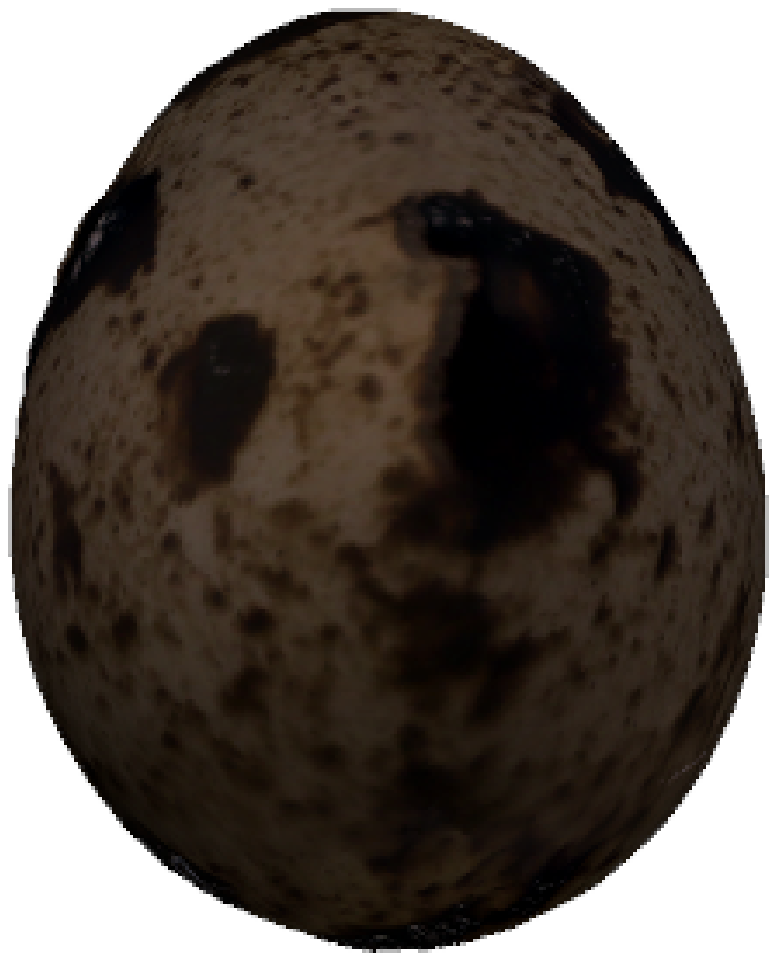}
& \includegraphics[width=0.06\hsize]{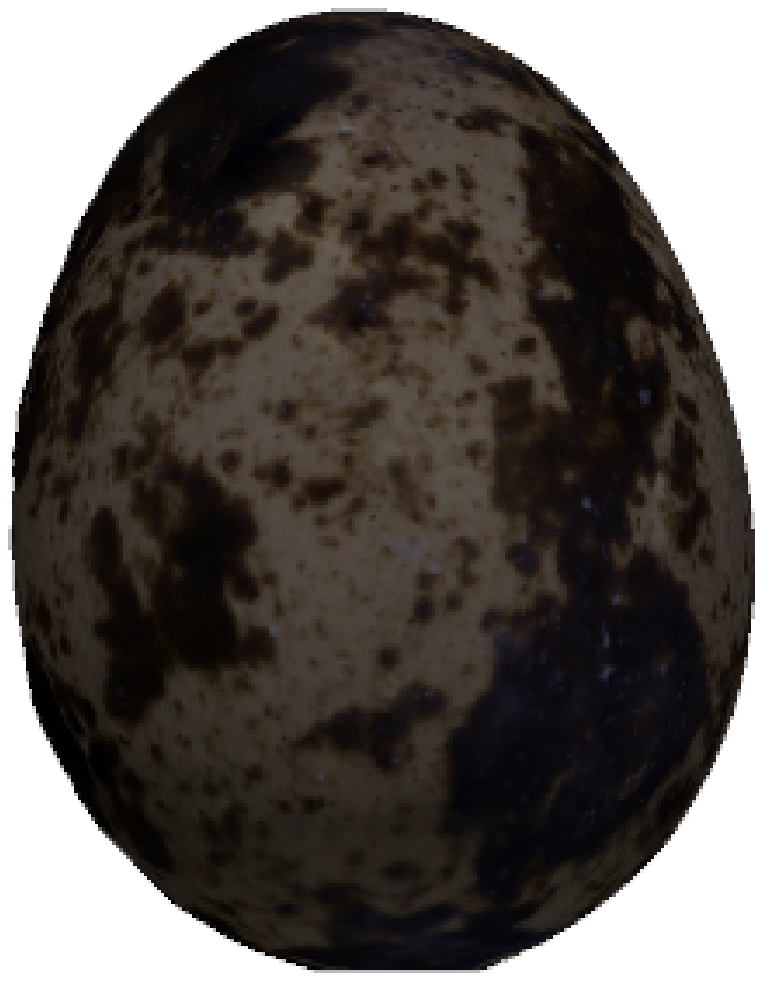}
& \includegraphics[width=0.06\hsize]{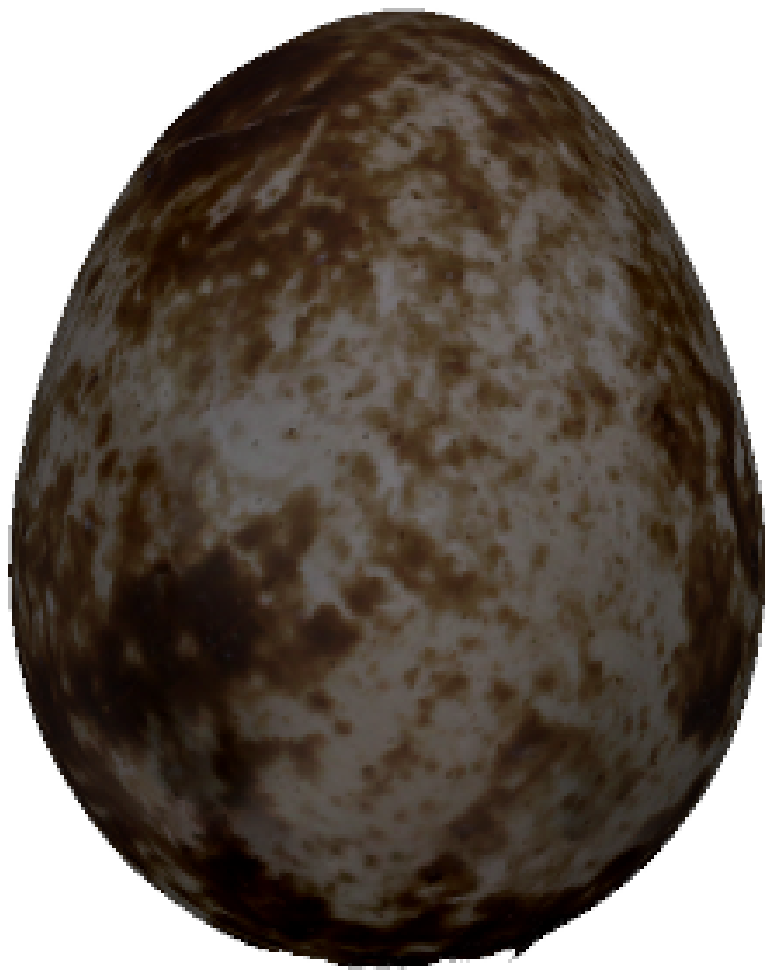}
\\

\end{tabular}
\caption{\label{fig:all_eggs}%
Evaluating different shape and patterns from a single exemplar (Test 6).}
\end{figure*}

%% file: fig-eval_graph.tex
\begin{figure}[h]
  \centering
    \includegraphics[width=0.95\linewidth]{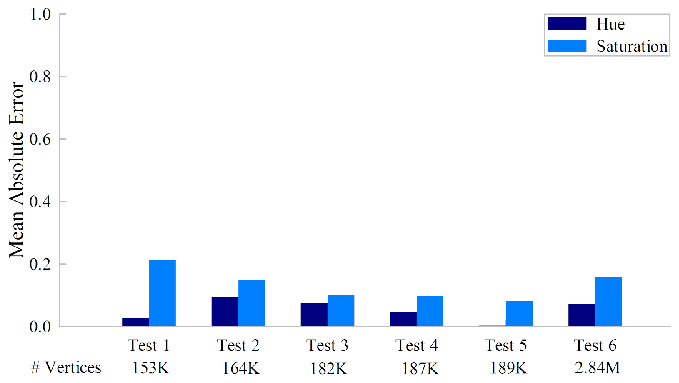}
    \caption[Evaluation Graph]{Quantitative evaluation shows that our algorithm can effectively reconstruct saturation and hue of patterned objects.}
	\label{fig:eval_graph}
\end{figure}

%% file: fig-userstudy.tex
\begin{figure}[t]
  \centering
    \includegraphics[width=0.95\linewidth]{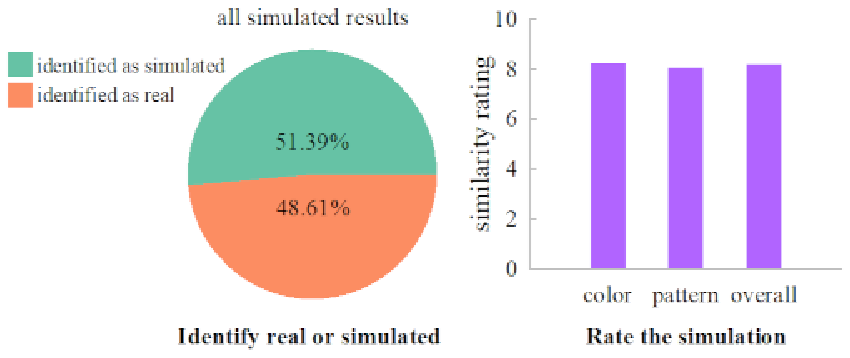}
    \caption[User study results]{(left) Participants identified $48.61\%$ of simulated results as real. (right) Users similarity ratings for color, patterns and overall appearance of simulated eggshells compared to ground truth on a scale of 1-10.}
	\label{fig:userstudy}
\end{figure}

%% file: fig-userstudy_results.tex
\begin{table}[t]
\centering
\setlength\tabcolsep{1.5pt}
\begin{tabular}{p{0.15\linewidth}|p{0.15\linewidth}|p{0.15\linewidth}|p{0.15\linewidth}|p{0.15\linewidth}|p{0.15\linewidth}}
\hline
\multirow{2}{*}{\small Tests} & \multicolumn{2}{c|}{\small Identification Results} & \multicolumn{3}{c}{\small Simulation Ratings} \\
\cline{2-6}
& \small Real & \small Simulated & \small Color & \small Patterns & \small Overall \\
\hline
\small Test 1  & \small \textbf{77.78\%} & \small 22.22\% & \small 8.00 & \small \textbf{9.60} & \small 8.96\\
\small Test 2 & \small 0\% & \small 100\% & \small 6.00 & \small 6.95 & \small 6.90\\
\small Test 3  & \small 44.44\% & \small 55.56\% & \small 7.10 & \small 6.90 & \small 7.15\\
\small Test 4 & \small \textbf{66.67\%} & \small 33.33\% & \small \textbf{9.20} & \small 7.90 & \small 8.40\\
\small Test 5 & \small 33.33\% & \small 66.67\% & \small 8.90 & \small 4.90 & \small 6.80\\ 
\hline
\end{tabular}
\caption{\label{fig:userstudy_results}%
User study results for the simulated eggshells from Tests 1-5.}
\end{table}

%% file: fig-experiment_results.tex
\begin{table}[h]
\centering
\setlength\tabcolsep{1.5pt}
\begin{tabular}{p{0.775\linewidth}|p{0.075\linewidth}|p{0.15\linewidth}}
\hline
\multirow{2}{*}{\small Experiments} & \multicolumn{2}{c}{\small Average Accuracy} \\
\cline{2-3}
& \small Hue & \small Saturation \\
\hline
\small Different Shape and Patterns from a Single Exemplar  & \small 92\% & \small 85\% \\
\small Evaluating Fluorescent and Non-Fluorescent Materials  & \small 82\% & \small 79\% \\ 
\hline
\end{tabular}
\caption{\label{fig:experiment_results}%
Summary of experimental results.}
\end{table}

%% file: fig-comparison.tex
\begin{figure}[h]
\centering
\includegraphics[width=0.7\hsize]{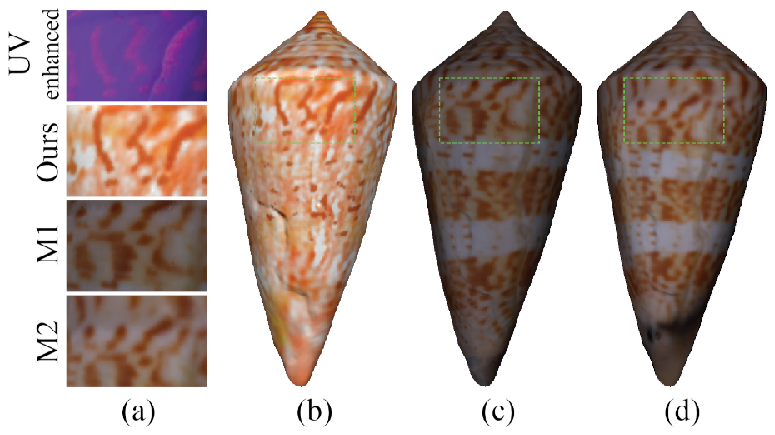}
\caption{\label{fig:comparison}%
(a) Pattern closeup (b) Our result (c) Geomagic (M1) (d) Conformal Surface Parameterization~\cite{haker} (M2). \emph{Conus delesertti} fossil, UF117269, FLMNH.}
\end{figure}

%% file: fig-mineral.tex
\begin{figure*}[h]
\centering
\includegraphics[width=0.8\linewidth]{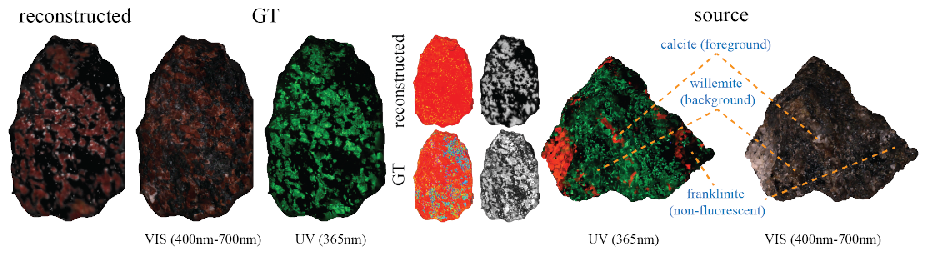}
\caption{\label{fig:mineral}%
Evaluating fluorescent and non-fluorescent materials. Mineral composition: willemite (brick-red), calcite (white), franklinite (black).}
\end{figure*}

%% file: applications.tex
\section{Applications}
\label{sec:applications}

There are opportunities to leverage our style transfer methods for digital restoration of aged or damaged materials at natural history (iDigBio~\cite{page2015}) and cultural heritage institutions. Texture synthesis for virtual environments is another promising area.

A unique application of our work reconstructs color in extinct shell fossils (Figure~\ref{fig:reconstruction}). The hue of cone shells is caused by biological processes that produce chemical secretions on the surface~\cite{williams2017}. Researchers in paleobiology illuminate these  shell fossils with ultraviolet radiation at different illumination angles to reveal the original patterns that uniquely identify the shell species~\cite{Hendricks2015}. False color from photo-editing software is used to infill patterns. In our example, the target dates back $3$ million years and the source is a modern descendant of the same species (Conus delliserti)~\cite{Hendricks2015,williams2017}. We transfer \hbox{3-D} color patterns from a source to a target object even when there are non-corresponding color variations.  Although we have no ground truth to compare our results (our target is extinct), our test evaluation with ground truth data on the avion eggshells and minerals provide evidence of the efficacy of our color reconstruction. 

Figure~\ref{fig:restoration} illustrates an application that restores faded color in painted patterns (flower petals) on broken tiles using a pristine exemplar from the same collection. Despite differences in shape and scale, we adapt the source color to the measured target materials effectively restoring the faded region. 

We can give the synthetic armadillo model in Figure~\ref{fig:material_assignment} the material appearance properties of willemite or a valley quail eggshell. In a process similar to \hbox{3-D} texture painting, the artist paints the intensity map on the object to indicate the material composition and concentration distribution (Figure~\ref{fig:material_assignment} \emph{top right}). Our algorithm automatically separates foreground patches from the background, and appropriately generates appearance properties to match the assigned material distribution (Figure~\ref{fig:material_assignment} \emph{bottom row}).

\input{fig-reconstruction_v002}

\input{fig-restoration}
\input{fig-material_assignment}

%% file: fig-reconstruction_v002.tex
\begin{figure}[ht!]
\centering
\renewcommand{\arraystretch}{.75}
\begin{tabular}{@{\hspace{0.5\tabcolsep}} c @{\hspace{0.5\tabcolsep}} c @{\hspace{0.5\tabcolsep}} c @{\hspace{0.5\tabcolsep}} c }
	\multicolumn{2}{@{\hspace{0.5\tabcolsep}} c}{\small fossil} & \small reconstructed & \small modern \\
	\scriptsize UV (365nm) & \scriptsize VIS (400nm-700nm) & & \\

	\includegraphics[width=0.175\hsize]{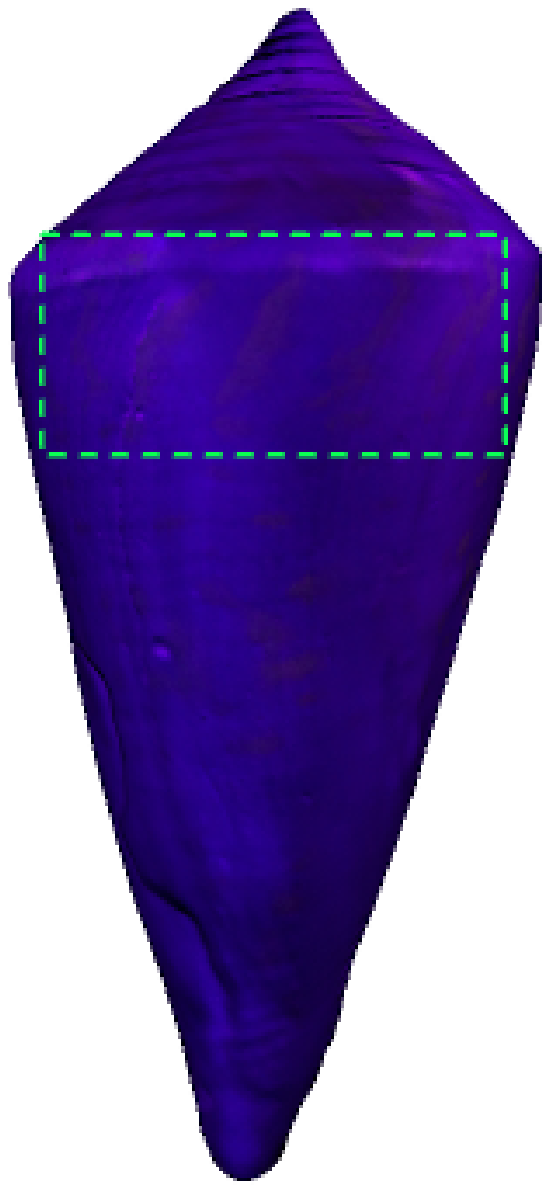} & 
    \includegraphics[width=0.175\hsize]{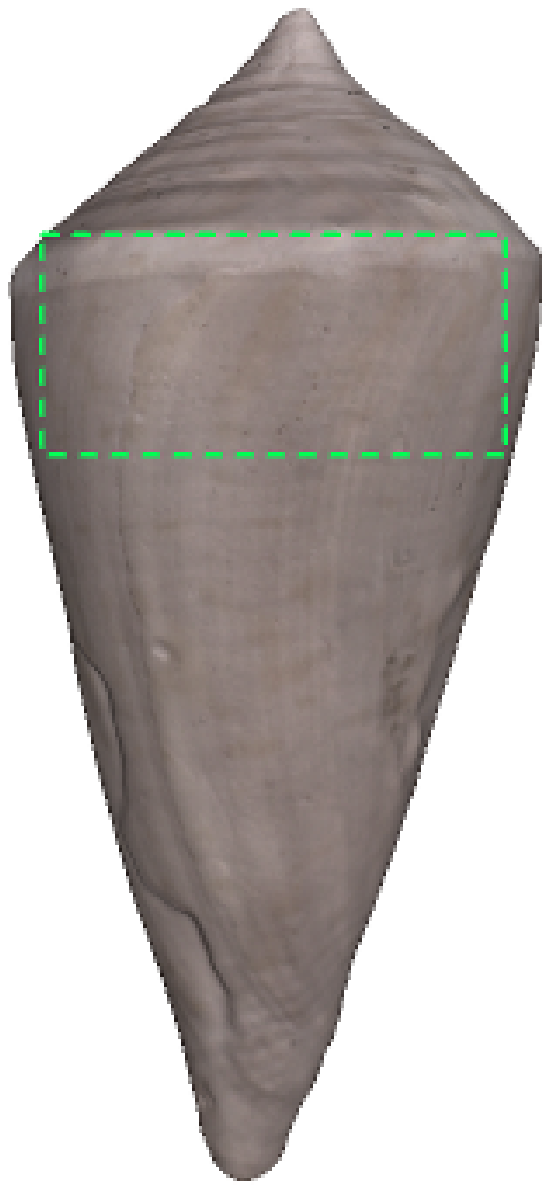} &
	\includegraphics[width=0.175\hsize]{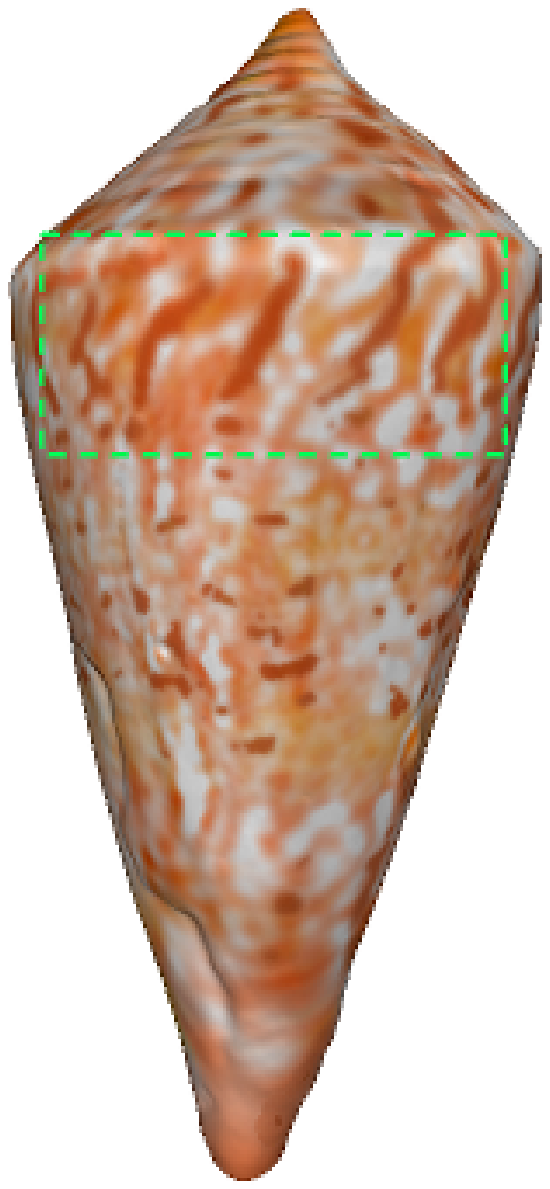} &	
	\includegraphics[width=0.175\hsize]{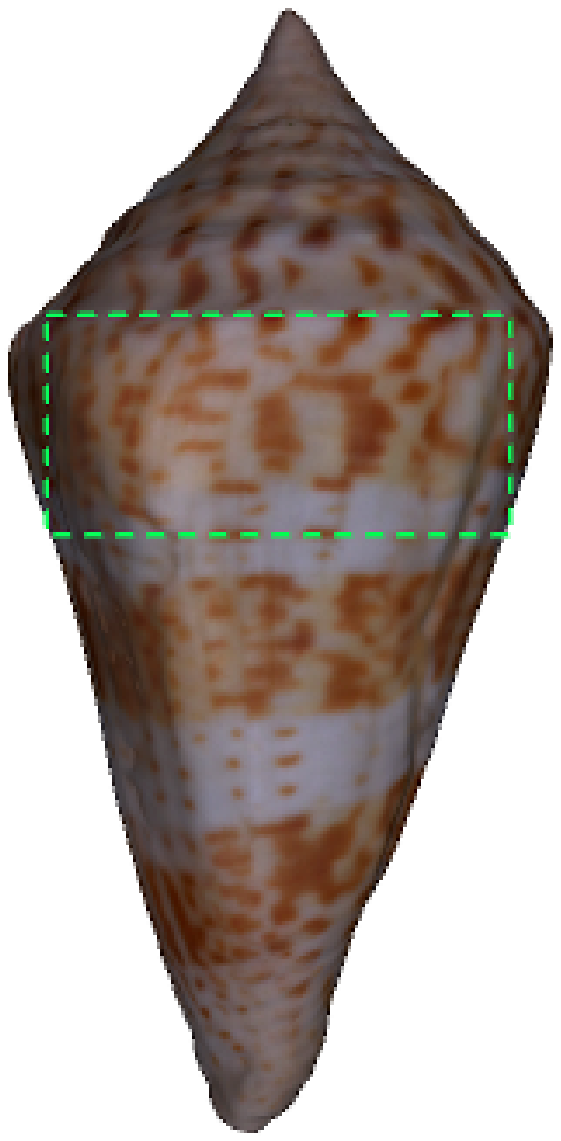} \\
	
	\includegraphics[width=0.24\linewidth]{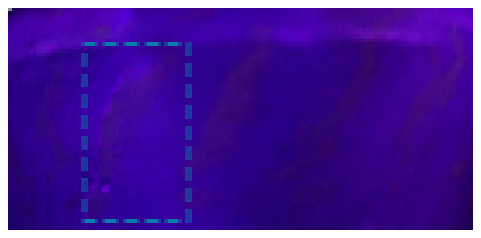} & 
    \includegraphics[width=0.24\linewidth]{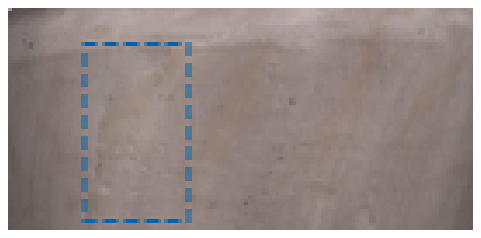} &
	\includegraphics[width=0.24\linewidth]{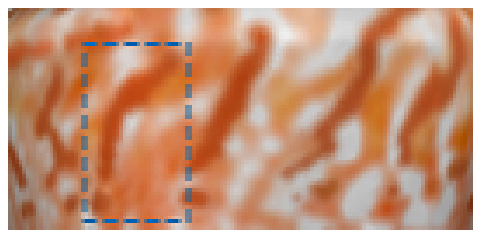} &	
	\includegraphics[width=0.24\linewidth]{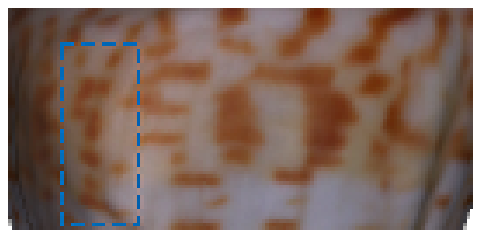} \\
	
	\includegraphics[width=0.225\hsize]{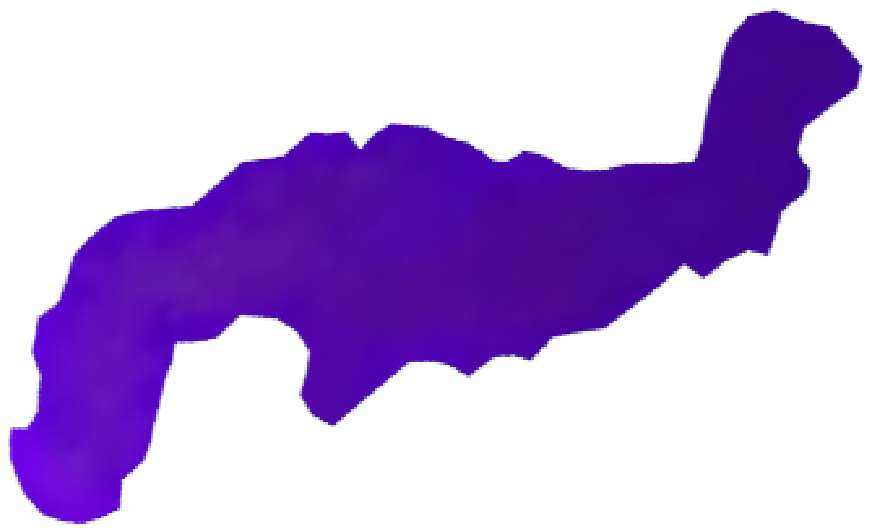} & 
    \includegraphics[width=0.225\hsize]{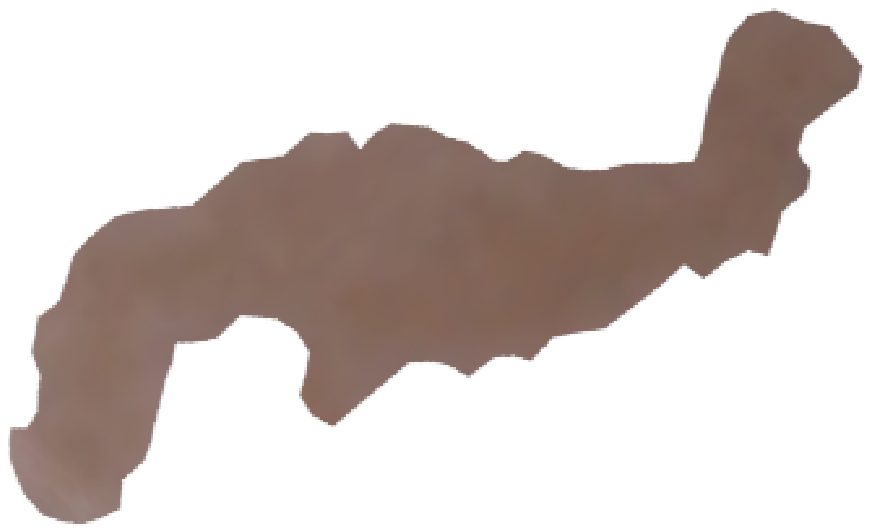} &
	\includegraphics[width=0.225\hsize]{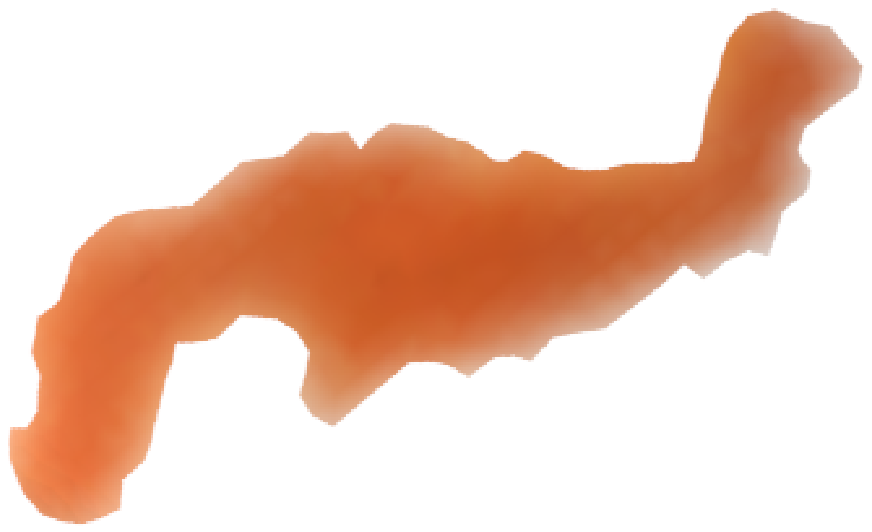} &	
	\includegraphics[width=0.225\hsize]{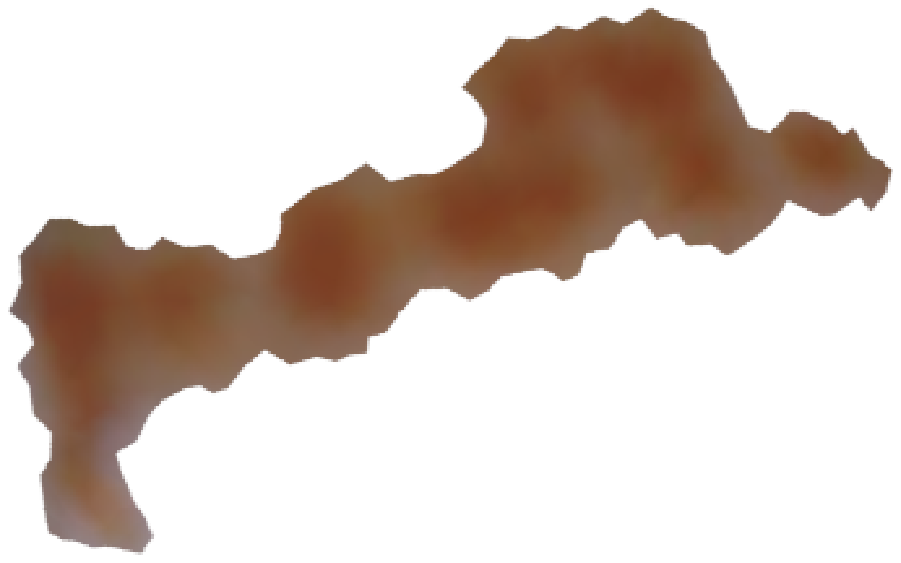} \\
\end{tabular}
\caption{\label{fig:reconstruction}%
Reconstructing color patterns in a colorless shell fossil. Fossil and modern shell patterns have similar shapes, but color variations within pattern boundaries differ. (bottom) We reconstruct color even when there are non-corresponding color variations. \emph{Conus delesertti} fossil, UF117269, \emph{Conus delesertti} modern, UF25623, FLMNH}
\end{figure}

%% file: fig-restoration.tex
\begin{figure}[h!]
\centering
\renewcommand{\arraystretch}{.75}
\begin{tabular}{@{\hspace{0.5\tabcolsep}} c @{\hspace{0.5\tabcolsep}} c @{\hspace{0.5\tabcolsep}} c @{\hspace{0.5\tabcolsep}} c }
	\multicolumn{2}{@{\hspace{0.5\tabcolsep}} c}{\small target} & \small reconstructed & \small source \\
	\scriptsize UV (254nm) & \scriptsize VIS (400nm-700nm) & & \\

	\includegraphics[width=0.225\hsize]{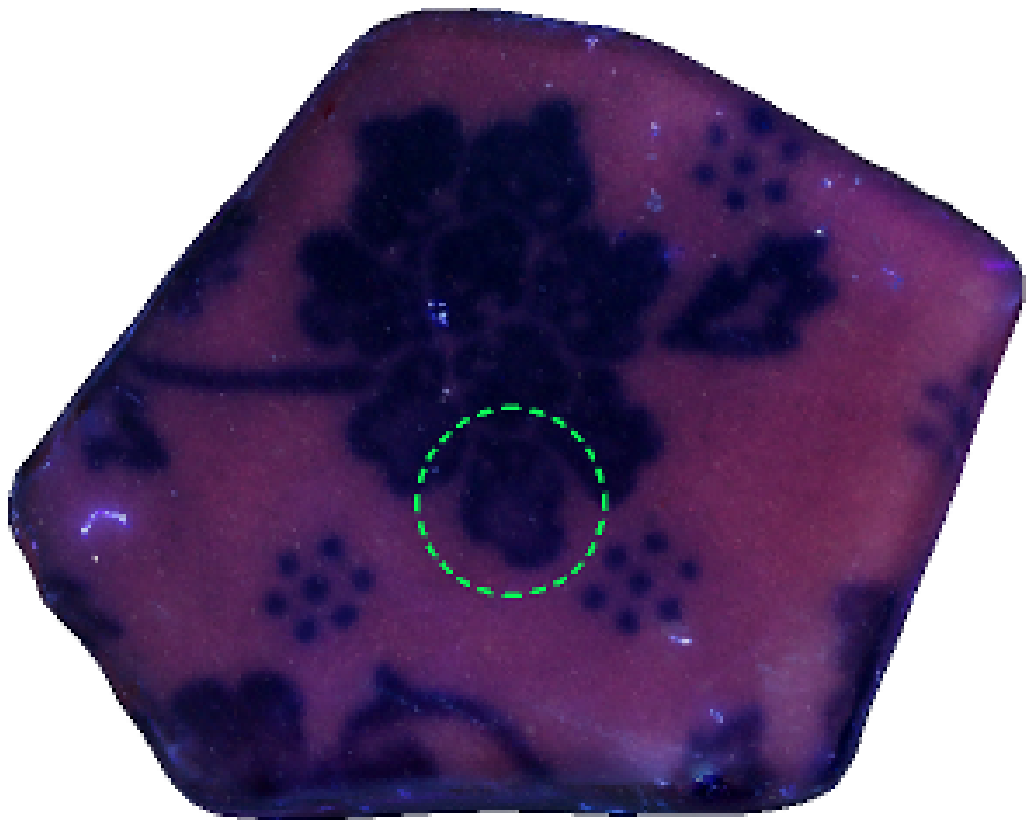} & 
    \includegraphics[width=0.225\hsize]{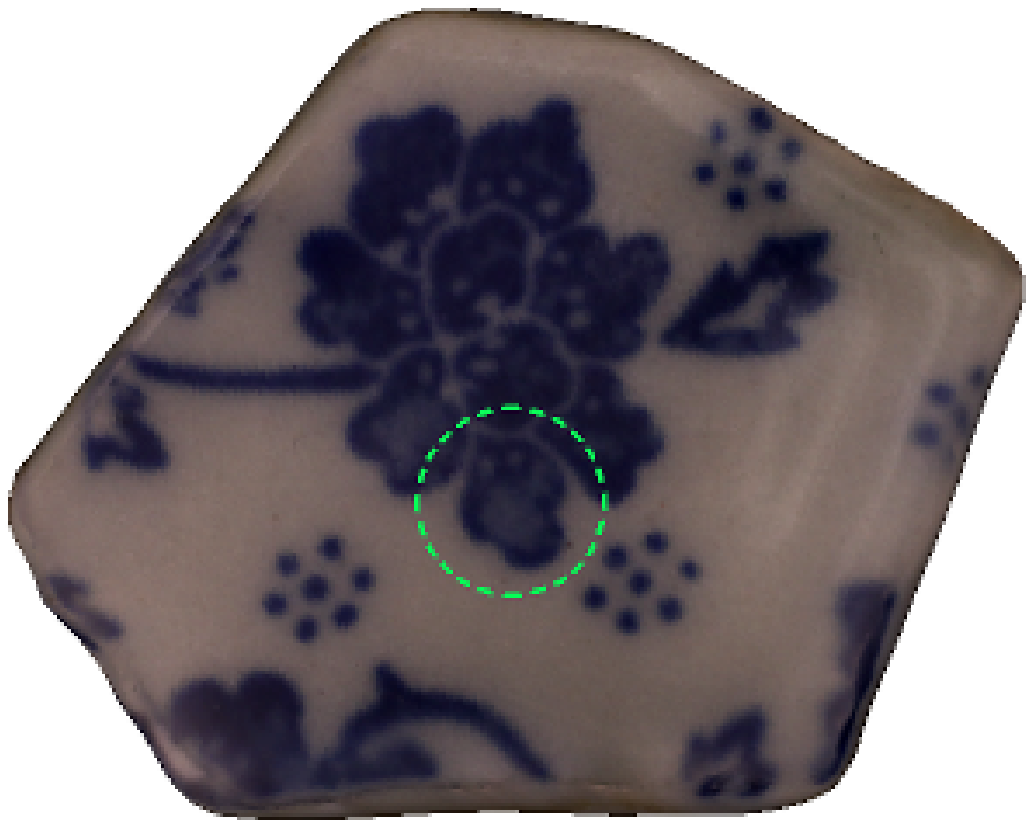} &
	\includegraphics[width=0.225\hsize]{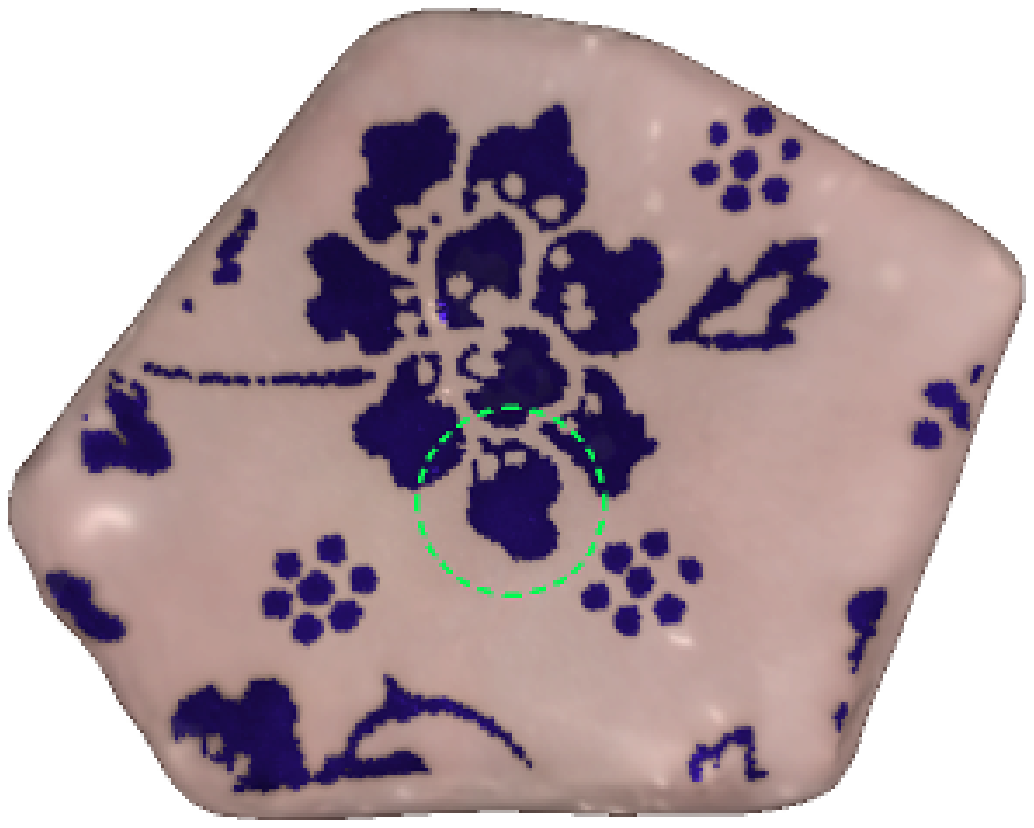} &	
	\includegraphics[width=0.225\hsize]{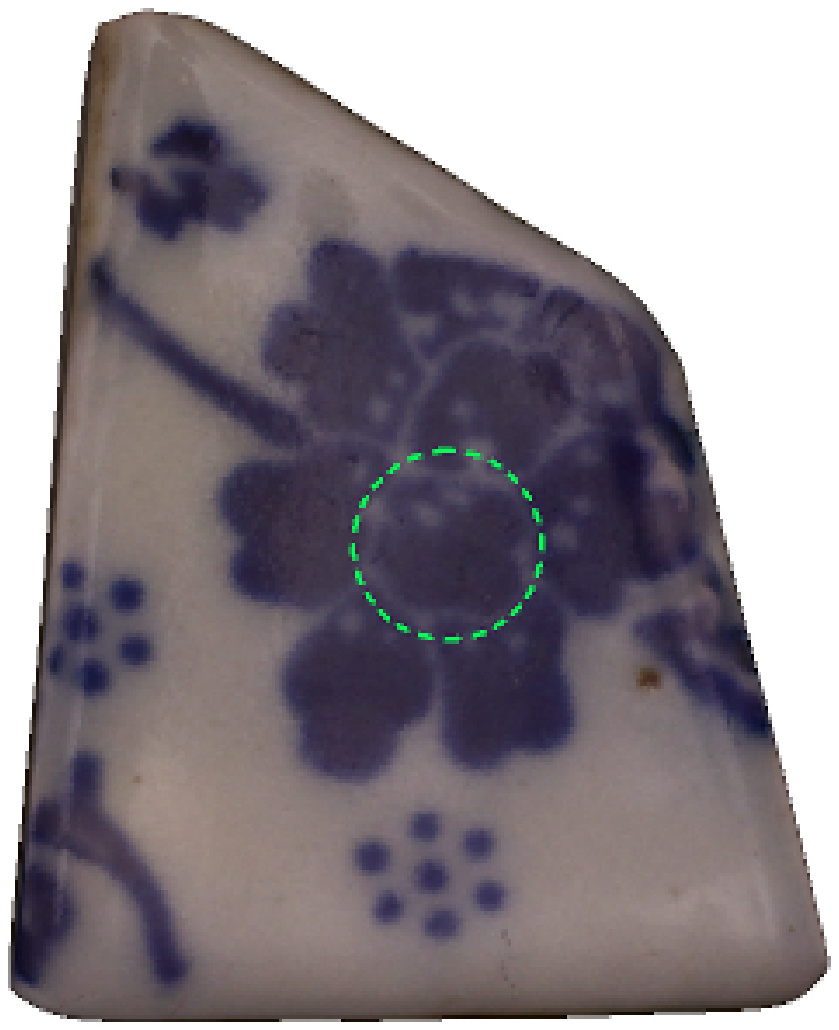} \\
	
	\includegraphics[width=0.225\hsize]{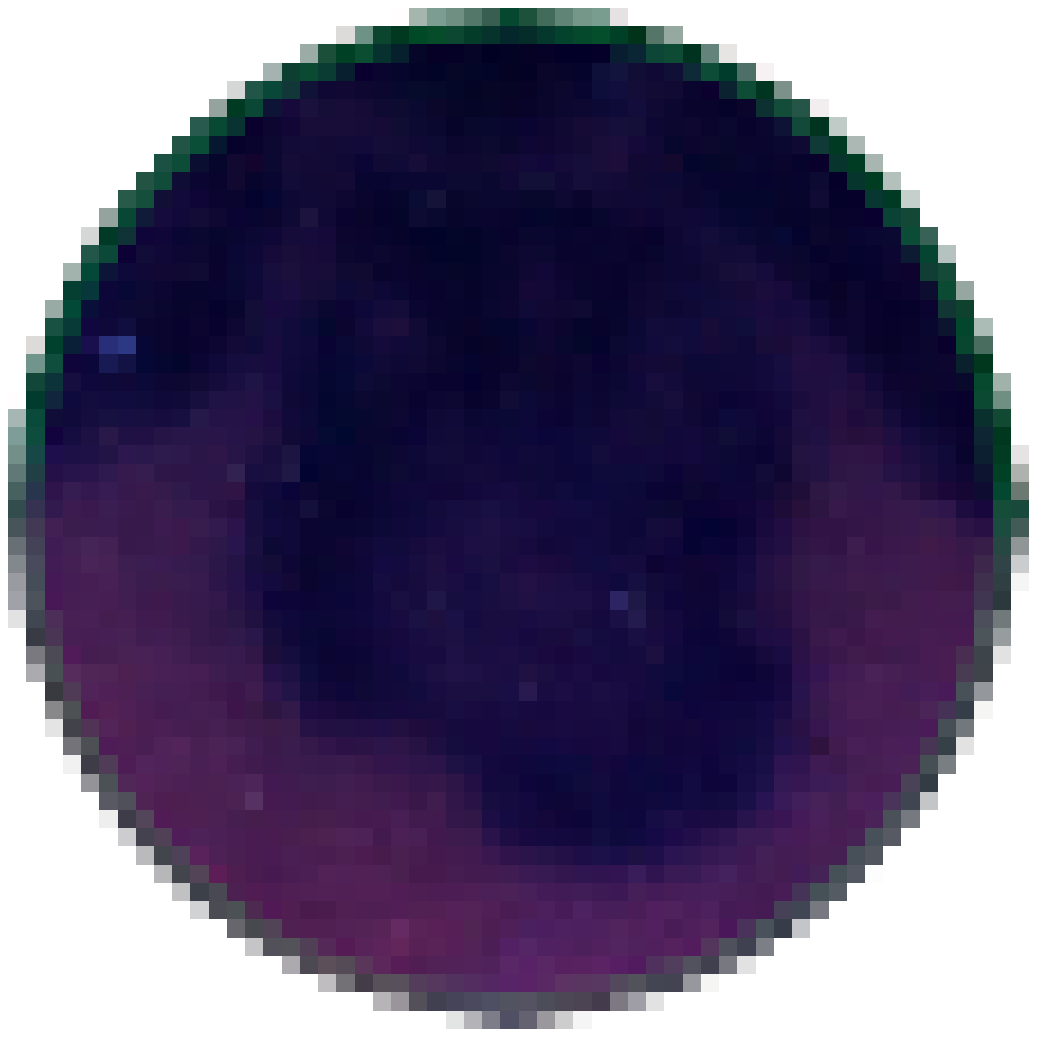} & 
    \includegraphics[width=0.225\hsize]{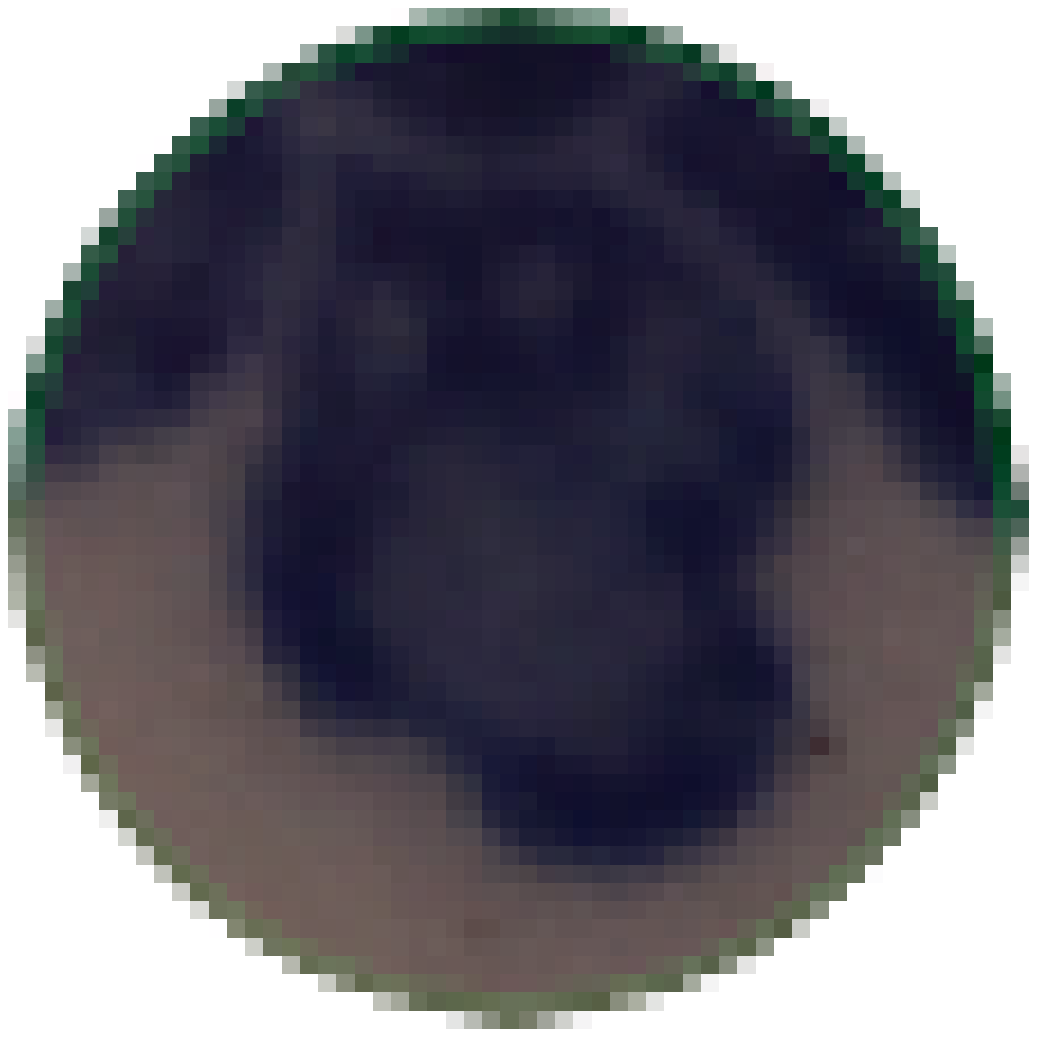} &
	\includegraphics[width=0.225\hsize]{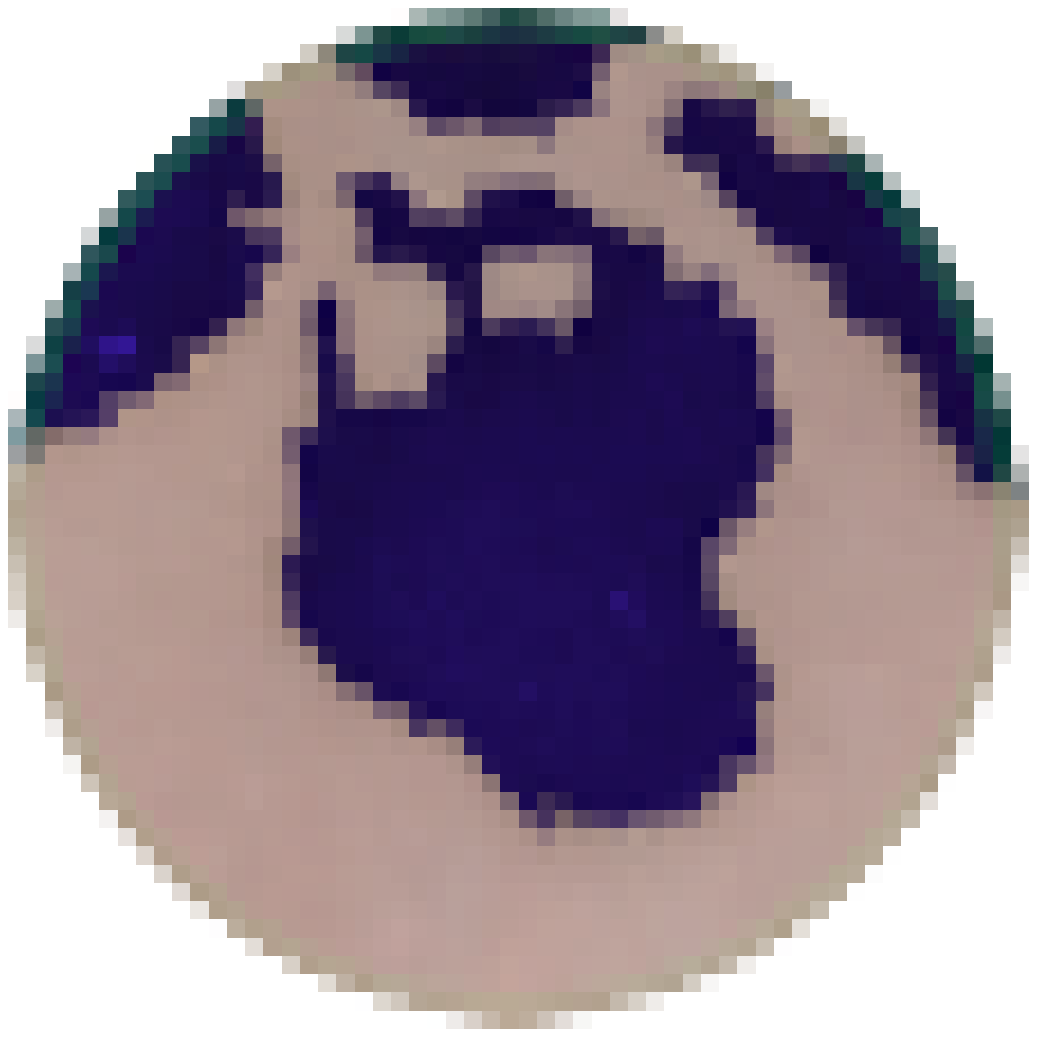} &	
	\includegraphics[width=0.225\hsize]{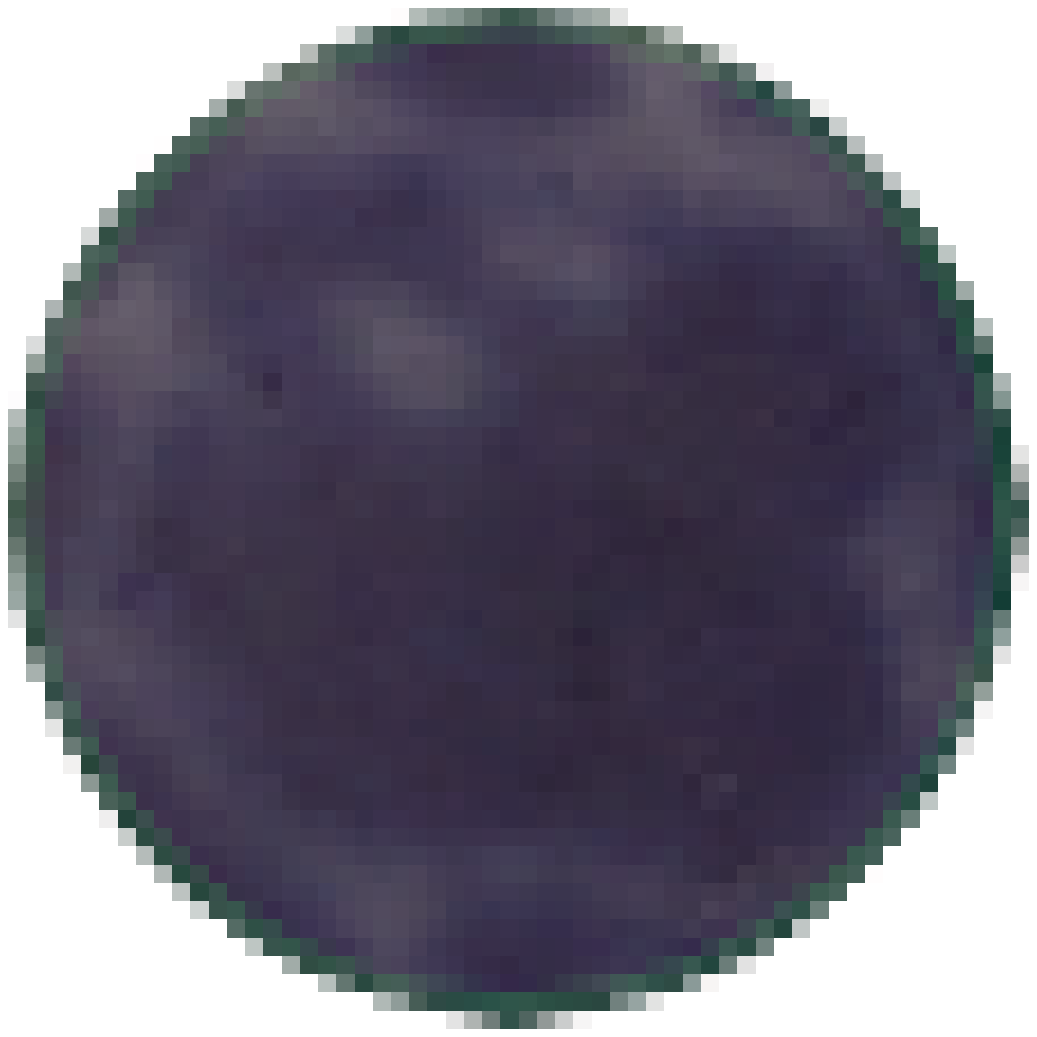} \\
\end{tabular}
\caption{\label{fig:restoration}%
Restoring color of a faded tile. The closeups show that we restore the color of faded petals despite variations in petal shape between the source and the target.}
\end{figure}

%% file: fig-material_assignment.tex
\begin{figure}[h]
\centering
\includegraphics[width=0.9\linewidth]{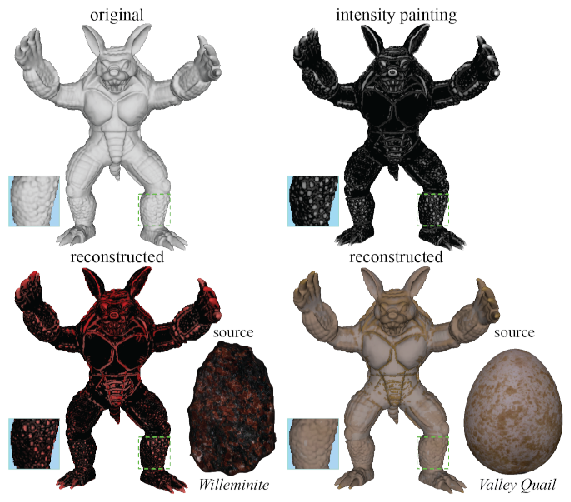}
\caption{\label{fig:material_assignment}% 
Assigning natural materials to a synthetic object. (top) (left-to-right) Synthetic object, intensity painted by an artist, (bottom) reconstructed.}
\end{figure}

%% file: conclusion.tex
\section{Discussion and Conclusion}
\label{sec:discussion}

We present a \hbox{3-D} material style transfer framework that replicates complex color patterns found in nature. An innovative component of our work is the use of fluorescent emission to measure and characterize surface materials. We use spherical harmonics to learn relationships between measured material properties and surface appearance on an exemplar, which we adapt to the material structure of a target object using novel bidirectional mapping functions. Quantitative and qualitative evaluation metrics show that we create highly plausible results.

Spherical harmonic functions over-smooth sharp edges making it difficult to reconstruct sharp changes in materials in a \emph{hackmanite} sample (Figure~\ref{fig:limitations_smoothing}). Increasing the number of data samples in sphere space mitigates this problem~\cite{McEwen2011,Leistedt2013} but requires increasingly higher order spherical harmonics, and exponential growth in computation time. Combining spherical harmonics with wavelet decomposition to capture high frequency details is a promising future direction.  

Run-time performance depends on the resolution of the input \hbox{3-D} meshes and the order of the spherical harmonic functions used to compute PDFs. High resolution meshes ($150K$-$185K$ vertices) and spherical harmonics of the order $n=150$ capture variations in material PDFs in a manner that balances efficiency with accuracy. Higher values of $n$ cause numerical errors due to hardware limitations with floating-point precision. It takes two days to run our algorithm end-to-end on a single dataset using an intel $2.6G$Hz $i9$ processor with $18$ cores. We optionally run datasets in parallel on a CPU cluster. Future work will explore a GPU implementation.

Specularities in $\mathbf{R}_{bis}$ occur when visible light overlaps the ultraviolet source. These artifacts cause patch segmentation and color reconstruction errors (Figure~\ref{fig:limitations_specularity}). We found only four examples of this in our data, and overall, our reconstruction errors were low. Future work will add a calibration step that removes reflected components from more diffuse fluorescent emission~\cite{Treibitz2012}) in $\mathbf{R}_{bis}$. Typically, this painstaking process records surface emission at narrow intervals ($5nm$) for all excitation wavelengths at the same intervals~\cite{tominaga2017appearance}, and decomposes measurements using a Donaldson Matrix~\cite{tominaga2017appearance}. We will explore a modified low-cost approximation over a smaller sample space ($370nm-480nm$) where visible light leakage may occur.

\input{fig-limitations_smoothing}

\input{fig-limitations_specularity}

Our system requires two objects of the same material, and does not compensate for other nanostructures that influence pattern coloration~\cite{FecheyrLippens2015}.  Multi-wavelength analysis is required to consider a broader range of complex materials. Although we do not reconstruct non-fluorescent materials (Section~\ref{sec:nonfluorescent} Figure~\ref{fig:mineral}), we anticipate that methods presented are extendible to other material reflectance relationships. Near-infrared detail enhancement may improve shape detail in eroded objects~\cite{tolerfranklin2021}.

%% file: fig-limitations_smoothing.tex
\begin{figure}[t]
  \centering
    \includegraphics[width=0.75\linewidth]{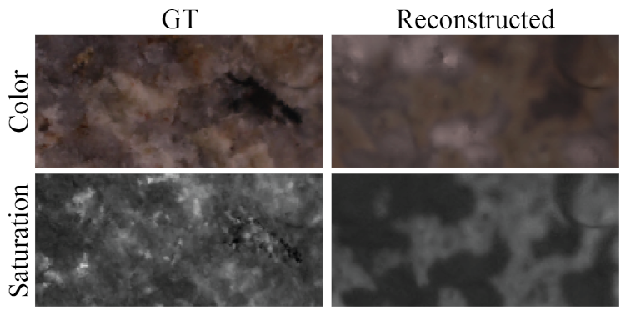}
    \caption{Spherical harmonic functions over-smooth sharp edges making it difficult to reconstruct sharp changes in material.}
	\label{fig:limitations_smoothing}
\end{figure}

%% file: fig-limitations_specularity.tex
\begin{figure}[h]
  \centering
    \includegraphics[width=0.75\linewidth]{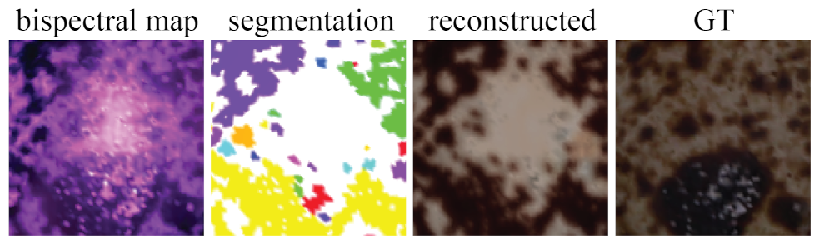}
    \caption{Specularities in the bispectral map interfere with material measurements resulting in errors in patch segmentation and material assignment.}
	\label{fig:limitations_specularity}
\end{figure}

%% file: acknowledgements.tex
\section{Acknowledgements}
This material is based upon work supported by the National Science Foundation under Grant No. 1510410. This work was conducted at the UF Graphics Imaging and Light Measurement Lab (GILMLab).

%% file: appendix.tex
\appendix

\subsection{Conformal Mapping}
\label{sec:appendconformalmapping}
An analytic function $f:U \rightarrow V$, $U, V \in \mathbf{R^{n}}$, is conformal at a point $u_{0} \in U$ if it preserves the magnitude and orientation of angles between the directed curves through $u_{0}$ ($f$ has a non-zero derivative at $u_{0}$).  Let $\Sigma \in \mathbf{R_{3}}$ be a genus zero smooth model. Let $p$ be a point on the surface of $\Sigma$ and let $\delta_{p}$ denote the Dirac delta (impulse) function at $p$. Let $\mathbf{S^{2}}$ denote a unit sphere in $\mathbf{R^{3}}$ and let $\mathbf{C}$ be a complex plane. A conformal map $z:\Sigma\setminus\{p\}\rightarrow \mathbf{S^{2}\setminus\{north pole\}}$ can be computed by solving the second order partial differential equation over the surface of $\Sigma$:

\begin{equation}
\label{eq:conformalpde}
\Delta z = \left(  \frac{\partial}{\partial u} - \iota \frac{\partial}{\partial v} \right) \delta_{p}
\end{equation}

\noindent where $u$ and $v$ are conformal coordinates defined in the neighborhood of $p$, $\iota$ is the square root of $-1$ and $\Delta$ is the Laplace-Beltrami operator. In pre-processing, landmarks in form curves are drawn over the two surfaces and correspondences assigned. Energy is computed by iteratively minimizing the Euclidean distance between the corresponding curves on the two surfaces.

\subsection{Harmonic Energies}
\label{sec:appendharmonicenergy}

The harmonic energy $E$ of a genus-0 triangulated mesh $\mathbf{u}$ is the sum of the string energies associated with all edges in $\mathbf{u}$: 

\begin{equation}
E = \sum_{[v_{i}, v_{j}] \in \mathbf{u}} k_{ij} \lvert \mathbf{f}(v_{i}) - \mathbf{f}(v_{j}) \rvert
\label{equ:diskharmonic}
\end{equation}.

\noindent where $\mathbf{f}(v_{i})$ is the conformal mapping of $\mathbf{u}$ to a unit disk and $k_{ij}$ is the string constant associated with the edge $[v_{i}, v_{j}]$~\cite{Yau2016ComputationalCG}. This mapping optimization is applicable to all topological disks.